\documentclass{article}

\PassOptionsToPackage{numbers, compress}{natbib}

\usepackage[main, final]{neurips_2026}

\usepackage[utf8]{inputenc} 
\usepackage[T1]{fontenc}    
\usepackage[backref=page]{hyperref}       
\usepackage{url}            
\usepackage{booktabs}       
\usepackage{multirow}       
\usepackage{amsfonts}       
\usepackage{nicefrac}       
\usepackage{microtype}      
\usepackage{xcolor}         
\usepackage{wrapfig}        

\newif\ificml 

\usepackage{mathtools}            
\usepackage{amssymb}              
\usepackage{bm}                   
\usepackage{xspace}               
\usepackage{pifont}               
\usepackage{tikz}                 
\usetikzlibrary{arrows.meta,positioning,calc,fit,shapes.geometric,backgrounds}
\usepackage{lato}
\newcommand{\latofont}{\fontfamily{\latofamily}\selectfont}
\usepackage[capitalize,noabbrev]{cleveref}  
\usepackage{subcaption}
\usepackage{wrapfig}

\usepackage{xcolor}
\usetikzlibrary{decorations.pathreplacing}

\definecolor{g1}{HTML}{3F5F7C}
\definecolor{g2}{HTML}{6B87A2}
\definecolor{g3}{HTML}{9EB1C5}
\definecolor{g4}{HTML}{CBD5E0}
\definecolor{clsCol}{HTML}{2F4858}     
\definecolor{maskCol}{HTML}{EAEAEA}    
\definecolor{gA}{HTML}{4C78A8}
\definecolor{gB}{HTML}{F58518}
\definecolor{gC}{HTML}{54A24B}
\definecolor{gD}{HTML}{B279A2}
\definecolor{clsCol}{HTML}{2F4858}
\definecolor{blockFill}{HTML}{E8ECF1}
\definecolor{blockEdge}{HTML}{8A94A3}
\definecolor{gA}{HTML}{4C78A8}
\definecolor{gB}{HTML}{F58518}
\definecolor{gC}{HTML}{54A24B}
\definecolor{gD}{HTML}{B279A2}
\definecolor{clsCol}{HTML}{2F4858}
\definecolor{blockFill}{HTML}{E8ECF1}
\definecolor{blockEdge}{HTML}{8A94A3}
\definecolor{gA}{HTML}{4C78A8}
\definecolor{gB}{HTML}{F58518}
\definecolor{gC}{HTML}{54A24B}
\definecolor{gD}{HTML}{B279A2}
\definecolor{clsCol}{HTML}{2F4858}
\definecolor{blockFill}{HTML}{E8ECF1}
\definecolor{blockEdge}{HTML}{8A94A3}

\DeclareRobustCommand{\etal}{\textit{et al.}\xspace}

\title{Hypnos: Learning Generalisable Representations of Sleep Physiology via Next-Token Prediction}
\hypersetup{
  pdftitle={Hypnos: Learning Generalisable Representations of Sleep Physiology via Next-Token Prediction},
  pdfauthor={Jonathan F. Carter, Lionel Tarassenko},
  pdfsubject={NeurIPS 2026},
  pdfkeywords={sleep, polysomnography, foundation models, self-supervised learning, next-token prediction}
}

\author{%
  Jonathan F. Carter\thanks{jonathan.carter@eng.ox.ac.uk} \\
  Institute of Biomedical Engineering\\
  University of Oxford\\
  \And
  Lionel Tarassenko \\
  Institute of Biomedical Engineering\\
  University of Oxford\\
}

\begin{document}

\maketitle

\begin{abstract}
  Foundation models offer a promising route to compress multi-modal physiological signals into compact representations of human health, with broad applications across sleep medicine, cardiology, neurology and other healthcare domains. Existing models have typically been trained with masked-reconstruction or contrastive objectives. However, masked reconstruction may be poorly suited to the stochastic nature of these signals, while contrastive approaches rely on positive-pair definitions despite the semantic invariances of physiological signals being poorly understood. In this work, we show that next-token prediction is a simple and scalable alternative. We develop Hypnos, a multi-modal sleep foundation model trained using eight different sensing modalities (e.g. EEG, ECG, respiratory signals) drawn from over 20{,}000 overnight polysomnography recordings. We tokenize each modality into streams of discrete tokens using residual vector quantization, then train a large auto-regressive RQ-Transformer to jointly predict the next token across all modalities in parallel. After training, Hypnos can be applied to continuous streams of sensor data from any subset of supported modalities, generating embeddings for downstream tasks. Hypnos significantly outperforms existing foundation models on most benchmarks evaluated. In sleep stage classification, we match the performance of strong supervised baselines on held-out test sets whilst using \(100\times\) less labelled data. Hypnos even generalises beyond sleep recordings, surpassing a dedicated ECG foundation model at detecting atrial fibrillation. Our results demonstrate that auto-regressive token modelling is a strong self-supervised approach for representation learning from multi-modal physiological signals.
\end{abstract}

\section{Introduction}

\begin{figure}[t]
  \centering
  \includegraphics[width=\textwidth]{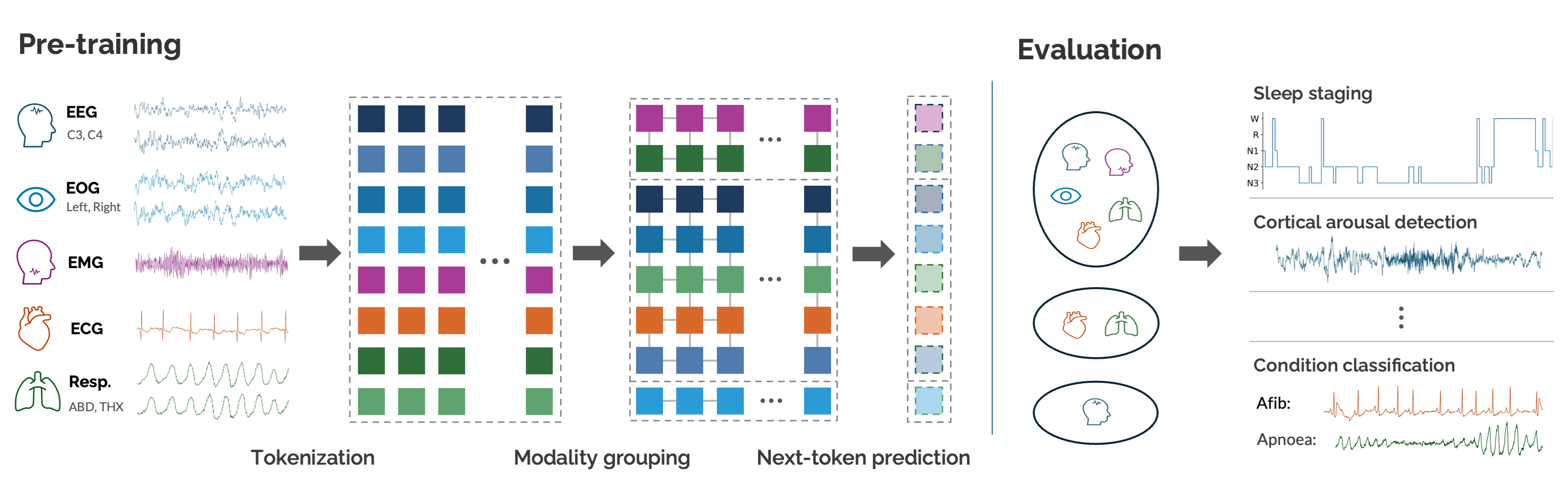}
  \caption{\textbf{Overview.} Hypnos is a large auto-regressive RQ-transformer trained via multi-modal next-token prediction on tokenized streams of physiological sensor data. During pre-training, cross-modal attention is restricted to randomly sampled sub-groups, improving test-time generalisation to subsets of supported modalities. After pre-training, Hypnos can be used to generate high-quality embeddings for a diverse range of sensor configurations and downstream tasks over different timescales. We evaluate on tasks including sleep stage classification of overnight recordings, the detection of cortical arousal events, and atrial fibrillation (Afib) detection.\label{fig:overview}}
\end{figure}

Physiological recordings such as polysomnography (PSG) capture hours of continuous, multi-modal sensor data from the brain and body. A single overnight study with eight channels recorded at 128 Hz yields over 30 million data points. How can we compress minutes, hours, or even days of continuous, multi-modal sensor data into better measures of health for tasks such as identifying neurodegenerative or cardiovascular disease? A key motivation for physiological foundation models is to use large quantities of unlabelled sensor data to address this challenge. Prior work has shown that self-supervised learning (SSL) techniques can be used to learn effective representations from a broad range of physiological sensors~\cite{banvilleUncoveringStructureClinical2021, abbaspourazadLargescaleTrainingFoundation2024,yuanSelfsupervisedLearningHuman2024}. These models have predominantly been trained using either contrastive learning~\cite{thapaMultimodalSleepFoundation2026,abbaspourazadLargescaleTrainingFoundation2024} or masked input-space reconstruction~\cite{leeHiMAEHierarchicalMasked2025,narayanswamyScalingWearableFoundation2024,foxFoundationalTransformerLeveraging2025} approaches. However, each has known shortcomings on stochastic, continuous signals (see \Cref{sec:background}). Next-token prediction is a simple alternative: it underpins modern Large Language Models (LLMs)~\cite{radfordImprovingLanguageUnderstanding2018,radfordLanguageModelsAre2019,brownLanguageModelsAre2020}, has been demonstrated to scale to context lengths exceeding 1\,M tokens~\cite{teamGemini15Unlocking2024}, and has been successfully applied in the analogous, continuous-signal domain of audio~\cite{borsosAudiolmLanguageModeling2023,defossezMoshiSpeechtextFoundation2024}.

In this work, we show that next-token prediction is a simple and scalable self-supervised learning objective for multi-modal physiological sensor data (\Cref{fig:overview}). We introduce Hypnos, a sleep foundation model trained on eight sensing modalities including electroencephalogram (EEG), electrocardiogram (ECG), and respiratory signals. Each modality is tokenized into a stream of discrete tokens using residual vector quantization (RVQ), and an auto-regressive Transformer is trained to jointly predict the next token across modalities in parallel. Using over 20,000 overnight PSG recordings drawn from nine public datasets, we find that both next-token perplexity and downstream probing performance continue to improve with scale. Our key contributions are as follows:
\begin{itemize}
  \item \textbf{Physiological next-token prediction:} We demonstrate that next-token prediction is an effective method for joint self-supervised learning from diverse physiological signals, unifying generative modelling and representation learning into a single architecture.
  \item \textbf{State-of-the-art performance:} Hypnos significantly outperforms prior sleep foundation models on most benchmark tasks and datasets evaluated. Embeddings from Hypnos even transfer beyond sleep, outperforming a dedicated ECG foundation model on external single-lead ECG benchmarks.
  \item \textbf{Deployment-focused design:} Hypnos can run in a streaming fashion, supports diverse sensor combinations and arbitrary length recordings, and generates embeddings at a convenient rate of 1\,Hz. This can enable real-time applications such as closed-loop neuromodulation.
\end{itemize}
Pre-trained model weights and inference code are available at: \href{https://github.com/joncarter1/hypnos}{https://github.com/joncarter1/hypnos}.
\section{Background and Motivation}\label{sec:background}

\paragraph{Vector Quantization}
Existing physiological foundation models typically operate on short windows of sensor data, e.g.\ 5 minutes or less~\cite{thapaMultimodalSleepFoundation2026,shuaiOSFPretrainingScaling2026,mckeenECGFMOpenElectrocardiogram2025}, or on derived quantities such as step counts and heart rate statistics~\cite{narayanswamyScalingWearableFoundation2024}. SleepMaMi~\cite{parkSleepMaMiUniversalSleep2026} is a notable exception, training a Mamba-based macro-encoder over full-night sequences of epoch-level embeddings from a frozen micro-encoder. High sampling rates are essential to capture fine details from modalities such as EEG or ECG, e.g.\ peak-peak timings. However, these signals are also highly compressible. For example, it has long been known that continuous ECG data can be modelled using coupled differential equations with a small number of latent variables~\cite{mcsharryDynamicalModelGenerating2003}. This motivates an initial compression of the data (e.g. tokenization) prior to sequence modelling. Vector quantization~\cite{vandenoordNeuralDiscreteRepresentation2017} learns a discrete codebook that maps continuous inputs to tokens, and has been used to build foundation models across domains including brain~\cite{jiangLargeBrainModel2024,xiaoBrainOmniBrainFoundation2025} and audio~\cite{copetSimpleControllableMusic2023,borsosAudiolmLanguageModeling2023} signals. NeuroLM~\cite{jiangNeuroLMUniversalMultitask2024} trained an autoregressive Transformer over single-codebook VQ tokens of EEG signals. We extend this line of work to multi-modal physiological signals, using residual vector quantization and an RQ-Transformer~\cite{leeAutoregressiveImageGeneration2022} to model multiple high-rate streams in parallel. Our technical approach, discussed in \Cref{sec:method}, is most similar to Moshi~\cite{defossezMoshiSpeechtextFoundation2024}, a state-of-the-art speech-text foundation model.

\begin{figure}[h]
  \centering
  \resizebox{0.9\linewidth}{!}{\latofont\begin{tikzpicture}[
    font=\footnotesize,
    >={Stealth[length=2.0mm,width=1.6mm]},
    trace/.style={draw=black!85, line width=0.65pt, line cap=round, line join=round},
    ghost/.style={line width=0.75pt, line cap=round, line join=round},
    bar/.style={draw=none},
    sublbl/.style={font=\scriptsize, text=black!65},
    panellbl/.style={font=\footnotesize\bfseries, text=black!80},
    masklbl/.style={font=\scriptsize\itshape, text=black!55},
]

\newcommand{\beatR}[1]{%
  ({#1+0.000},0)
  ({#1+0.060},0)
  ({#1+0.090},0.07) ({#1+0.120},0.09) ({#1+0.150},0.07)
  ({#1+0.180},0)
  ({#1+0.225},0)
  ({#1+0.255},0.65)
  ({#1+0.285},0)
  ({#1+0.500},0)
  ({#1+0.580},0.14) ({#1+0.640},0.18) ({#1+0.700},0.14) ({#1+0.760},0)
  ({#1+1.000},0)
}

\newcommand{\beatA}[1]{\beatR{#1}}

\def\topy{1.35}     
\def\boty{0.0}      
\def\xobsL{0.0}     
\def\xmaskL{4.6}    
\def\xmaskR{5.9}    
\def\histx{7.2}     
\def\masktop{0.78}  
\def\maskbot{-0.35} 
\def\histtop{0.78}  
\def\histamp{0.65}  

\fill[maskCol, draw=black!25, line width=0.25pt, rounded corners=0.6pt]
  (\xmaskL,{\topy+\maskbot}) rectangle (\xmaskR,{\topy+\masktop});
\node[masklbl, anchor=south] at ({(\xmaskL+\xmaskR)/2}, {\topy+\masktop+0.02}) {$z_{t+1}$};

\draw[trace, yshift=\topy cm] plot coordinates {
  \beatR{0.00} \beatR{1.10} \beatR{2.20} \beatR{3.30}
  ({4.30},0) ({\xmaskL},0)
};

\draw[ghost, color=gA, yshift=\topy cm] plot coordinates {
  ({\xmaskL},0) \beatR{4.70} ({\xmaskR},0)
};

\def\hbT{\topy}
\foreach \i/\h/\c in {0/0.04/black!30, 1/0.06/black!30, 2/0.10/black!30, 3/0.85/gA, 4/0.10/black!30, 5/0.06/black!30, 6/0.04/black!30} {
  \fill[\c] ({\histx + 0.16*\i}, \hbT)
    rectangle ({\histx + 0.16*\i + 0.12}, {\hbT + \h*\histamp});
}
\draw[->, draw=black!55, line width=0.25pt]
  ({\histx-0.05}, \hbT) -- ({\histx + 1.20}, \hbT);
\draw[->, draw=black!55, line width=0.25pt]
  ({\histx-0.05}, \hbT) -- ({\histx-0.05}, {\hbT + \histtop});
\node[sublbl, anchor=north] at ({\histx + 0.55}, {\hbT - 0.04}) {next token};
\node[sublbl, anchor=south, rotate=90] at ({\histx - 0.32}, {\hbT + \histtop/2}) {$p$};

\node[panellbl, anchor=east] at (-0.15, \topy) {Regular};

\fill[maskCol, draw=black!25, line width=0.25pt, rounded corners=0.6pt]
  (\xmaskL,{\boty+\maskbot}) rectangle (\xmaskR,{\boty+\masktop});

\draw[trace, yshift=\boty cm] plot coordinates {
  \beatA{0.00} \beatA{1.05} \beatA{2.50} \beatA{3.55}
  ({\xmaskL},0)
};

\draw[ghost, color=gB, opacity=0.70, yshift=\boty cm] plot coordinates {
  ({\xmaskL},0) \beatA{4.55} ({\xmaskR},0)
};
\draw[ghost, color=gA, opacity=0.70, yshift=\boty cm] plot coordinates {
  ({\xmaskL},0) \beatA{4.75} ({\xmaskR},0)
};
\draw[ghost, color=gC, opacity=0.70, yshift=\boty cm] plot coordinates {
  ({\xmaskL},0) \beatA{4.95} ({\xmaskR},0)
};

\def\hbB{\boty}
\foreach \i/\h/\c in {0/0.04/black!30, 1/0.06/black!30, 2/0.55/gB, 3/0.65/gA, 4/0.55/gC, 5/0.06/black!30, 6/0.04/black!30} {
  \fill[\c] ({\histx + 0.16*\i}, \hbB)
    rectangle ({\histx + 0.16*\i + 0.12}, {\hbB + \h*\histamp});
}
\draw[->, draw=black!55, line width=0.25pt]
  ({\histx-0.05}, \hbB) -- ({\histx + 1.20}, \hbB);
\draw[->, draw=black!55, line width=0.25pt]
  ({\histx-0.05}, \hbB) -- ({\histx-0.05}, {\hbB + \histtop});
\node[sublbl, anchor=north] at ({\histx + 0.55}, {\hbB - 0.04}) {next token};
\node[sublbl, anchor=south, rotate=90] at ({\histx - 0.32}, {\hbB + \histtop/2}) {$p$};

\node[panellbl, anchor=east] at (-0.15, \boty) {Arrhythmia};

\end{tikzpicture}}
  \caption{For stochastic signals such as an ECG (left), the distribution over the future may be well-characterised by a peaked distribution $p$ over some discretisation (i.e. tokenization) $z_{t+1}$. For example, during arrhythmia, a subset of tokens with similar QRS morphologies may be equally likely.\label{fig:ntp-stochasticity}}
\end{figure}
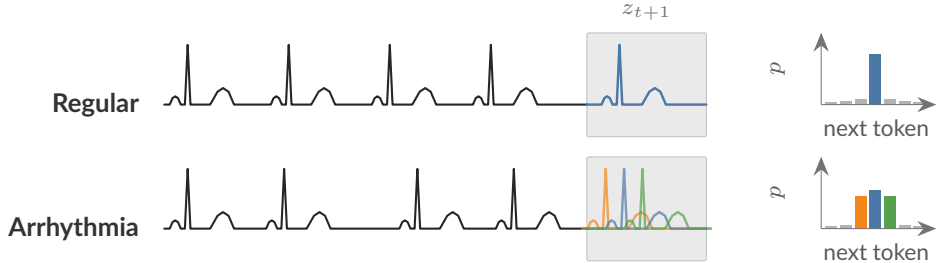

\paragraph{Stochasticity} Next-token prediction is well-suited to handle the stochastic nature of physiological signals. Rather than trying to exactly reconstruct physiological signals in input space, e.g.~\cite{leeHiMAEHierarchicalMasked2025, foxFoundationalTransformerLeveraging2025,wangEegptPretrainedTransformer2024}, which may be sensitive to exact peak timings and waveform morphology, next-token prediction can assign probability mass to the plausible subset of next tokens (see \Cref{fig:ntp-stochasticity}). Alternatively, the signal could be modelled using a continuous distribution, e.g.\ via diffusion~\cite{sohl-dicksteinDeepUnsupervisedLearning2015,hoDenoisingDiffusionProbabilistic2020}, as recently used in the analogous domain of audio~\cite{rouardContinuousAudioLanguage2026}. However, we adopt discrete next-token prediction in this work to inherit architectures, training recipes, and simple tractable likelihoods from audio and language modelling.

  \paragraph{Discretisation} Tokenization is itself a design bias. However, sensor data already has a finite sampling rate and precision, and its effective resolution is limited by the noise floor. Our tokenizers achieve low reconstruction error (\Cref{fig:tokenizer-recon} in the Appendix), and downstream performance is stable across a range of token lengths (\Cref{fig:token-duration}).

\paragraph{Multimodality} Contrastive methods are characterised by the use of positive pairs. For example, prior work has constructed positive pairs using physiological signals from different modalities over the same time range~\cite{zhangBrantXUnifiedPhysiological2024,thapaMultimodalSleepFoundation2026}, segments from adjacent time ranges~\cite{kiyassehCLOCSContrastiveLearning2021}, and segments from the same subject at different time ranges~\cite{pillaiPaPaGeiOpenFoundation2025}. Positive pair definitions and augmentations are inductive biases that implicitly shape the learnt latent space. For example, the motivation for the leave-one-out method used by SleepFM is to `encourage each embedding to align semantically with all other modalities'~\cite{thapaMultimodalSleepFoundation2026}. A potential limitation of SleepFM's leave-one-out construction is that it encourages the model to extract information \textit{shared} between modalities, and may discard information that is not~\cite{daunhawerIdentifiabilityResultsMultimodal2023}. SleepMaMi addresses this directly, pairing private per-modality encoders with a shared encoder and using additional non-contrastive losses~\cite{parkSleepMaMiUniversalSleep2026}. However, this approach widens the design space.

Different sensing modalities are intentionally recorded because they give complementary `views' of physiology, with independent variations providing useful information for downstream tasks. For example, brain activity measured by the EEG often looks similar between Wake and Rapid Eye Movement (REM) sleep, but is easily distinguished using activity measured by the electrooculogram (EOG)~\cite{iberAASMManualScoring2007}. This is unlike contrastive learning in vision~\cite{chenExploringSimpleSiamese2021}, where augmentations such as crops or colour jitters are chosen precisely because they do not change the underlying semantic content. For physiological signals, the space of invariant augmentations is not well understood. For example, should ECG embeddings from one subject during exercise be closer to that of another subject during exercise, or the same subject when stationary? Human-defined augmentations may limit the effectiveness of both contrastive learning and the related approach of predicting transformations applied to the input~\cite{yuanSelfsupervisedLearningHuman2024,jayalathBrainsBitterLesson2025}.

\paragraph{Sequential Learning} Finally, existing physiological foundation models have not been designed for sequential updates and streaming inference. This may be beneficial across a range of applications, ranging from closed loop neuromodulation to remote patient monitoring, by enabling real-time tracking of physiological state. The autoregressive nature of next-token prediction is well-suited for this downstream use case.
\section{Method}\label{sec:method}
\subsection{Datasets}
To train and evaluate our models, we use overnight polysomnography (PSG) recordings drawn from datasets available from the National Sleep Research Resource (NSRR,~\cite{zhangNationalSleepResearch2018}): SHHS~\cite{quanSleepHeartHealth1997}, CCSHS~\cite{rosenPrevalenceRiskFactors2003} CFS~\cite{redlineFamilialAggregationObstructive1995}, CHAT~\cite{marcusRandomizedTrialAdenotonsillectomy2013}, MESA~\cite{chenRacialEthnicDifferences2015}, MrOS~\cite{songRelationshipsSleepStages2015}, NCHSDB~\cite{leeLargeCollectionRealworld2022}, and WSC~\cite{youngBurdenSleepApnea2009}; using the same training, validation and test set splits for each dataset as Shuai~\etal~\cite{shuaiOSFPretrainingScaling2026}. We use the Dreem Open Datasets (DOD-H and DOD-O) for further external evaluation~\cite{guillotDreemOpenDatasets2020}. Collectively, these datasets contain over 20,000 overnight polysomnography recordings spanning a broad range of patient demographics, sensors and recording configurations.

\paragraph{Modalities}
\begin{wraptable}[11]{r}{0.45\textwidth}
  \centering
  \footnotesize
  \setlength{\tabcolsep}{4pt}
  \renewcommand{\arraystretch}{0.95}
  \caption{\textbf{Supported modalities.}\label{tab:modalities}}
  \begin{tabular}{lll}
    \toprule
    Modality         & Rate (Hz) & Tokenizer group \\
    \midrule
    EEG (C3--M2)     & 128       & EEG             \\
    EEG (C4--M1)     & 128       & EEG             \\
    EOG (E1--M2)     & 128       & EOG             \\
    EOG (E2--M1)     & 128       & EOG             \\
    Chin EMG         & 128       & EMG             \\
    ECG              & 128       & ECG             \\
    Abdominal effort & 32        & Resp            \\
    Thoracic effort  & 32        & Resp            \\
    \bottomrule
  \end{tabular}
\end{wraptable}
We use up to \(M=8\) modalities per recording, drawn from five physiological modality groups commonly available across the NSRR cohorts: central electroencephalogram (EEG), electrooculogram (EOG), chin electromyogram (EMG), electrocardiogram (ECG), and abdominal (ABD) / thoracic (THX) respiratory effort. These were selected based on their prevalence across cohorts and their relevance to downstream sleep analysis tasks. \Cref{tab:modalities} summarises the supported input channels to Hypnos and the sampling rates used for each modality.

\paragraph{Pre-processing}
Each channel was minimally pre-processed by resampling to a consistent rate across recordings before filtering and normalisation. Full pre-processing details are given in \Cref{app:preproc}.

\subsection{Tokenizing Physiological Signals}\label{section:tokenization}
\begin{wrapfigure}{R}{0.45\textwidth}
  \centering
  \resizebox{\linewidth}{!}{\latofont
\begin{tikzpicture}[
    font=\scriptsize,
    >={Stealth[length=2.2mm,width=1.8mm]},
    tok/.style={draw=none,minimum width=3.5mm,minimum height=3.0mm,
                inner sep=0pt,rounded corners=0.3pt},
    sigbox/.style={draw=blockEdge,fill=white,rounded corners=1.5pt,
                   minimum width=8.4mm,minimum height=26mm,inner sep=0pt},
    nnbox/.style={draw=blockEdge,fill=blockFill,rounded corners=1.5pt,
                  minimum width=6.8mm,minimum height=22mm,inner sep=2pt,
                  font=\footnotesize},
    groupbox/.style={draw=black!35,dashed,line width=0.5pt,rounded corners=2.5pt,
                     fill=black!4},
    flow/.style={->,draw=black!70,line width=0.5pt},
    sublbl/.style={font=\footnotesize,text=black!60},
    grouplbl/.style={font=\footnotesize,text=black!75},
    vbox/.style={draw=black!35,rounded corners=1.5pt,line width=0.4pt},
    sigwave/.style={line width=0.55pt,gA!85!black},
]

\def\modcol{gD}
\def\T{6}
\def\dx{0.42}
\def\dy{0.32}

\def\sigHW{0.42}                     
\def\sigHH{1.30}                     
\def\nnHW{0.34}                      
\def\arrowGap{0.40}                  
\def\innerGap{0.28}                  
\def\xSig{0}                         

\pgfmathsetmacro{\xEncSea}{\sigHW + \arrowGap + \nnHW}
\pgfmathsetmacro{\xEncTr}{\xEncSea + 2*\nnHW + \innerGap}

\pgfmathsetmacro{\xRVQ}{\xEncTr + \nnHW + \arrowGap + \nnHW}        

\pgfmathsetmacro{\xGridL}{\xRVQ + \nnHW + \arrowGap}                
\pgfmathsetmacro{\xTokL}{\xGridL + 0.25}                            
\pgfmathsetmacro{\xTokR}{\xTokL + (\T-1)*\dx}                       
\pgfmathsetmacro{\xGridR}{\xTokR + 0.25}                            

\pgfmathsetmacro{\xDecTr}{\xGridR + \arrowGap + \nnHW}
\pgfmathsetmacro{\xDecSea}{\xDecTr + 2*\nnHW + \innerGap}
\pgfmathsetmacro{\xSigOut}{\xDecSea + \nnHW + \arrowGap + \sigHW}

\def\yGrid{0.72}                     
\def\yGroup{1.25}                    
\def\groupPad{0.12}                  

\node[sigbox] (sigin) at (\xSig,0) {};
\begin{scope}[shift={(\xSig,0)}]
  \path[clip] (-0.38,-1.22) rectangle (0.38,1.22);
  \draw[sigwave]
    plot[domain=-1.20:1.20,samples=300,smooth]
      ({0.21*sin(720*\x*0.95)+0.11*sin(1500*\x+30)+0.045*sin(2700*\x+15)},\x);
\end{scope}
\node[above=2pt of sigin,font=\footnotesize] {$X^{i}\in\mathbb{R}^{f\cdot T}$};

\node[groupbox,fit={({\xEncSea-\nnHW-\groupPad},-\yGroup)
                    ({\xEncTr+\nnHW+\groupPad}, \yGroup)},
      inner sep=0pt] (encgrp) {};
\node[grouplbl,above=2pt of encgrp.north] {Encoder};

\node[nnbox] (encSea) at (\xEncSea,0) {\rotatebox{90}{SeaNet}};
\node[nnbox] (encTr)  at (\xEncTr,0)  {\rotatebox{90}{Transformer}};

\draw[flow] (sigin.east) -- (encSea.west);
\draw[flow] (encSea.east) -- (encTr.west);

\node[nnbox] (rvq) at (\xRVQ,0) {\rotatebox{90}{RVQ}};
\draw[flow] (encTr.east) -- (rvq.west);

\foreach \t in {1,...,\T} {
  \foreach \k/\shade in {1/100,2/72,3/48,4/28} {
    \pgfmathsetmacro{\yk}{(2.5 - \k)*\dy}
    \pgfmathsetmacro{\xt}{\xTokL + (\t-1)*\dx}
    \node[tok,fill=\modcol!\shade] at (\xt,\yk) {};
  }
}
\coordinate (gridSW) at (\xGridL,-\yGrid);
\coordinate (gridNE) at (\xGridR, \yGrid);
\coordinate (gridW)  at (\xGridL, 0);
\coordinate (gridE)  at (\xGridR, 0);
\coordinate (gridN)  at ({(\xGridL+\xGridR)/2}, \yGrid);
\draw[vbox] (gridSW) rectangle (gridNE);

\draw[flow] (rvq.east) -- (gridW);

\node[above=2pt,font=\footnotesize] at (gridN)
  {$V^{i}\in\{1,\dots,C\}^{K\times T}$};

\node[groupbox,fit={({\xDecTr-\nnHW-\groupPad},-\yGroup)
                    ({\xDecSea+\nnHW+\groupPad}, \yGroup)},
      inner sep=0pt] (decgrp) {};
\node[grouplbl,above=2pt of decgrp.north] {Decoder};

\node[nnbox] (decTr)  at (\xDecTr,0)  {\rotatebox{90}{Transformer}};
\node[nnbox] (decSea) at (\xDecSea,0) {\rotatebox{90}{SeaNet}};

\draw[flow] (gridE) -- (decTr.west);
\draw[flow] (decTr.east) -- (decSea.west);

\node[sigbox] (sigout) at (\xSigOut,0) {};
\begin{scope}[shift={(\xSigOut,0)}]
  \path[clip] (-0.38,-1.22) rectangle (0.38,1.22);
  \draw[sigwave]
    plot[domain=-1.20:1.20,samples=300,smooth]
      ({0.20*sin(720*\x*0.95+8)+0.10*sin(1500*\x+25)+0.045*sin(2700*\x+30)},\x);
\end{scope}
\draw[flow] (decSea.east) -- (sigout.west);
\node[above=2pt of sigout,font=\footnotesize] {$\hat{X}^{i}\in\mathbb{R}^{f\cdot T}$};

\end{tikzpicture}}
  \caption{\textbf{Tokenizer training.} Each signal $X^{i}$ is encoded with RVQ into discrete residual tokens $V^{i}$. Encoder and decoder are trained jointly to reconstruct $X^{i}$.}
  \label{fig:tokenizer}
\end{wrapfigure}
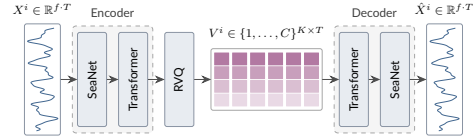
For each modality, we train tokenizers to transform the continuous raw signals into discrete tokens. We do this using an encoder-decoder architecture with a residual vector quantization (RVQ) layer, previously used by several foundation models, including for audio~\cite{defossezMoshiSpeechtextFoundation2024} and brain data~\cite{xiaoBrainOmniBrainFoundation2025}. Residual vector quantization~\cite{juangMultipleStageVector1982,martinezStackedQuantizersCompositional2014} extends vector quantization by representing each input unit (e.g.\ a signal segment) as a sum of $K$ tokens drawn from $K$ separate codebooks. The first token quantizes the input vector; each subsequent token quantizes the residual error left by the preceding partial reconstruction.

We tokenize each 1D channel $X^i\in\mathbb{R}^{f\cdot T}$ into a stream of discrete residual tokens $V^i\in\{1, \ldots, C\}^{K\times T}$, where $f$ is the sampling rate in Hz, $T$ is the sequence length in seconds, $K$ is the number of residual levels, and $C$ is the per-codebook vocabulary size. This differs from the design of BrainTokenizer~\cite{xiaoBrainOmniBrainFoundation2025}, which compresses EEG or MEG data with varying numbers of channels into a fixed number of `virtual' channels. This allows us to delegate the handling of missing modalities to downstream sequence modelling described in \Cref{section:Hypnos}.

\paragraph{Architecture}
Our encoders and decoders consist of stacks of convolutional (SeaNet~\cite{tagliasacchiSEANetMultimodalSpeech2020}) and Transformer layers~\cite{vaswaniAttentionAllYou2017}, as illustrated in \Cref{fig:tokenizer}. We configure the stride in the convolutional layers such that the encoders produce embeddings at 1\,Hz. All convolutions are causal~\cite{oordWaveNetGenerativeModel2016}, and each Transformer layer uses a causal sliding window. This design means that after training the tokenizer can be applied to arbitrary length input sequences, supporting streaming inference and enabling us to apply the tokenizers to an entire night of data in a single forward pass.

Because they have similar signal characteristics, we share tokenizers between EEG channels, between EOG channels, and between respiratory channels, resulting in five tokenizers for eight modalities. The number of quantizers $K$ was chosen for each modality to maintain high reconstruction accuracy. Full architecture and training hyper-parameters are given in \Cref{app:codec-hp}. An investigation into the effect of quantization depth is performed in \Cref{section:appendix:tokenizer_design}.

\paragraph{Optimisation}
Each tokenizer is trained on 64-second windows sampled from the training split of the pre-training datasets. We use a multi-term reconstruction loss combined with an RVQ commitment penalty closely following BrainTokenizer~\cite{xiaoBrainOmniBrainFoundation2025}. Full loss specifications and hyper-parameters are given in \Cref{app:codec-hp}.

\subsection{Hypnos}\label{section:Hypnos}
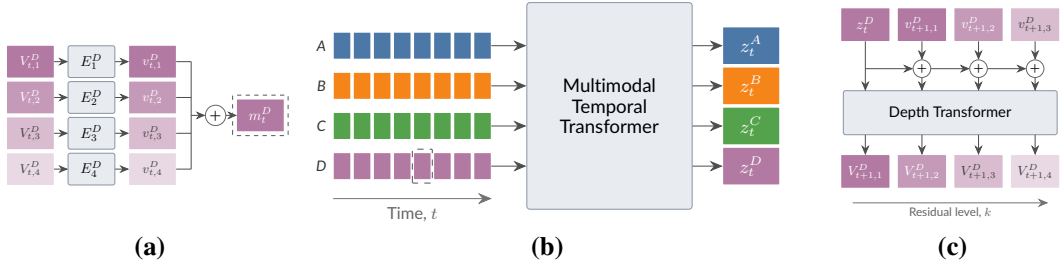
\begin{figure}
  \centering
  \def\panelht{3cm}
  \begin{minipage}[b]{0.27\linewidth}
    \centering
    \begin{minipage}[c][\panelht][c]{\linewidth}
      \centering
      \resizebox{\linewidth}{!}{\latofont
\begin{tikzpicture}[
    font=\footnotesize,
    >={Stealth[length=2.2mm,width=1.8mm]},
    tok/.style={draw=none,minimum width=11.5mm,minimum height=7.8mm,
                inner sep=0pt,rounded corners=0.4pt,font=\normalsize\color{white}},
    embbox/.style={draw=blockEdge,fill=blockFill,rounded corners=1.5pt,
                   minimum width=11.5mm,minimum height=7.8mm,
                   inner sep=1pt,font=\normalsize,align=center},
    sumnode/.style={draw=black!60,circle,inner sep=0pt,minimum size=5mm,
                    font=\small,fill=white},
    flow/.style={->,draw=black!70,line width=0.45pt},
    wire/.style={draw=black!70,line width=0.45pt},
]

\def\modcol{gD}
\def\K{4}
\def\dy{0.88}

\def\xV{0.00}
\def\xE{1.50}
\def\xv{3.00}
\def\xBus{3.95}
\def\xSum{4.55}
\def\xOut{5.65}

\foreach \k/\y/\shade/\txt in {%
  1/{1.5*\dy}/100/white,
  2/{0.5*\dy}/75/white,
  3/{-0.5*\dy}/50/{black!75},
  4/{-1.5*\dy}/28/{black!75}%
} {
  \node[tok,fill=\modcol!\shade,font=\small\color{\txt}]
    (V\k) at (\xV, {\y}) {$V^{D}_{t,\k}$};
}

\foreach \k/\y in {1/{1.5*\dy},2/{0.5*\dy},3/{-0.5*\dy},4/{-1.5*\dy}} {
  \node[embbox] (e\k) at (\xE, {\y}) {$E_{\k}^{D}$};
  \draw[flow] (V\k.east) -- (e\k.west);
}

\foreach \k/\y/\shade/\txt in {%
  1/{1.5*\dy}/100/white,
  2/{0.5*\dy}/75/white,
  3/{-0.5*\dy}/50/{black!75},
  4/{-1.5*\dy}/28/{black!75}%
} {
  \node[tok,fill=\modcol!\shade,font=\small\color{\txt}]
    (vv\k) at (\xv, {\y}) {$v^{D}_{t,\k}$};
  \draw[flow] (e\k.east) -- (vv\k.west);
}

\foreach \k in {1,...,\K} {
  \draw[wire] (vv\k.east) -- ({\xBus}, {(2.5 - \k)*\dy});
}
\draw[wire] ({\xBus}, {1.5*\dy}) -- ({\xBus}, {-1.5*\dy});

\node[sumnode] (sum) at (\xSum, 0) {\(\boldsymbol{+}\)};
\draw[flow] ({\xBus}, 0) -- (sum.west);

\node[tok,fill=\modcol] (out) at (\xOut, 0) {$m^{D}_{t}$};
\draw[flow] (sum.east) -- (out.west);
\draw[dashed,draw=black!65,line width=0.5pt,rounded corners=1pt]
  ($(out.south west)+(-0.10,-0.10)$) rectangle ($(out.north east)+(0.10,0.10)$);

\end{tikzpicture}}
    \end{minipage}\par
    \vspace{2pt}\textbf{(a)}
  \end{minipage}\hspace{0.01\linewidth}
  \begin{minipage}[b]{0.45\linewidth}
    \centering
    \begin{minipage}[c][\panelht][c]{\linewidth}
      \centering
      \resizebox{\linewidth}{!}{\latofont

\begin{tikzpicture}[
    font=\footnotesize,
    >={Stealth[length=2.2mm,width=1.8mm]},
    tok/.style={draw=none,minimum width=9.5mm,minimum height=6.2mm,
        inner sep=0pt,rounded corners=0.3pt,font=\small\color{white}},
    smalltok/.style={draw=none,minimum width=2.8mm,minimum height=4.4mm,
        inner sep=0pt,rounded corners=0.2pt},
    rowlbl/.style={font=\scriptsize\itshape,text=black!80},
    block/.style={draw=blockEdge,fill=blockFill,rounded corners=1.5pt,
        align=center,font=\small,inner sep=3pt},
    flow/.style={->,draw=black!70,line width=0.45pt},
    sublbl/.style={font=\footnotesize,text=black!60},
  ]

  \def\sx{0.34}   
  \def\dy{0.68}
  \def\Ntok{8}   
  \def\xL{0.95}   
  \def\xRlast{3.33}  
  \def\xRowLbl{0.60} 

  \foreach \r/\lab in {0/A,1/B,2/C,3/D} {
      \node[rowlbl] at (\xRowLbl, {-\dy*\r}) {\lab};
    }

  \foreach \r/\col in {0/gA,1/gB,2/gC,3/gD} {
      \foreach \t in {1,...,\Ntok} {
          \node[smalltok,fill=\col] (tok\r\t) at ({\xL + \sx*(\t - 1)}, {-\dy*\r}) {};
        }
    }

  \draw[dashed,draw=black!65,line width=0.5pt,rounded corners=0.5pt]
  ($(tok35.south west)+(-0.022,-0.06)$) rectangle
  ($(tok35.north east)+(0.022,0.06)$);

  \draw[->,draw=black!55,line width=0.3pt]
  ({\xL - 0.15}, {-\dy*3 - 0.55}) -- ({\xRlast + 0.15}, {-\dy*3 - 0.55});
  \node[sublbl] at ({(\xL + \xRlast)/2}, {-\dy*3 - 0.82}) {Time, $t$};

  \node[block,minimum width=2.8cm,minimum height=3.5cm]
  (temporal) at ({\xRlast + 2.15}, {-\dy*1.5})
  {Multimodal\\Temporal\\Transformer};

  \foreach \r in {0,...,3} {
      \draw[flow] (tok\r8.east) -- (temporal.west |- {0,-\dy*\r});
    }

  \foreach \r/\sup/\col in {%
      0/{\scriptstyle A}/gA,
      1/{\scriptstyle B}/gB,
      2/{\scriptstyle C}/gC,
      3/{\scriptstyle D}/gD%
    } {
      \node[tok,fill=\col] (ctx\r) at ({\xRlast + 4.55}, {-\dy*\r}) {$z^{\sup}_{t}$};
      \draw[flow] (temporal.east |- ctx\r) -- (ctx\r.west);
    }

\end{tikzpicture}}
    \end{minipage}\par
    \vspace{2pt}\textbf{(b)}
  \end{minipage}\hfill
  \begin{minipage}[b]{0.21\linewidth}
    \centering
    \begin{minipage}[c][\panelht][c]{\linewidth}
      \centering
      \resizebox{\linewidth}{!}{\latofont\begin{tikzpicture}[
    font=\footnotesize,
    >={Stealth[length=2.2mm,width=1.8mm]},
    tok/.style={draw=none,minimum width=11.5mm,minimum height=7.8mm,
                inner sep=0pt,rounded corners=0.4pt,font=\normalsize\color{white}},
    ztoki/.style={draw=none,fill=\modcol,minimum width=11.5mm,minimum height=7.8mm,
                  inner sep=0pt,rounded corners=0.4pt,font=\normalsize\color{white}},
    block/.style={draw=blockEdge,fill=blockFill,rounded corners=1.5pt,
                  align=center,font=\normalsize,inner sep=3pt},
    sumnode/.style={draw=black!60,circle,inner sep=0pt,minimum size=4.2mm,
                    font=\scriptsize,fill=white},
    flow/.style={->,draw=black!70,line width=0.45pt},
    sublbl/.style={font=\footnotesize,text=black!60},
]

\def\modcol{gD}
\def\K{4}
\def\colX#1{\ifcase#1\or 1.20\or 2.55\or 3.90\or 5.25\fi}

\def\topY{2.10}
\def\plusY{1.05}

\node[ztoki] (zti) at ({\colX{1}}, \topY) {$z^{D}_{t}$};

\foreach \k/\shade/\txt in {2/100/white,3/75/white,4/50/{black!75}} {
  \pgfmathtruncatemacro{\km}{\k - 1}
  \node[tok,fill=\modcol!\shade,font=\small\color{\txt}]
    (vin\k) at ({\colX{\k}}, \topY) {$v^{D}_{t+1,\km}$};
}

\foreach \k in {2,...,\K} {
  \node[sumnode] (plus\k) at ({\colX{\k}}, \plusY) {\(\boldsymbol{+}\)};
}

\foreach \k in {2,3,4} {
  \draw[flow] (vin\k.south) -- (plus\k.north);
}

\draw[flow] ({\colX{1}}, \plusY) -- (plus2.west);
\foreach \k in {3,...,\K} {
  \pgfmathtruncatemacro{\km}{\k - 1}
  \draw[flow] (plus\km.east) -- (plus\k.west);
}

\def\bxW{5.10}
\def\bxH{1.05}
\def\bxCX{{(\colX{1} + \colX{\K})/2}}
\def\bxCY{0}
\node[block,minimum width=\bxW cm,minimum height=\bxH cm]
  (depth) at (\bxCX, \bxCY) {Depth Transformer};

\draw[flow] (zti.south) -- ({\colX{1}}, {\bxCY + \bxH/2});
\foreach \k in {2,...,\K} {
  \draw[flow] (plus\k.south) -- ({\colX{\k}}, {\bxCY + \bxH/2});
}

\def\botY{-1.40}
\foreach \k/\shade/\txt in {1/100/white,2/75/white,3/50/{black!75},4/28/{black!75}} {
  \node[tok,fill=\modcol!\shade,font=\small\color{\txt}]
    (vout\k) at ({\colX{\k}}, \botY) {$V^{D}_{t+1,\k}$};
  \draw[flow] ({\colX{\k}}, {\bxCY - \bxH/2}) -- (vout\k.north);
}

\draw[->,draw=black!55,line width=0.3pt]
  ({\colX{1} - 0.30}, {\botY - 0.75}) -- ({\colX{\K} + 0.30}, {\botY - 0.75});
\node[sublbl] at ({(\colX{1} + \colX{\K})/2}, {\botY - 0.98}) {Residual level, $k$};

\end{tikzpicture}}
    \end{minipage}\par
    \vspace{2pt}\textbf{(c)}
  \end{minipage}
  \caption{\textbf{Hypnos training.} (a) For each time $t$, the $K$ discrete residual tokens (illustrated with $K=4$) from modality $i$ are combined to form an embedding $m^i_t$. (b) A Transformer backbone mixes information to produce embeddings $z^i_t$ for each modality, e.g. $i\in \{A, B, C, D\}$. (c) For all $i,t$, the Depth Transformer auto-regressively predicts the next residual token $V^i_{t+1,k}$ conditioned on $z^i_t$.}
  \label{fig:arch}
\end{figure}

After tokenization, we train a RQ-Transformer~\cite{leeAutoregressiveImageGeneration2022} to minimise the conditional log-likelihood over next residual tokens with all modalities, timesteps and residual levels weighted equally:
\begin{equation}\label{eq:hypnos-objective}
  \mathcal{L}(\theta)
  = -\sum_{i=1}^{M}\sum_{t=1}^{T}\sum_{k=1}^{K}
  \log p_\theta\!\left(V^i_{t,k}\mid V_{<t,1:K},\, V^i_{t,1:k-1}\right).
\end{equation}
Here $V_{<t,1:K}$ denotes the tokens at all residual levels and modalities before time $t$, restricted to the same attention group as modality $i$ during training (\Cref{sec:modality-masking}). A key advantage of the RQ-Transformer design is that, rather than flattening the $K\cdot T$ tokens per modality into a single sequence, the architecture decouples temporal modelling from residual-depth modelling, meaning self-attention scales with $T$ rather than $K\cdot T$. This is illustrated in \Cref{fig:arch}.

The first stage of the model is a learnt embedding layer which aggregates discrete residual tokens $V^i_{t,k}$ into a single embedding $m^i_t$ for each modality and timestep following prior work~\cite{leeAutoregressiveImageGeneration2022,defossezMoshiSpeechtextFoundation2024}. This is followed by a stack of Transformer layers~\cite{vaswaniAttentionAllYou2017} which aggregate information over time and across modalities from the sequences of embeddings $m^i_t$. To reduce computational complexity, each layer alternates between temporal and modality attention, rather than all-to-all attention. Finally, a depth transformer which auto-regressively predicts residual tokens for the next timestep conditioned on the modality embedding $z^i_t$. To allow the shared depth transformer to disambiguate modalities, a learnt per-modality embedding $e^i$ is also added to the conditioning. Predictions are made in parallel for all modalities $i$ and timesteps $t$. The backbone and depth transformer parameters are shared across all modalities.

After training, we use the output embeddings $z^i_t$ from the temporal transformer for downstream tasks. The RQ-Transformer design means that the outputs of the temporal transformer are a natural choice to use as embeddings for downstream tasks. The training process encourages these embeddings to simultaneously encode coarse to fine details required by the depth transformer to predict each output residual token.

\paragraph{Design and Optimisation}
We use modern Transformer components throughout, with sliding-window attention~\cite{beltagyLongformerLongDocumentTransformer2020} used to enable length-generalisation during inference. Models are trained with AdamW~\cite{loshchilovDecoupledWeightDecay2019} and a cosine learning rate schedule, following prior language and audio work~\cite{brownLanguageModelsAre2020,defossezMoshiSpeechtextFoundation2024}. By default, we use batch size $B=512$ and context length $T=512$, i.e.\ $\approx 8.5$ minutes of data per sequence at 1\,Hz. We evaluated three model scales (\Cref{tab:hypnos-configs}) informed by Vision Transformer configurations~\cite{dosovitskiyImageWorth16x162020}. Full architecture and training hyper-parameters are given in \Cref{app:hypnos-hp}.

\begin{table}[h]
  \centering
  \caption{\textbf{Hypnos model configurations}. Parameter counts include token embedding tables, both Transformer stacks, and per-codebook output heads.\label{tab:hypnos-configs}}
  \begin{tabular}{lccccccr}
    \toprule
                 & \multicolumn{3}{c}{Temporal Transformer} & \multicolumn{3}{c}{Depth Transformer} &                                                \\
    \cmidrule(lr){2-4} \cmidrule(lr){5-7}
    Model        & Layers                                   & Hidden \(D\)                          & Heads & Layers & Hidden \(D\) & Heads & Params \\
    \midrule
    Hypnos-Tiny  & 12                                       & 192                                   & 3     & 4      & 128          & 2     & 37M    \\
    Hypnos-Small & 12                                       & 384                                   & 6     & 4      & 192          & 3     & 81M    \\
    Hypnos-Base  & 12                                       & 768                                   & 12    & 4      & 384          & 6     & 222M   \\
    \bottomrule
  \end{tabular}
\end{table}

\subsection{Modality Masking}\label{sec:modality-masking}
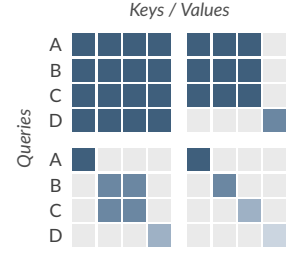
\begin{wrapfigure}{R}{0.35\linewidth}
  \centering
  \resizebox{0.75\linewidth}{!}{\latofont

\begin{tikzpicture}[
    font=\scriptsize,
    cell/.style={draw=none,minimum size=3.0mm,inner sep=0pt},
    lbl/.style={font=\scriptsize,text=black!75},
    axis/.style={font=\scriptsize\itshape,text=black!70},
]

\foreach \i/\n in {1/A,2/B,3/C,4/D} {
  \node[lbl] at (-0.38,-0.34*\i) {\n};
  \node[lbl] at (-0.38,-1.55-0.34*\i) {\n};
}

\node[axis,anchor=south] at (1.29,-0.12) {Keys / Values};
\node[axis,rotate=90,anchor=south] at (-0.56,-1.60) {Queries};

\foreach \X/\Y/\ga/\gb/\gc/\gd in {%
    0.0/0.0/0/0/0/0,
    1.55/0.0/0/0/0/1,
    0.0/-1.55/0/1/1/2,
    1.55/-1.55/0/1/2/3%
} {
  \foreach \ii/\gi in {1/\ga,2/\gb,3/\gc,4/\gd} {
    \foreach \jj/\gj in {0/\ga,1/\gb,2/\gc,3/\gd} {
      \ifnum\gi=\gj
        \ifcase\gi\relax
          \node[cell,fill=g1] at (\X + 0.34*\jj, \Y - 0.34*\ii) {};
        \or
          \node[cell,fill=g2] at (\X + 0.34*\jj, \Y - 0.34*\ii) {};
        \or
          \node[cell,fill=g3] at (\X + 0.34*\jj, \Y - 0.34*\ii) {};
        \or
          \node[cell,fill=g4] at (\X + 0.34*\jj, \Y - 0.34*\ii) {};
        \fi
      \else
        \node[cell,fill=maskCol] at (\X + 0.34*\jj, \Y - 0.34*\ii) {};
      \fi
    }
  }
}
\end{tikzpicture}%
}
  \caption{\textbf{Example cross-modal attention matrices ($M=4$).} During training, attention is restricted to random sub-groups.}
  \label{fig:modality-masking}
\end{wrapfigure}
A desired property of our model is robustness to missing modalities. Sleep studies use a wide range of sensor configurations, from full polysomnography (PSG) in clinical settings to single-channel EEG or cardio-respiratory only recordings in the home using wearable devices. To improve generalisation to subsets of supported modalities during inference, we randomly divide modalities into one or more groups of varying sizes during training. We then restrict attention in each Transformer layer so that modalities within a group can only attend to each other, as illustrated in \Cref{fig:modality-masking}.

We split modalities into a variable number of groups $G$ by sampling from a Chinese Restaurant Process~\cite{aldousExchangeabilityRelatedTopics1985} with concentration parameter $\alpha$. Prior work has improved robustness to missing modalities by randomly masking out sensing modalities during training, e.g.~\cite{carterWav2sleepUnifiedMultiModal2025,xuLSM2LearningIncomplete2025,shuaiOSFPretrainingScaling2026}.  We use the Chinese Restaurant Process as a convenient way to interpolate between using a single group ($\alpha\!\to\!0$, $G{=}1$) and fully independent ($\alpha\!\to\!\infty$, $G{=}M$), i.e. no cross-modal fusion during training. We use $\alpha=1$ by default, which exposes the model to a wide range of group sizes during training, and which gives $\mathbb{E}[G] \approx 2.72$ for $M{=}8$. We expected this to give better test-time performance than random dropout, since no tokens go unused during training, but found that both perform comparably (\Cref{sec:modality-masking-ablation}).

\subsection{Embedding Aggregation}\label{sec:agg}
We take a simple approach to produce embeddings over different timescales and modalities. To produce a summary vector $z_t$ for each timestep $t$, we simply take the average over the embeddings from available modalities $z^i_t$. To produce a single embedding for a time range $[t_1, t_2]$, we simply average embeddings over that time range. For example, to produce an embedding for a 30-second interval for sleep stage classification, we average embeddings $z_t$ over that interval. We leave the design of more effective embedding aggregation strategies to future work.

\subsection{Evaluation set-up}
In our main experiments, we compare Hypnos with OSF~\cite{shuaiOSFPretrainingScaling2026}, SleepFM~\cite{thapaMultimodalSleepFoundation2026}, and sleep2vec~\cite{yuanSleep2vecUnifiedCrossModal2026}. We re-train SleepFM using the open-source code and re-implement sleep2vec from the paper description. Both re-trained models use the same data splits, modalities and pre-processing as Hypnos. We directly use the public OSF model weights. Additional compute-matched comparisons with re-trained OSF, and a comparison with SleepMaMi~\cite{parkSleepMaMiUniversalSleep2026}, are given in \Cref{app:compute,app:sleepmami}. Further implementation details are given in \Cref{app:fm-comparison}.

To simplify comparison across foundation models, which produce embeddings at different timescales, we formulate sleep analysis tasks as classification over 30-second windows of data, following~\cite{shuaiOSFPretrainingScaling2026}. For example, a 30-second window is marked as `positive' for apnoea if it has any overlap with an apnoea event. Recordings are chunked to match the context length of each model. Using Hypnos, we can generate embeddings for an arbitrary length recording in a single forward pass.

For each model, we use all supported modalities as inputs. All probes are trained using only the training and validation splits of the pre-training datasets, such that MrOS, DOD-H and DOD-O serve as fully external test sets to evaluate generalisation. For our supervised baselines, we use the SLEEPYLAND framework~\cite{rossiSLEEPYLANDTrustBegins2025a} to train and evaluate SleepTransformer~\cite{phanSleepTransformerAutomaticSleep2022} and U-Sleep~\cite{perslevUSleepResilientHighfrequency2021} on the same dataset splits. Additional details of our evaluation set-up are described in \Cref{app:fm-comparison}.
\section{Experiments}\label{sec:experiments}

\subsection{Comparison with existing foundation models}
In \Cref{tab:robustness}, we compare Hypnos with existing foundation models across common sleep analysis tasks using four clinically motivated modality configurations: full PSG ($M=8$), single-channel EEG ($M=1$), EOG ($M=2$), and cardio-respiratory signals (ECG, ABD and THX; $M=3$). Additional information on the evaluation tasks is provided in \Cref{app:fm-comparison}. We report performance using mean AUROC and AUPRC values over recordings. To assess statistical significance, we perform a Wilcoxon signed-rank test over recordings with Benjamini-Hochberg FDR correction ($q<0.05$). The same procedure is used for all comparisons throughout this section. Across sensor configurations, Hypnos consistently achieves better performance across downstream tasks.

\begin{table}[h]
  \centering
  \small
  \setlength{\tabcolsep}{4pt}
  \caption{\textbf{Foundation model comparison across tasks and sensor configurations.} MLP probe results on held-out MrOS, under full-modality ($M{=}8$) and restricted-modality configurations. AUROC / AUPRC, mean over subjects; staging AUROC/AUPRC are macro-averaged over the five sleep stages. \textbf{Best} per metric in bold; $^{*}$ indicates Hypnos is significantly better.\label{tab:robustness}}
  \resizebox{\textwidth}{!}{%
    \begin{tabular}{ll cc cc cc cc}
      \toprule
                  &           & \multicolumn{2}{c}{Staging} & \multicolumn{2}{c}{Arousal} & \multicolumn{2}{c}{Apnoea} & \multicolumn{2}{c}{Desat.}                                                                                             \\
      \cmidrule(lr){3-4} \cmidrule(lr){5-6} \cmidrule(lr){7-8} \cmidrule(lr){9-10}
      Setting     & Method    & AUROC                       & AUPRC                       & AUROC                      & AUPRC                      & AUROC                & AUPRC                & AUROC                & AUPRC                \\
      \midrule
      Full        & SleepFM   & $0.952$                     & $0.718$                     & $0.883$                    & $0.698$                    & $0.777$              & $0.396$              & $0.731$              & $0.714$              \\
      ($M{=}8$)   & sleep2vec & $0.960$                     & $0.741$                     & $0.886$                    & $0.704$                    & $0.754$              & $0.382$              & $0.736$              & $0.712$              \\
                  & OSF       & $0.960$                     & $0.747$                     & $0.922$                    & $0.777$                    & $0.775$              & $0.392$              & $0.737$              & $0.715$              \\
                  & Hypnos    & $\mathbf{0.973}^{*}$        & $\mathbf{0.796}^{*}$        & $\mathbf{0.934}^{*}$       & $\mathbf{0.808}^{*}$       & $\mathbf{0.795}^{*}$ & $\mathbf{0.418}^{*}$ & $\mathbf{0.767}^{*}$ & $\mathbf{0.755}^{*}$ \\
      \midrule
      EEG-C3      & SleepFM   & $0.925$                     & $0.650$                     & $0.849$                    & $0.635$                    & $0.686$              & $0.305$              & $0.621$              & $0.609$              \\
      ($M{=}1$)   & sleep2vec & $0.946$                     & $0.707$                     & $0.834$                    & $0.610$                    & $0.693$              & $0.321$              & $0.636$              & $0.624$              \\
                  & OSF       & $0.937$                     & $0.674$                     & $0.906$                    & $0.749$                    & $0.659$              & $0.291$              & $\mathbf{0.643}$     & $0.620$              \\
                  & Hypnos    & $\mathbf{0.969}^{*}$        & $\mathbf{0.779}^{*}$        & $\mathbf{0.922}^{*}$       & $\mathbf{0.790}^{*}$       & $\mathbf{0.723}^{*}$ & $\mathbf{0.351}^{*}$ & $0.635$              & $\mathbf{0.628}^{*}$ \\
      \midrule
      EOG         & SleepFM   & $0.916$                     & $0.638$                     & $0.824$                    & $0.593$                    & $0.681$              & $0.303$              & $0.654$              & $0.627$              \\
      ($M{=}2$)   & sleep2vec & $0.933$                     & $0.671$                     & $0.804$                    & $0.545$                    & $0.667$              & $0.304$              & $0.653$              & $0.631$              \\
                  & OSF       & $0.940$                     & $0.691$                     & $0.907$                    & $0.747$                    & $0.688$              & $0.309$              & $0.651$              & $0.630$              \\
                  & Hypnos    & $\mathbf{0.962}^{*}$        & $\mathbf{0.752}^{*}$        & $\mathbf{0.912}^{*}$       & $\mathbf{0.766}^{*}$       & $\mathbf{0.717}^{*}$ & $\mathbf{0.335}^{*}$ & $\mathbf{0.666}^{*}$ & $\mathbf{0.645}^{*}$ \\
      \midrule
      ECG+ABD+THX & SleepFM   & $0.849$                     & $0.512$                     & $0.798$                    & $0.544$                    & $0.753$              & $0.373$              & $0.737$              & $0.714$              \\
      ($M{=}3$)   & sleep2vec & $0.909$                     & $0.618$                     & $0.830$                    & $0.607$                    & $0.748$              & $0.378$              & $0.737$              & $0.718$              \\
                  & OSF       & $0.870$                     & $0.545$                     & $0.849$                    & $0.632$                    & $0.762$              & $0.379$              & $0.742$              & $0.722$              \\
                  & Hypnos    & $\mathbf{0.931}^{*}$        & $\mathbf{0.659}^{*}$        & $\mathbf{0.878}^{*}$       & $\mathbf{0.684}^{*}$       & $\mathbf{0.797}^{*}$ & $\mathbf{0.417}^{*}$ & $\mathbf{0.795}^{*}$ & $\mathbf{0.777}^{*}$ \\
      \bottomrule
    \end{tabular}%
  }
\end{table}

\subsection{Comparison with supervised sleep stage classification models} In \Cref{tab:sleep-staging}, we compare the performance of Hypnos against strong supervised baselines on the task of sleep stage classification, evaluating against AASM (5-class) expert-annotated sleep stages: Wake, N1, N2, N3 and REM. We report Cohen's $\kappa$ (the most common metric in the automated sleep staging literature~\cite{phanSleepTransformerAutomaticSleep2022}), macro-averaged AUROC and macro-averaged AUPRC, for one representative in-domain cohort (SHHS) and one held-out cohort (MrOS). Additional per-dataset results across eight cohorts (SHHS, CCSHS, CFS, NCHSDB, MrOS, DOD-H, DOD-O, MESA), including MLP probe results, are reported in the 100\% column of \Cref{tab:fewshot-all,tab:fewshot-all-mlp}. Using only a linear probe, Hypnos outperforms strong supervised baselines on every metric on both SHHS and MrOS, and (as detailed in \Cref{app:fewshot-all}) outperforms both baseline models on the majority of datasets and metrics.
\begin{table}[h]
  \centering
  \small
  \caption{\textbf{Sleep stage classification performance.} Cohen's $\kappa$, macro-averaged AUROC and macro-averaged AUPRC (per-subject mean) on one in-domain (SHHS) and one held-out (MrOS) cohort. \textbf{Best} per (cohort, metric) in bold; $^{*}$ indicates Hypnos is significantly better than every other foundation model (paired Wilcoxon, one-sided, per-subject; Benjamini--Hochberg FDR-corrected $q<0.05$).\label{tab:sleep-staging}}

  \begin{tabular}{l ccc ccc}
    \toprule
                                                                                  & \multicolumn{3}{c}{SHHS (in-domain)} & \multicolumn{3}{c}{MrOS (held-out)}                                                                                             \\
    \cmidrule(lr){2-4} \cmidrule(lr){5-7}
    Method                                                                        & $\kappa$                             & AUROC                               & AUPRC                & $\kappa$             & AUROC                & AUPRC                \\
    \midrule
    \multicolumn{7}{l}{\textit{Supervised}}                                                                                                                                                                                                                \\
    \hspace*{0.5em}U-Sleep~\cite{perslevUSleepResilientHighfrequency2021}         & $0.799$                              & $0.969$                             & $0.825$              & $0.730$              & $0.949$              & $0.748$              \\
    \hspace*{0.5em}SleepTransformer~\cite{phanSleepTransformerAutomaticSleep2022} & $0.806$                              & $0.974$                             & $0.832$              & $0.753$              & $0.958$              & $0.758$              \\
    \midrule
    \multicolumn{7}{l}{\textit{Self-supervised + linear probe}}                                                                                                                                                                                            \\
    \hspace*{0.5em}SleepFM~\cite{thapaMultimodalSleepFoundation2026}              & $0.716$                              & $0.941$                             & $0.731$              & $0.658$              & $0.919$              & $0.659$              \\
    \hspace*{0.5em}sleep2vec~\cite{yuanSleep2vecUnifiedCrossModal2026}            & $0.763$                              & $0.964$                             & $0.792$              & $0.751$              & $0.958$              & $0.737$              \\
    \hspace*{0.5em}OSF~\cite{shuaiOSFPretrainingScaling2026}                      & $0.791$                              & $0.968$                             & $0.809$              & $0.740$              & $0.957$              & $0.738$              \\
    \hspace*{0.5em}Hypnos                                                         & $\mathbf{0.811}^{*}$                 & $\mathbf{0.976}^{*}$                & $\mathbf{0.836}^{*}$ & $\mathbf{0.799}^{*}$ & $\mathbf{0.973}^{*}$ & $\mathbf{0.794}^{*}$ \\
    \bottomrule
  \end{tabular}%
\end{table}

\subsection{Few-shot learning}
\begin{figure}[h]
  \centering
  \begin{subfigure}[t]{0.45\linewidth}
    \centering
    \includegraphics[width=\linewidth]{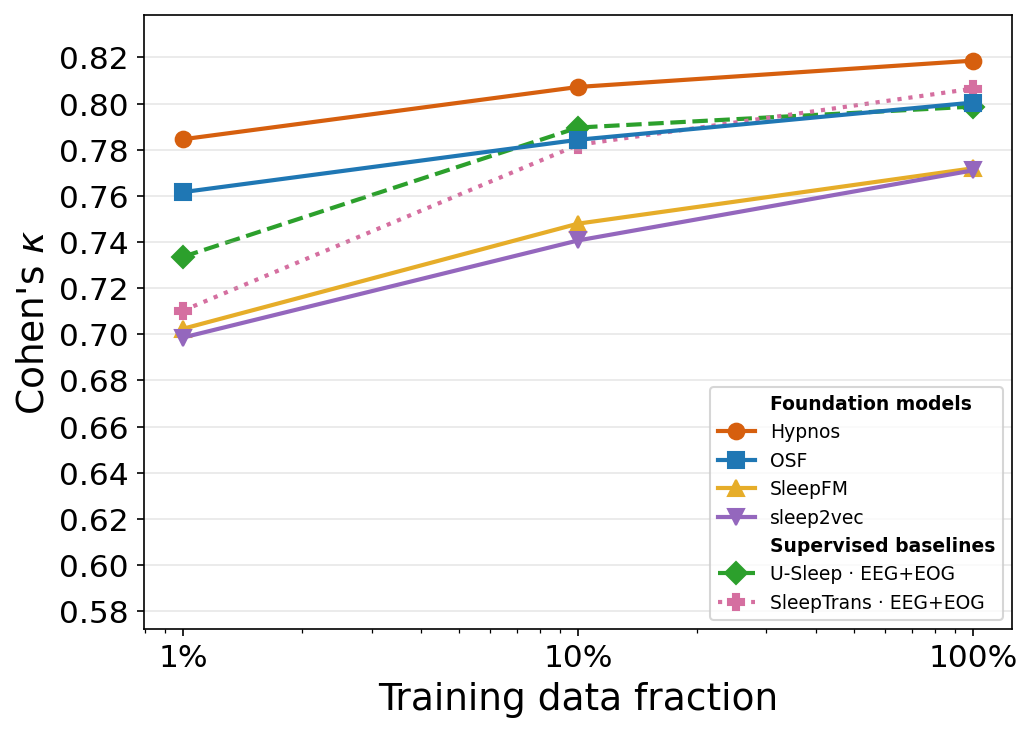}
    \caption{SHHS (in-domain)}\label{fig:fewshot-shhs}
  \end{subfigure}\hfill
  \begin{subfigure}[t]{0.45\linewidth}
    \centering
    \includegraphics[width=\linewidth]{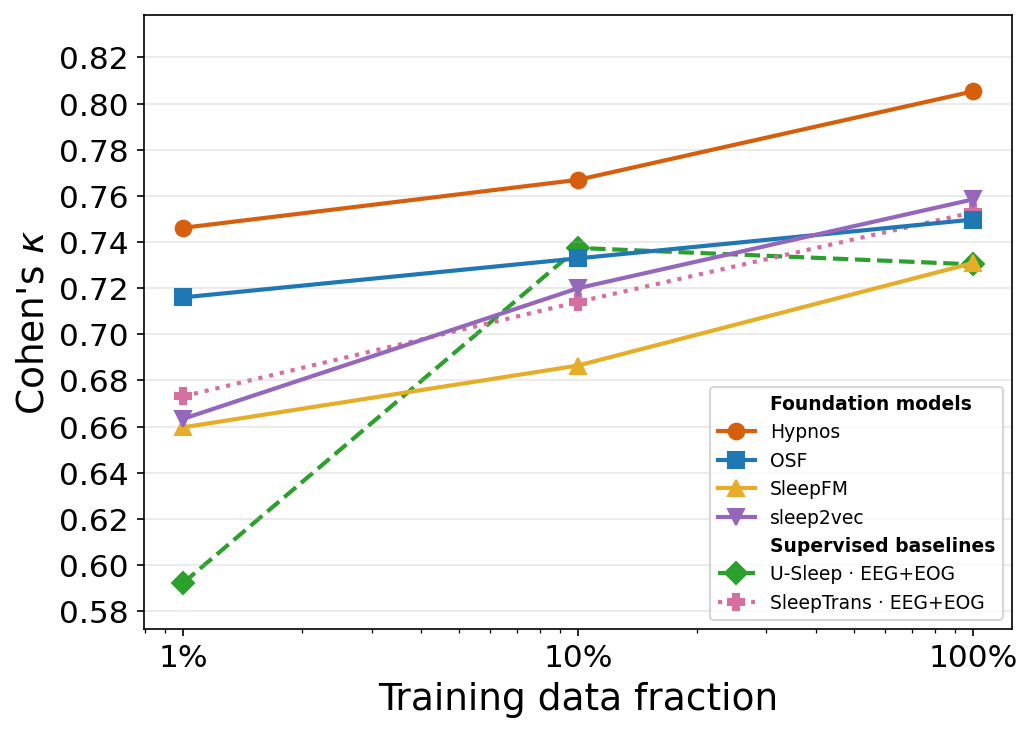}
    \caption{MrOS (held-out)}\label{fig:fewshot-mros}
  \end{subfigure}
  \caption{\textbf{Few-shot sleep stage classification.} We train MLP probes on each foundation model and re-train supervised baselines using varying fractions of in-domain data. Using as little as 1\% of the probe-training data, Hypnos matches U-Sleep trained on the full dataset on held-out MrOS.\label{fig:fewshot-staging}}
\end{figure}

Next, we evaluate model performance in the few-shot learning regime, scaling the proportion of labelled recordings used for training and validation from 1\% up to 100\%. \Cref{fig:fewshot-staging} reports sleep staging performance for an in-domain (SHHS) and a held-out (MrOS) dataset, alongside supervised baselines re-trained on the same data fractions. Hypnos outperforms existing foundation models and supervised baselines across all proportions of labelled data; on MrOS, Hypnos trained on as little as 1\% of recordings matches U-Sleep trained on the full dataset.

\subsection{Transfer to external ECG benchmarks}\label{sec:external-ecg}
\begin{wraptable}{r}{0.45\textwidth}
  \vspace{-1.2em}
  \centering
  \small
  \setlength{\tabcolsep}{4pt}
  \caption{\textbf{External ECG benchmarks.} Frozen-encoder linear-probe AUROC; full details in \Cref{app:external-ecg}.\label{tab:external-ecg}}
  \begin{tabular}{l c c}
    \toprule
    Dataset   & Hypnos           & xECG             \\
    \midrule
    CinC 2017 & $\mathbf{0.984}$ & $\mathbf{0.985}$ \\
    Apnea-ECG & $\mathbf{0.925}$ & $0.884$          \\
    CPSC 2021 & $\mathbf{0.985}$ & $0.934$          \\
    \bottomrule
  \end{tabular}
  \vspace{-0.8em}
\end{wraptable}
To assess whether Hypnos's representations transfer beyond PSG recordings, we also evaluated performance on three external single-lead ECG benchmarks: PhysioNet/CinC 2017 (atrial fibrillation detection), Apnea-ECG (overnight per-minute apnoea detection) and CPSC 2021 (paroxysmal AF detection). We compare against xECG~\cite{lunelliBenchECGXECGBenchmark2025}, a foundation model pre-trained on 12-lead clinical ECG. In \Cref{tab:external-ecg}, we see that Hypnos matches xECG on atrial fibrillation (AF) detection using CinC 2017, and beats it by 4 and 5 percentage points on Apnea-ECG and CPSC 2021 respectively, indicating the model effectively generalises beyond PSG recordings.

\subsection{Generative modelling}
\begin{figure}[h]
  \centering
  \includegraphics[width=\linewidth]{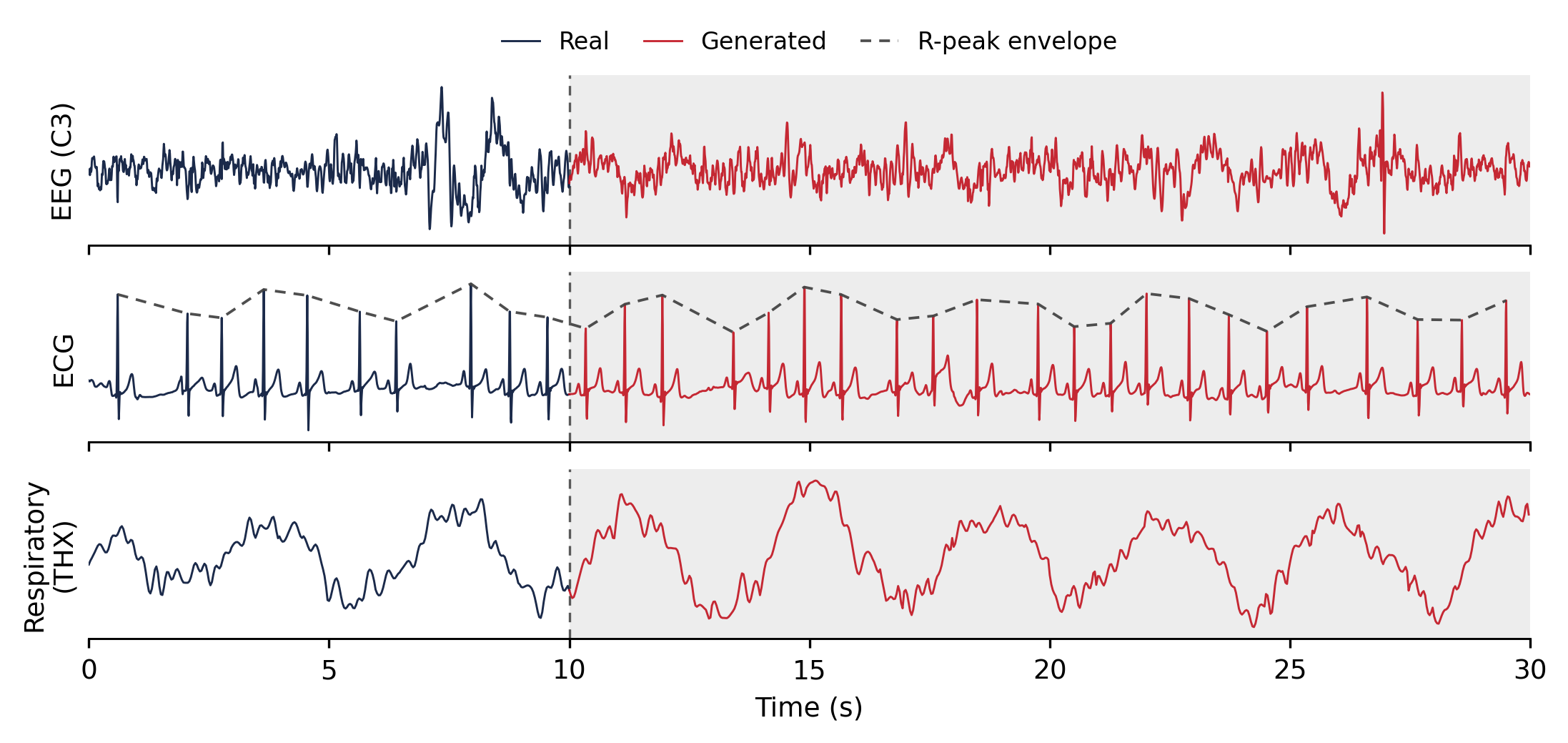}
  \caption{\textbf{Autoregressive generation of physiological signals.} Hypnos can be used to jointly generate physiological signals for any subset of supported modalities. Here we see that conditioned on 10\,s of real context (blue), Hypnos generates plausible signals with cross-modal consistency. For example, we can observe respiration-induced amplitude modulation of R-peaks in the ECG.\label{fig:generation}}
\end{figure}

Although the focus of our work is representation learning, the next-token prediction objective makes Hypnos fully generative: for any subset of supported modalities, tokens can be auto-regressively sampled and decoded back to waveforms via the tokenizers. \Cref{fig:generation} shows that, conditioned on real multi-modal context, Hypnos produces coherent continuations that preserve waveform morphology and cross-modal structure, indicating that next-token prediction captures the joint distribution of the underlying signals. Additional generated ECG examples during atrial fibrillation and regular rhythm are shown in \Cref{fig:ecg-rhythm} in the Appendix.

\subsection{Modality-masking}\label{sec:modality-masking-ablation}
We ablate our group masking approach (\textit{Default}, $\alpha=1$) against three alternatives: \textit{No masking} ($\alpha\!\to\!0$), where modalities are processed with full cross-modal attention during training; \textit{Independent} ($\alpha\!\to\!\infty$), where each modality is processed with no cross-modal attention; and \textit{Random} where modalities are randomly masked out with probability $p=0.5$. \Cref{tab:modality-masking-ablation} reports linear-probe performance on full ($M{=}8$) and restricted-modality (\textit{Restr.}) inputs, with full results in \Cref{tab:modality-masking-ablation-app}. The \textit{Independent} variant lags across every configuration, highlighting the benefit of cross-modal fusion. Training-time masking improves restricted-modality robustness over \textit{No masking} whilst maintaining full-modality performance. Surprisingly, both masking strategies perform comparably, despite a roughly $20\times$ difference in exposure to single-channel sequences during training, indicating that missing-modality robustness is insensitive to the specific masking strategy.

\begin{table}[h]
  \centering
  \small
  \setlength{\tabcolsep}{4pt}
  \caption{\textbf{Modality-masking ablation.} Per-subject mean of the primary metric per task. \textit{Restr.} is the mean across three restricted-modality configurations (EEG-C3, EOG, ECG+ABD+THX); \textit{Full} uses all eight modalities. Full results are reported in \Cref{tab:modality-masking-ablation-app}. \label{tab:modality-masking-ablation}}
  \begin{tabular}{l cc cc cc cc}
    \toprule
                                             & \multicolumn{2}{c}{Staging ($\kappa$)} & \multicolumn{2}{c}{Arousal (AUROC)} & \multicolumn{2}{c}{Apnoea (AUROC)} & \multicolumn{2}{c}{Desat. (AUROC)}                                         \\
    \cmidrule(lr){2-3} \cmidrule(lr){4-5} \cmidrule(lr){6-7} \cmidrule(lr){8-9}
    Variant                                  & Full                                   & Restr.                              & Full                               & Restr.                             & Full    & Restr.  & Full    & Restr.  \\
    \midrule
    \multicolumn{9}{l}{\textit{No training-time masking}}                                                                                                                                                                                     \\
    Independent ($\alpha\rightarrow \infty$) & $0.773$                                & $0.656$                             & $0.866$                            & $0.838$                            & $0.847$ & $0.798$ & $0.817$ & $0.776$ \\
    No masking ($\alpha\rightarrow 0$)       & $0.792$                                & $0.659$                             & $0.900$                            & $0.864$                            & $0.861$ & $0.802$ & $0.830$ & $0.769$ \\
    \cmidrule(lr){1-9}
    \multicolumn{9}{l}{\textit{Training-time modality masking}}                                                                                                                                                                               \\
    Random ($p=0.5$)                         & $0.795$                                & $0.672$                             & $0.902$                            & $0.873$                            & $0.860$ & $0.813$ & $0.831$ & $0.796$ \\
    Default ($\alpha=1$)                     & $0.797$                                & $0.676$                             & $0.901$                            & $0.874$                            & $0.859$ & $0.814$ & $0.831$ & $0.798$ \\
    \bottomrule
  \end{tabular}
\end{table}

\subsection{Model scaling}
\begin{figure}[h]
  \centering
  \begin{subfigure}[t]{0.24\linewidth}
    \centering
    \includegraphics[width=\linewidth]{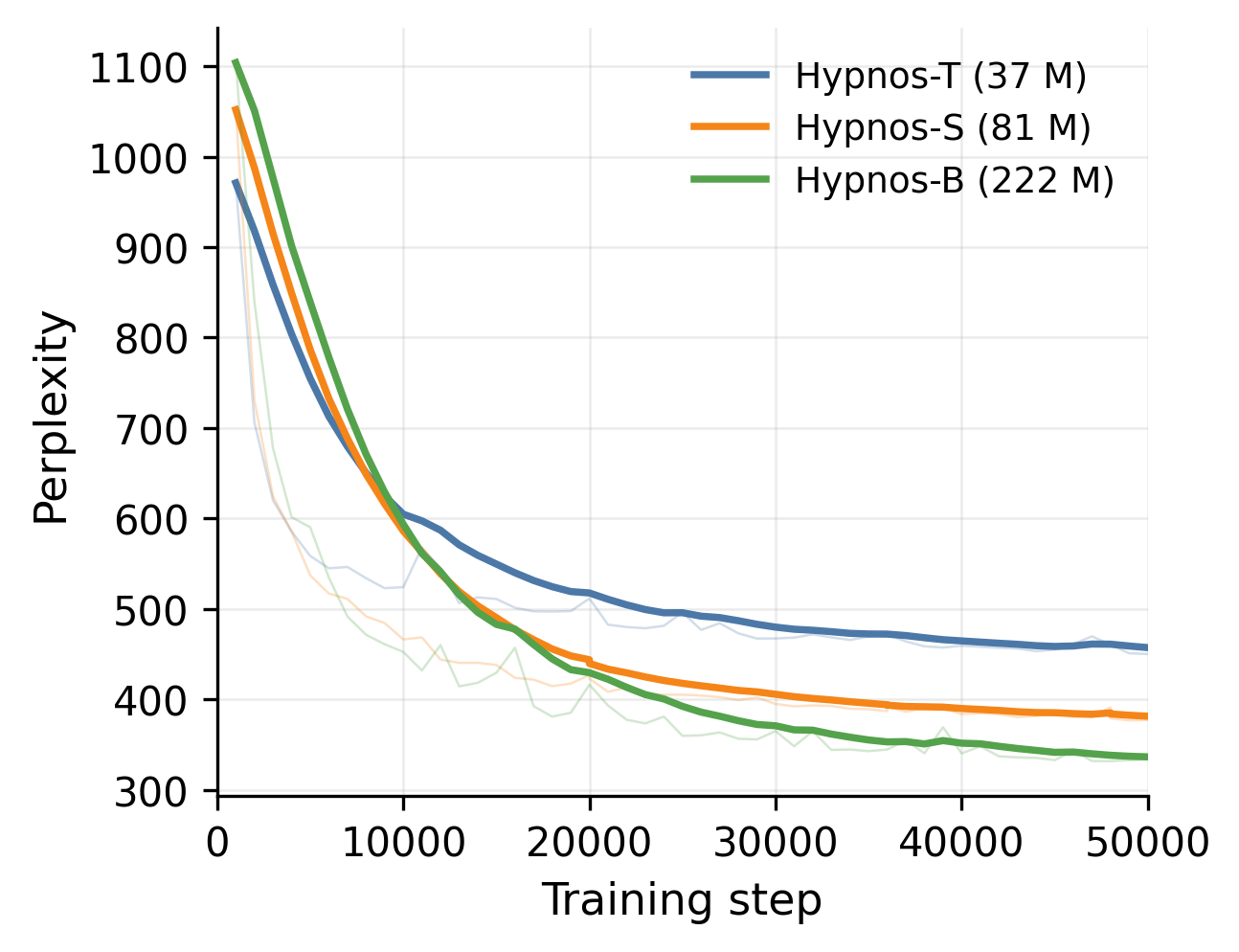}
    \caption{Validation loss}\label{fig:scaling-loss}
  \end{subfigure}\hfill
  \begin{subfigure}[t]{0.24\linewidth}
    \centering
    \includegraphics[width=\linewidth]{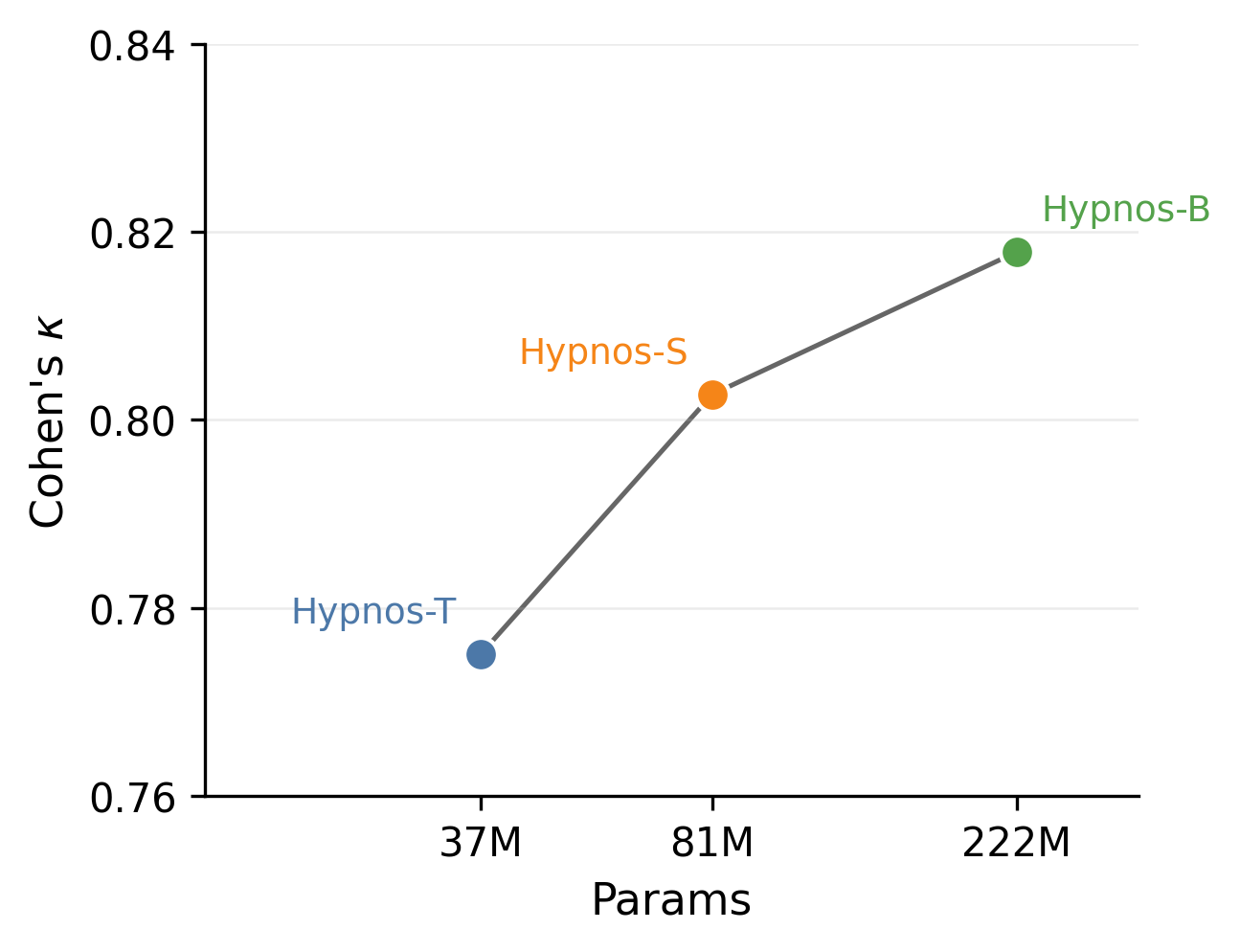}
    \caption{Sleep staging}\label{fig:scaling-staging}
  \end{subfigure}\hfill
  \begin{subfigure}[t]{0.24\linewidth}
    \centering
    \includegraphics[width=\linewidth]{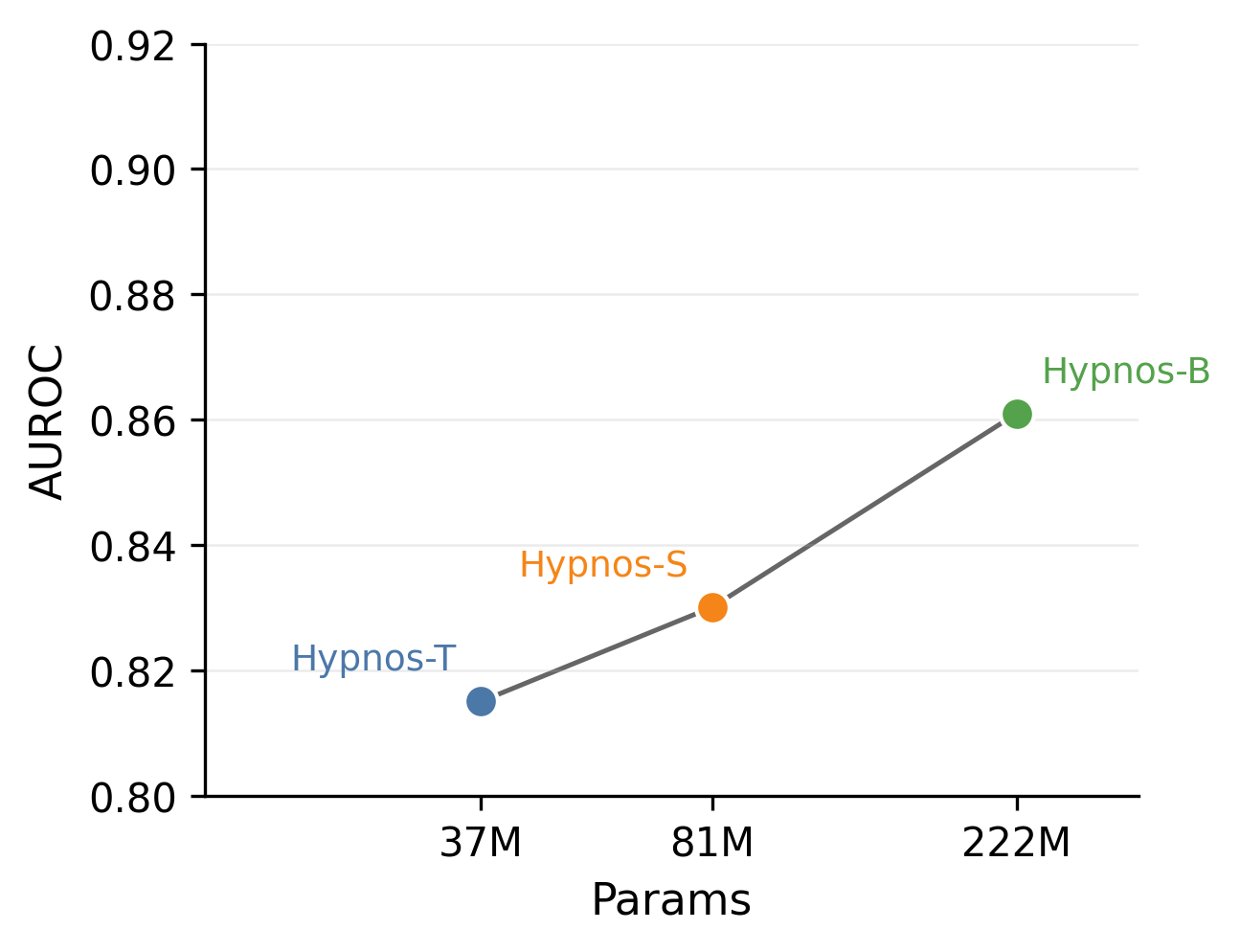}
    \caption{Apnoea detection}\label{fig:scaling-apnea}
  \end{subfigure}\hfill
  \begin{subfigure}[t]{0.24\linewidth}
    \centering
    \includegraphics[width=\linewidth]{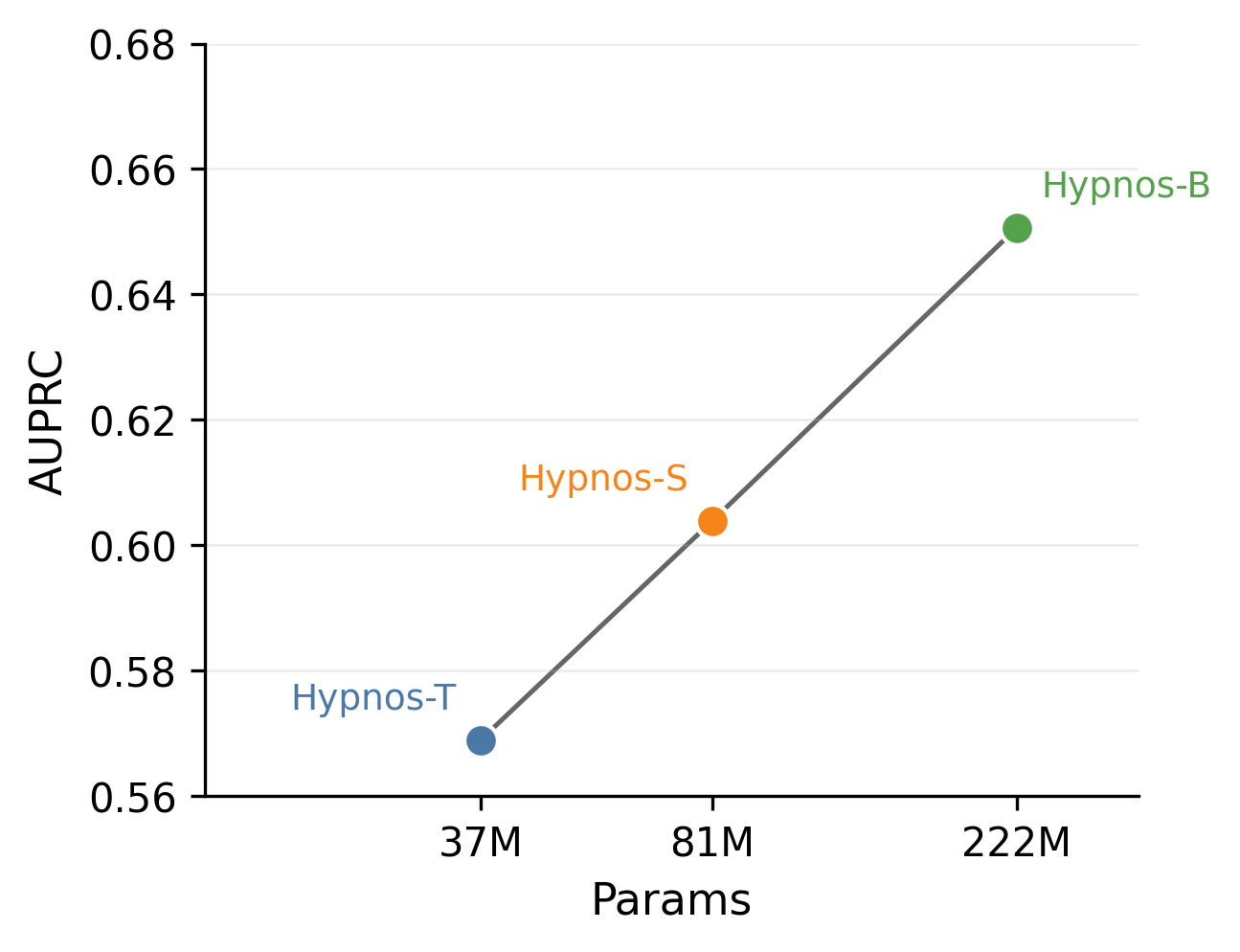}
    \caption{Arousal detection}\label{fig:scaling-arousal}
  \end{subfigure}
  \caption{\textbf{Scaling model size improves the performance of Hypnos.} Next-token perplexity and downstream metrics all improve with model scale. Sleep stage classification, apnoea detection and arousal detection performance are reported using a linear probe on the SHHS validation set.\label{fig:scaling-multimodal}}
\end{figure}

We investigated the effect of model scaling from Tiny to Base configurations. Further scaling experiments, including to larger model sizes and context lengths of over an hour ($T=4096$) using unimodal model variants, can be found in \Cref{app:scaling-eeg}. In \Cref{fig:scaling-multimodal}, we observe monotonic improvements in both next-token perplexity and performance on downstream tasks as we increase model size.

\section{Limitations and Future Work}

  \paragraph{Latent-space objectives} Our results suggest that a key ingredient may be prediction in a compressed latent space rather than in input space: masked-token prediction over the same tokens performs similarly to next-token prediction, while input-space MAE is weaker (\Cref{app:objective}). We expect other latent-space approaches, such as JEPA~\cite{assranSelfSupervisedLearningImages2023} and latent diffusion~\cite{rouardContinuousAudioLanguage2026}, to also work well, and comparing them at scale is an important direction for future work.

\paragraph{Long-context learning} Many clinically meaningful properties of physiological signals are only visible across hours or days of sensor data, including circadian phase, multi-night sleep regularity, and the clustering of rare events such as nocturnal seizures or periodic limb movements. Recent work has begun to reach these timescales by first compressing short segments into embeddings, then modelling sequences of those embeddings over a full night~\cite{parkSleepMaMiUniversalSleep2026}, a full day~\cite{wangLearningHumanHealth2026}, or multiple days~\cite{caoWavesFMHierarchicalRepresentation2026}. Combining this kind of hierarchical compression with next-token prediction could extend Hypnos to days of continuous multi-modal data while retaining its generative and streaming capabilities.

Further discussion of limitations and future directions, including clinical outcomes and biomarker discovery, is provided in \Cref{app:limitations}.

\section{Conclusions}
We have presented Hypnos, a multi-modal sleep foundation model trained with next-token prediction over residual vector quantised (RVQ) tokens drawn from eight physiological sensing modalities. Hypnos can be applied to real-time continuous streams of physiological sensor data from modalities such as EEG and ECG signals, generating high-quality embeddings for a range of downstream tasks. Our model outperforms existing foundation models and strong supervised baselines on most tasks and datasets evaluated, across diverse physiological sensing tasks such as sleep staging, atrial fibrillation and apnoea detection. Hypnos is one step toward a broader goal: foundation models that compress hours to days of multi-modal physiological signals into better measures of human health. That this simple recipe scales with model size, transfers to downstream tasks, generalises to held-out datasets, and is robust to sensor configurations suggests it can extend beyond sleep.

\section*{Acknowledgements}
Funding from ARIA and DSIT and Pillar VC under the Encode: AI for Science Fellowship. JC thanks Will Bolton, Botos Csaba and Bryan Li for their feedback on drafts of the paper. We kindly thank the National Sleep Research Resource (NSRR) for providing access to the datasets used. The National Sleep Research Resource was supported by the National Heart, Lung, and Blood Institute (R24 HL114473, 75N92019R002). Additional acknowledgements for the datasets used in this work are listed in \Cref{section:appendix:acks}.
\newpage
\section*{References}
{
  \small
  \bibliographystyle{plainnat}
  \renewcommand{\bibsection}{}
  \bibliography{references,references_extra}
}

\appendix

\section{Additional Implementation Details}\label{section:additional_implementation}

\subsection{Preprocessing}\label{app:preproc}

\paragraph{Referencing and filtering}
EEG and EOG channels were re-referenced against the contralateral mastoid (C3--M2, C4--M1 for EEG; E1--M2, E2--M1 for EOG). Chin EMG was derived bipolarly from the chin electrode pair. ECG and respiratory effort (ABD, THX) were used directly. All signals were resampled using a polyphase filter with an anti-aliasing low-pass. Brain, muscle, and cardiac channels were resampled to 128\,Hz. Respiratory effort signals (ABD, THX) were resampled to 32\,Hz, reflecting their lower frequency content. All channels were notch-filtered to suppress mains interference. Per-modality bandpass filters were then applied:
\begin{itemize}
  \item EEG, EOG, EMG: 0.5--45\,Hz bandpass.
  \item ECG: 0.05--45\,Hz bandpass.
  \item ABD, THX: 0.05\,Hz high-pass.
\end{itemize}

\paragraph{Normalisation and amplitude compression}
For each channel of each recording we performed an online rolling z-score normalisation: the mean and variance were tracked online by a 60-second exponential moving average. To prevent transient artefacts from inflating the running variance, the per-sample squared deviation was clipped at $6\sigma$ of the current variance estimate before accumulation. After normalisation, values lying beyond $\pm 8\sigma$ were log-compressed, so that large artefacts remain order-preserving without dominating the reconstruction loss.

\subsection{Tokenizer training details}\label{app:codec-hp}

\paragraph{Architecture}
Each tokenizer consists of a SeaNet and Transformer components plus an RVQ bottleneck, as illustrated in \Cref{fig:tokenizer}. SeaNet stride ratios are chosen so that the product of strides corresponds to one second of input samples, producing tokens at 1\,Hz at every modality's sampling rate. The encoder right-pads by one hop length and drops the first warm-up token, so each output token is right-aligned to its segment boundary. All other architecture and optimisation hyperparameters are listed in \Cref{tab:codec-hp}; weight decay is applied to Transformer parameters only, following the Mimi convention~\cite{defossezMoshiSpeechtextFoundation2024}.

\begin{table}[h]
  \centering
  \caption{\textbf{Tokenizer hyperparameters.}\label{tab:codec-hp}}
  \begin{tabular}{ll}
    \toprule
    Component                   & Setting                                               \\
    \midrule
    SeaNet \texttt{n\_filters}  & 64                                                    \\
    SeaNet stride ratios        & $[2,4,4,4]$ at 128\,Hz; $[4,4,2]$ at 32\,Hz           \\
    SeaNet residual blocks      & 1 per stage, dilation base 2                          \\
    Encoder embedding dim       & 512                                                   \\
    Codebook entry dim          & 256                                                   \\
    Codebook size $C$           & 2048                                                  \\
    Residual levels $K$         & 8 (EEG, EOG, EMG); 4 (ECG, respiratory)               \\
    EMA decay                   & 0.99                                                  \\
    Quantization dropout        & 0.5                                                   \\
    Encoder/decoder Transformer & 4 layers, 8 heads, FFN dim 2048, sliding window 32    \\
    LayerScale init             & $10^{-2}$                                             \\
    Loss weights                & $\lambda_\phi=0.5$, $\lambda_{\text{RVQ}}=0.25$       \\
    Optimiser                   & AdamW, lr $2\times 10^{-4}$, wd $10^{-2}$ (Tx params) \\
    Gradient clipping           & 1.0 (norm)                                            \\
    Schedule                    & Cosine, 500-step linear warm-up                       \\
    Training steps              & 50{,}000                                              \\
    Batch size $\times$ window  & $1024 \times 64$\,s                                   \\
    Precision                   & bf16 mixed                                            \\
    \bottomrule
  \end{tabular}
\end{table}

\paragraph{Loss formulation}
Following BrainTokenizer~\cite{xiaoBrainOmniBrainFoundation2025}, the tokenizer is trained with a multi-term reconstruction loss combined with an RVQ commitment penalty. The reconstruction loss has (i) a time-domain term between the original waveform $X$ and the reconstruction $\hat{X}$:
\begin{equation}
  \mathcal{L}_{\text{time}} \;=\; \|X - \hat{X}\|_1,
\end{equation}
where $\|\cdot\|_1$ denotes the $\ell_1$ distance; and (ii) a frequency-domain term between the amplitude spectra $A$, $\hat{A}$ and phase spectra $\Phi$, $\hat{\Phi}$ of the Hamming-windowed signals:
\begin{equation}
  \mathcal{L}_{\text{freq}} \;=\; \|A - \hat{A}\|_1 + \lambda_\phi \cdot \|\Phi - \hat{\Phi}\|_1.
\end{equation}
Letting $z_k$ and $z_{q_k}$ denote the residual entering the $k$-th codebook and its nearest entry, the RVQ commitment loss is:
\begin{equation}
  \mathcal{L}_{\text{rvq}} \;=\; \sum_{k=1}^{K}\big\|z_k - \mathrm{sg}[z_{q_k}]\big\|_2^{2},
\end{equation}
where $\mathrm{sg}[\cdot]$ denotes a stop-gradient. This term pulls the encoder output $z_k$ toward its assigned codebook entry. Each tokenizer is trained to minimise:
\begin{equation}
  \mathcal{L}_{\text{tok}} \;=\; \mathcal{L}_{\text{time}} + \mathcal{L}_{\text{freq}} + \lambda_{RVQ} \cdot \mathcal{L}_{\text{rvq}},
\end{equation}
with scalar weights $\lambda_\phi=0.5$ and $\lambda_{RVQ}=0.25$, identical to BrainTokenizer~\cite{xiaoBrainOmniBrainFoundation2025}.

\subsection{Hypnos hyperparameters and training}\label{app:hypnos-hp}
The temporal Transformer applies rotary position embeddings (RoPE,~\cite{suRoFormerEnhancedTransformer2024}) to attention queries and keys, while the depth Transformer uses a learnt position embedding indexed by codebook level. Linear and embedding parameters are initialised from $\mathcal{N}(0, 0.02^2)$ following GPT-2~\cite{brownLanguageModelsAre2020}. We use a residual dropout of $0.1$. Per-scale learning rates are set inversely to hidden width: $1.2 \times 10^{-3}$ for Hypnos-Tiny, $6 \times 10^{-4}$ for Hypnos-Small, and $3 \times 10^{-4}$ for Hypnos-Base. All other training hyperparameters are listed in \Cref{tab:hypnos-hp}.

\begin{table}[h]
  \centering
  \caption{\textbf{Hypnos training hyperparameters.}\label{tab:hypnos-hp}}
  \begin{tabular}{ll}
    \toprule
    Component              & Setting                                                                   \\
    \midrule
    Norm / activation      & LayerNorm + SwiGLU; RMSNorm on QK                                         \\
    Position encoding      & RoPE (temporal); learnt level-index embedding (depth)                     \\
    Sliding-window pattern & Local window 64; every 4th layer global ($T/2$)                           \\
    CRP concentration      & $\alpha = 1.0$                                                            \\
    Dropout                & 0.1                                                                       \\
    LayerScale init        & $10^{-2}$                                                                 \\
    Weight init            & $\mathcal{N}(0, 0.02^2)$ for Linear / Embedding                           \\
    Optimiser              & AdamW, $(\beta_1, \beta_2)=(0.9, 0.95)$, wd $0.1$                         \\
    Gradient clipping      & 1.0 (norm)                                                                \\
    Per-scale lr           & Tiny $1.2\times 10^{-3}$; Small $6\times 10^{-4}$; Base $3\times 10^{-4}$ \\
    Schedule               & Cosine, 500-step linear warm-up                                           \\
    Training steps         & 50{,}000                                                                  \\
    Batch size             & 512                                                                       \\
    Context length         & 512                                                                       \\
    Precision              & bf16 mixed, activation checkpointing                                      \\
    \bottomrule
  \end{tabular}
\end{table}

\subsection{Comparing sleep foundation models}\label{app:fm-comparison}

\paragraph{Baseline implementation} SleepFM~\cite{thapaMultimodalSleepFoundation2026} and sleep2vec~\cite{yuanSleep2vecUnifiedCrossModal2026} were re-implemented and re-trained using the same modalities, pre-processing steps, and dataset splits as Hypnos, which match the dataset splits used by OSF. We used the `Large' configuration of sleep2vec, which has around 240 million parameters, including a slightly larger Transformer backbone than Hypnos-Base. For completeness, in \Cref{tab:sleepfm-ots} we additionally report performance using the open-source SleepFM checkpoint, evaluated using all supported modalities and the original pre-processing steps. We adopt our re-trained variant as the stronger baseline throughout.

\begin{table}[h]
  \centering
  \small
  \caption{\textbf{Off-the-shelf vs.\ re-trained SleepFM.} Performance on held-out MrOS with an MLP probe.}\label{tab:sleepfm-ots}
  \begin{tabular}{l cc cc cc cc}
    \toprule
                 & \multicolumn{2}{c}{Staging} & \multicolumn{2}{c}{Arousal} & \multicolumn{2}{c}{Apnoea} & \multicolumn{2}{c}{Desat.}                                                                             \\
    \cmidrule(lr){2-3} \cmidrule(lr){4-5} \cmidrule(lr){6-7} \cmidrule(lr){8-9}
    SleepFM      & AUROC                       & AUPRC                       & AUROC                      & AUPRC                      & AUROC            & AUPRC            & AUROC            & AUPRC            \\
    \midrule
    Open weights & $0.928$                     & $0.657$                     & $0.771$                    & $0.465$                    & $0.729$          & $0.349$          & $0.694$          & $0.679$          \\
    Re-trained   & $\mathbf{0.952}$            & $\mathbf{0.718}$            & $\mathbf{0.883}$           & $\mathbf{0.698}$           & $\mathbf{0.777}$ & $\mathbf{0.396}$ & $\mathbf{0.731}$ & $\mathbf{0.714}$ \\
    \bottomrule
  \end{tabular}
\end{table}

\paragraph{Evaluation set-up}  We believe an important property of a `foundation model' is effective `out-of-the-box' embeddings, enabling strong downstream performance with limited task-specific supervision, and downstream multimodal fusion with other information streams, e.g.~electronic healthcare records. In computer vision, representation quality is commonly evaluated with linear or MLP probes, e.g.~\cite{assranSelfSupervisedLearningImages2023, caronEmergingPropertiesSelfSupervised2021}. However, existing sleep foundation models have varied drastically in their evaluation set-up.

We aimed to standardise the use of embeddings to perform a fair comparison across different foundation models. We primarily report linear and MLP probing performance, applying simple temporal and modality pooling for each model where necessary to evaluate at 30-second resolution on downstream tasks. We believe this most closely matches evaluations in domains such as computer vision, e.g.~mean-pooled embeddings from image patches. For OSF, which uses 30-second windows of input data, we use the dedicated CLS token for downstream probing experiments. For SleepFM and sleep2vec, which generate embeddings for each modality at different temporal resolutions, we first mean-pooled in the temporal dimension to obtain embeddings for each 30-second window. To combine embeddings from each modality, we evaluated both mean-pooling and concatenation and found the latter to work best across baselines.

\paragraph{Sleep analysis tasks} We primarily compare performance on the following tasks:
\begin{itemize}
  \item \textbf{Staging}: Sleep stage classification into one of five classes: Wake, N1, N2, N3 (deep) or rapid-eye-movement (REM) sleep.
  \item \textbf{Arousal}: Detection of cortical arousal events, characterised by an abrupt shift to wakefulness-like EEG activity. A high rate of cortical arousals during sleep correlates with excessive daytime sleepiness, impaired vigilance/cognition, and reduced quality of life.
  \item \textbf{Apnoea}: Detection of apnoea events, i.e. a temporary cessation in breathing. All apnoeas (central, obstructive, mixed) are aggregated into a single class for the purpose of evaluation.
  \item \textbf{Desat.}: Detection of blood oxygen desaturation events (transient drops in $\mathrm{SpO}_2$).
\end{itemize}
Annotations for each task come from the NSRR-harmonised scoring provided alongside each overnight recording.

\paragraph{Probe training} Each foundation model was evaluated with an identical probing pipeline. Before fitting any probe, we standardised each embedding dimension to zero mean and unit variance, with the scaling statistics computed on the training split only and then applied to the validation and test splits. The \emph{linear probe} is a single linear layer (multinomial logistic regression), trained to minimise cross-entropy with AdamW (learning rate $10^{-3}$, weight decay $10^{-4}$) for 30 epochs. The \emph{MLP probe} has two hidden layers of width 512 and 256 with ReLU activations and dropout ($p\!=\!0.1$), and is trained to minimise cross-entropy with AdamW (learning rate $10^{-3}$, weight decay $10^{-4}$) for up to 10 epochs, with early stopping on a held-out validation split (patience 2). A separate probe is fit per task, and a fixed random seed is used throughout. We used the same probe configurations and hyper-parameters for all models reported. Probe hyper-parameters were chosen as sensible defaults and were not optimised.

\subsection{Compute usage}\label{section:compute}
Each tokenizer described in \Cref{section:tokenization} was trained using a single NVIDIA H100 GPU using bf16 mixed-precision, requiring 60 GB of GPU RAM and around 5 hours of training time. After training, tokenization was performed using a single NVIDIA L40S GPU, with each entire overnight recording of each channel (10+ hours) tokenized in a single forward pass in around 250 ms. All Hypnos models were trained using H100 GPUs with bf16 mixed-precision training and activation checkpointing in each Transformer layer. Training Hypnos-Base to 50k steps required 1.5 days distributed across 8 GPUs, using around 45 GB of RAM on each GPU. For downstream probing evaluations, embeddings were generated using a single NVIDIA L40S GPU, with the embedding of each overnight recording taking around 3 seconds.

To reduce compute usage, many of our experiments including initial model design and ablation studies were performed using a unimodal variant of Hypnos-Small using EEG data, which took around 1 day to train on a single H100 GPU. Across all experiments our total compute usage was approximately 8000 H100 GPU-hours.

\section{Additional Experiments}

\subsection{Tokenizer design}\label{section:appendix:tokenizer_design}

\begin{figure}[h]
    \centering
    \includegraphics[width=\linewidth]{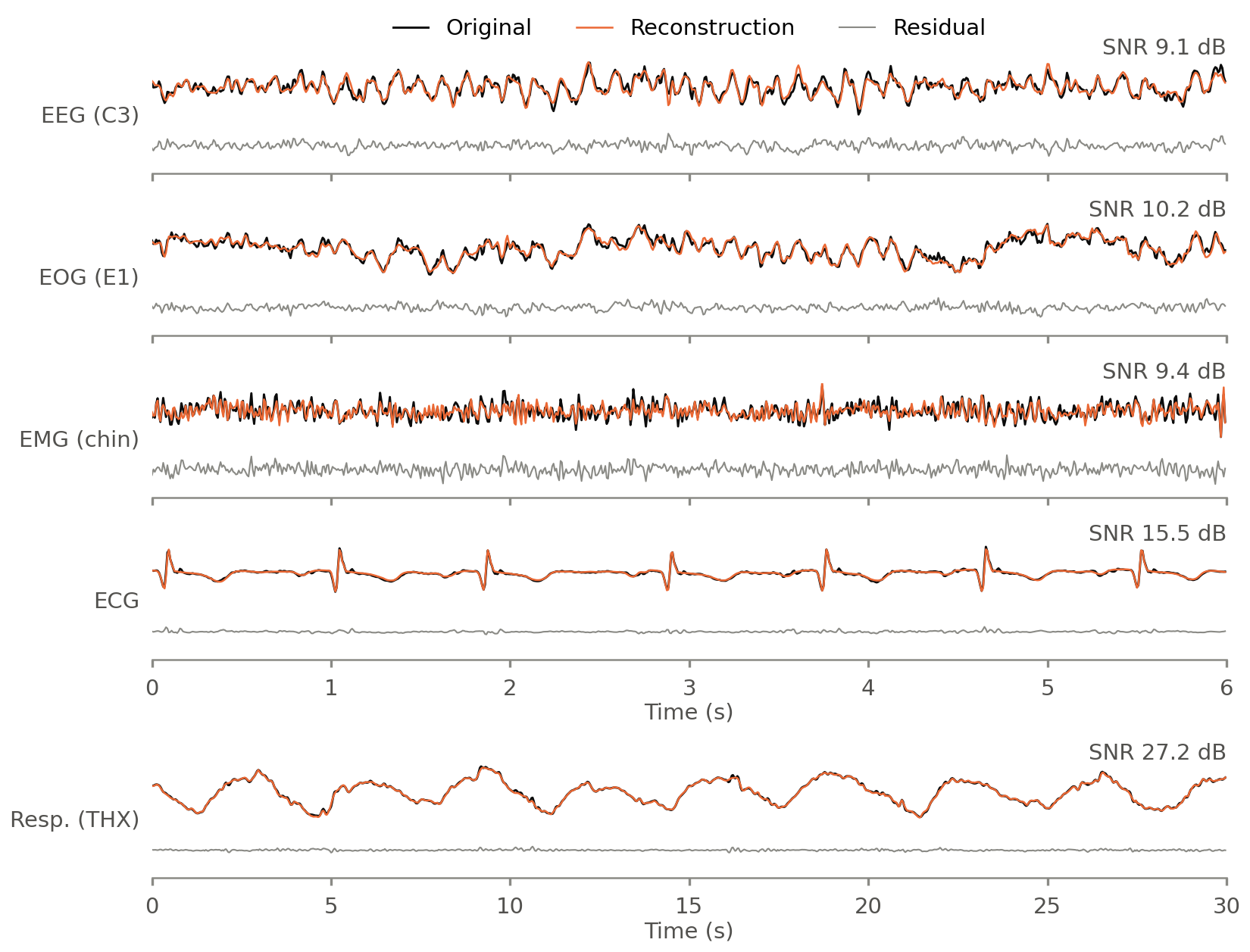}
    \caption{\textbf{Tokenizer reconstructions on held-out MrOS.} One time-aligned example per tokenizer, with the residual (original minus reconstruction) on the same scale below each trace. Respiration is shown over 30\,s to span several breaths. The SNR on each row is the mean over held-out records.\label{fig:tokenizer-recon}}
  \end{figure}

  \paragraph{Reconstruction quality} Tokenizer model capacities were chosen to achieve near lossless compression before downstream sequence modelling. \Cref{fig:tokenizer-recon} shows reconstructions from each of our five tokenizers on held-out MrOS recordings. Over eight random 64-second windows from each of 32 records, mean SNR is $9.1$\,dB for EEG (C3; $9.2$\,dB for C4), $10.2$\,dB for EOG (E1 and E2), $9.4$\,dB for chin EMG (28 records with this channel), $15.5$\,dB for ECG, and $27.2$\,dB and $29.0$\,dB for thoracic and abdominal respiration, with mean Pearson correlations from $0.91$ to $0.99$. Reconstruction is weakest for EEG, EOG and EMG, where small residuals remain visible.

\begin{figure}[h]
  \centering
  \includegraphics[width=0.7\linewidth]{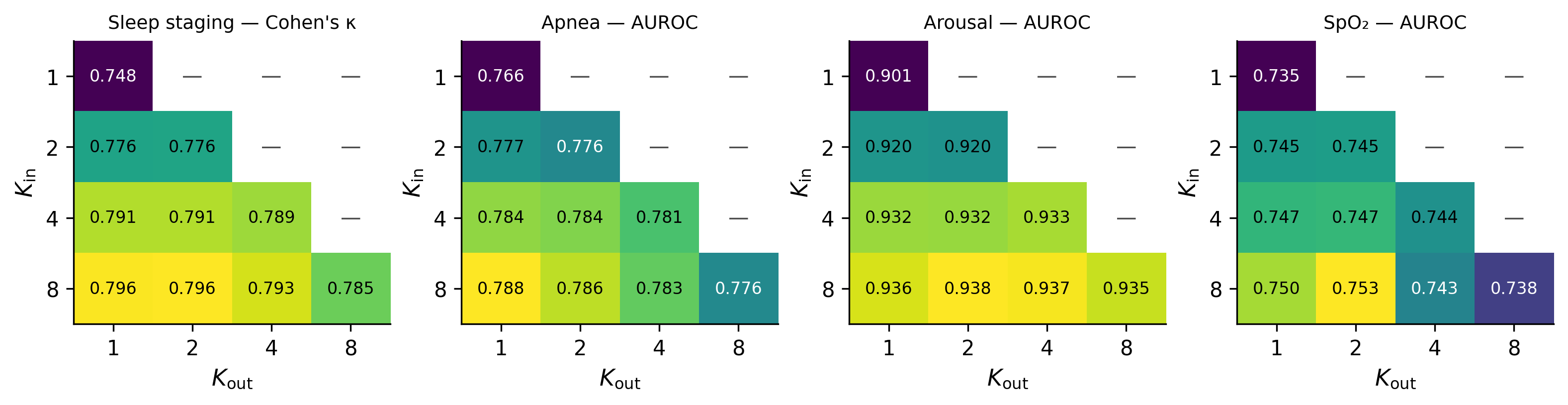}
  \caption{\textbf{Effect of input and output residual depth on downstream performance.} Linear probe metrics for unimodal EEG models on the SHHS validation set varying $K_{in}$ and $K_{out}$. Performance slightly improves when increasing the number of input residual tokens but worsens when increasing the number of output residual tokens.\label{fig:q-depth}}
\end{figure}

\paragraph{Residual depth} BrainOmni uses a codebook with \(K=4\) quantization layers, which leads to visibly smoothed reconstructions of EEG data (see Fig.~4 of~\cite{xiaoBrainOmniBrainFoundation2025}). Instead, we used \(K=8\) quantizers for neural signals to increase reconstruction accuracy of higher frequency details, e.g.\ gamma activity. Here we investigate the effect of residual depth on downstream performance. We vary the quantization depth at both the input and output to Hypnos, $K_{in}$ and $K_{out}$, which determine the residual tokens available for sequence modelling (\Cref{fig:arch}a) and the residual tokens to be predicted (\Cref{fig:arch}c) respectively. \Cref{fig:q-depth} shows the performance of unimodal Hypnos variants across downstream tasks using EEG data as we vary $K_{in}$ and $K_{out}$. We generally observe improved performance as we increase $K_{in}$, but a decrease in performance as we increase $K_{out}$. This indicates that high-frequency information is useful for sequence modelling, but trying to predict high-frequency information does not improve the quality of the learnt representation on the tasks evaluated. This suggests that $K_{out}$ could be decreased during training. Reducing $K_{out}$ from 8 to 2 would reduce Depth Transformer FLOPs by 75\%, and overall FLOPs by around 25\% for Hypnos-Base. However, this would come at the expense of generative capabilities.

\begin{figure}[h]
  \centering
  \begin{subfigure}[t]{0.19\linewidth}
    \centering
    \includegraphics[width=\linewidth]{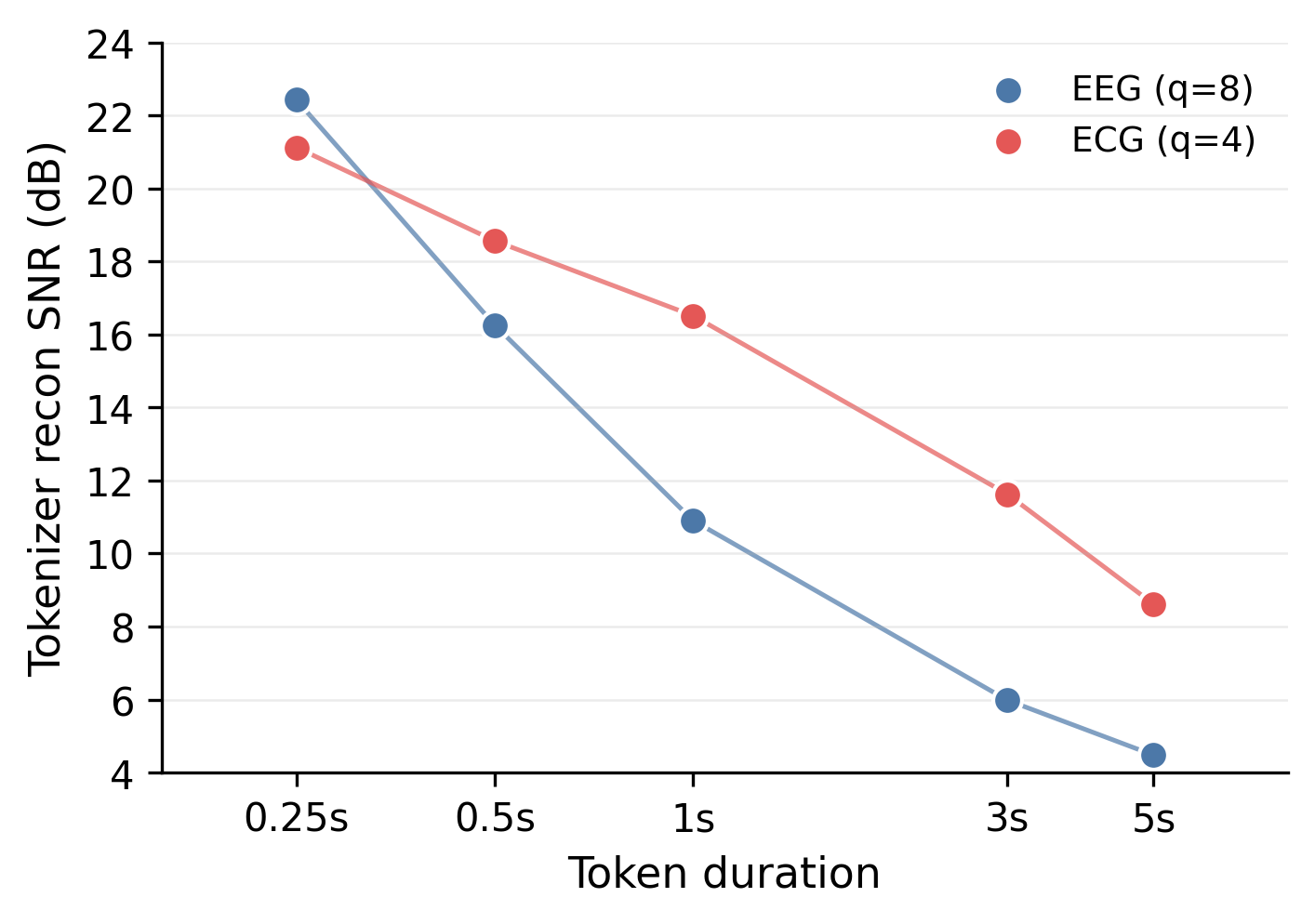}
    \caption{Recon. SNR}\label{fig:tokdur-snr}
  \end{subfigure}\hfill
  \begin{subfigure}[t]{0.19\linewidth}
    \centering
    \includegraphics[width=\linewidth]{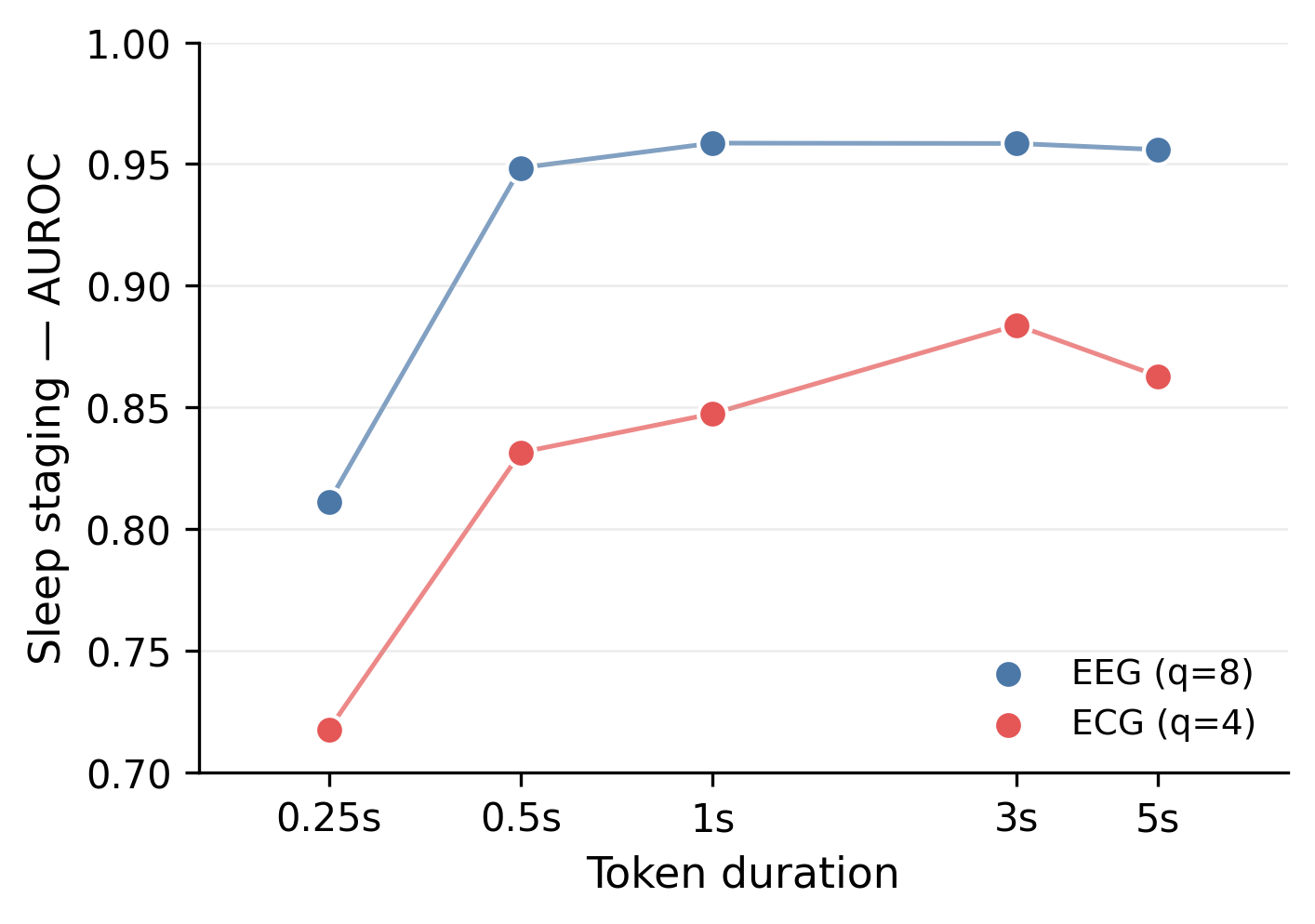}
    \caption{Sleep staging}\label{fig:tokdur-staging}
  \end{subfigure}\hfill
  \begin{subfigure}[t]{0.19\linewidth}
    \centering
    \includegraphics[width=\linewidth]{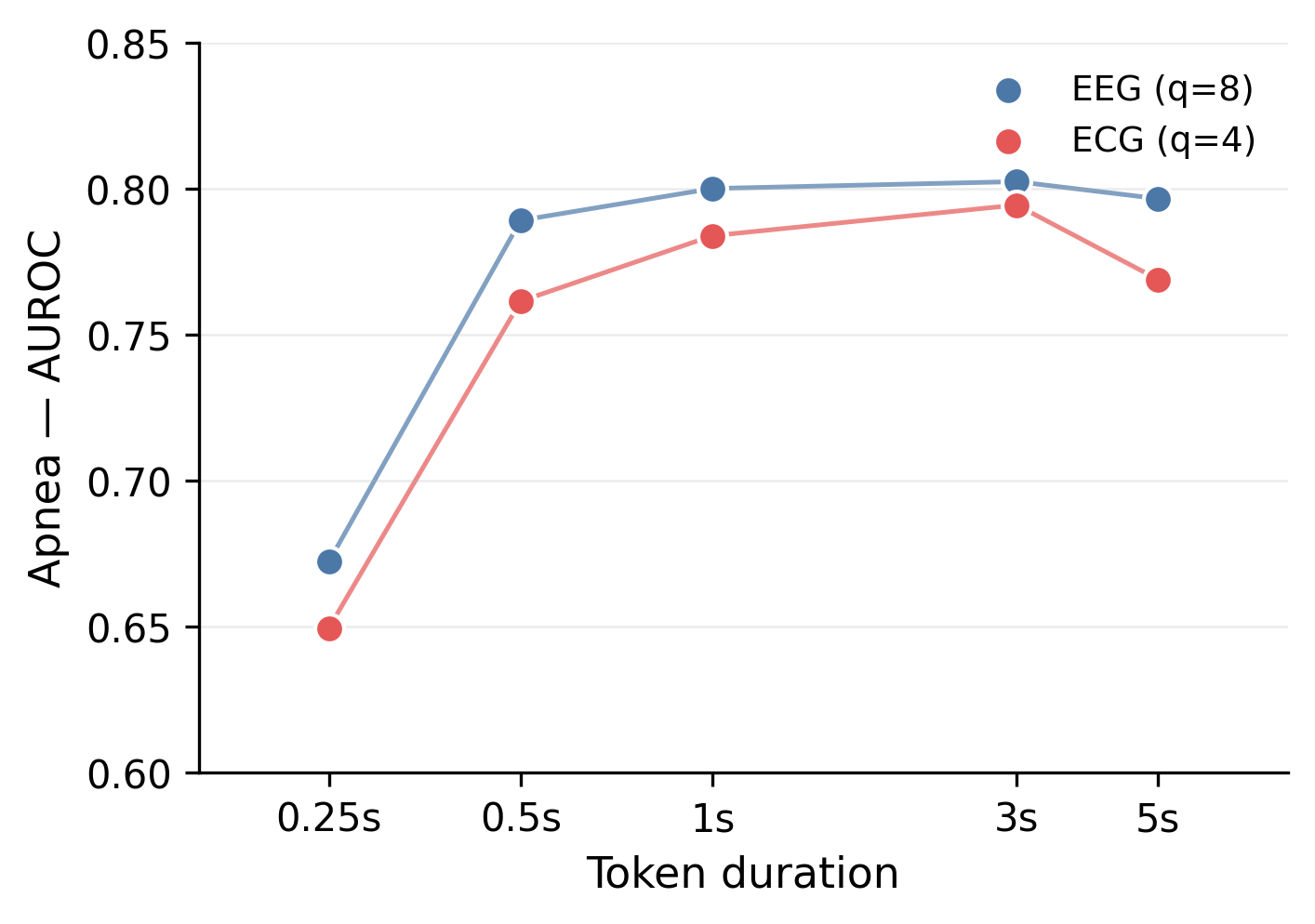}
    \caption{Apnoea detection}\label{fig:tokdur-apnea}
  \end{subfigure}\hfill
  \begin{subfigure}[t]{0.19\linewidth}
    \centering
    \includegraphics[width=\linewidth]{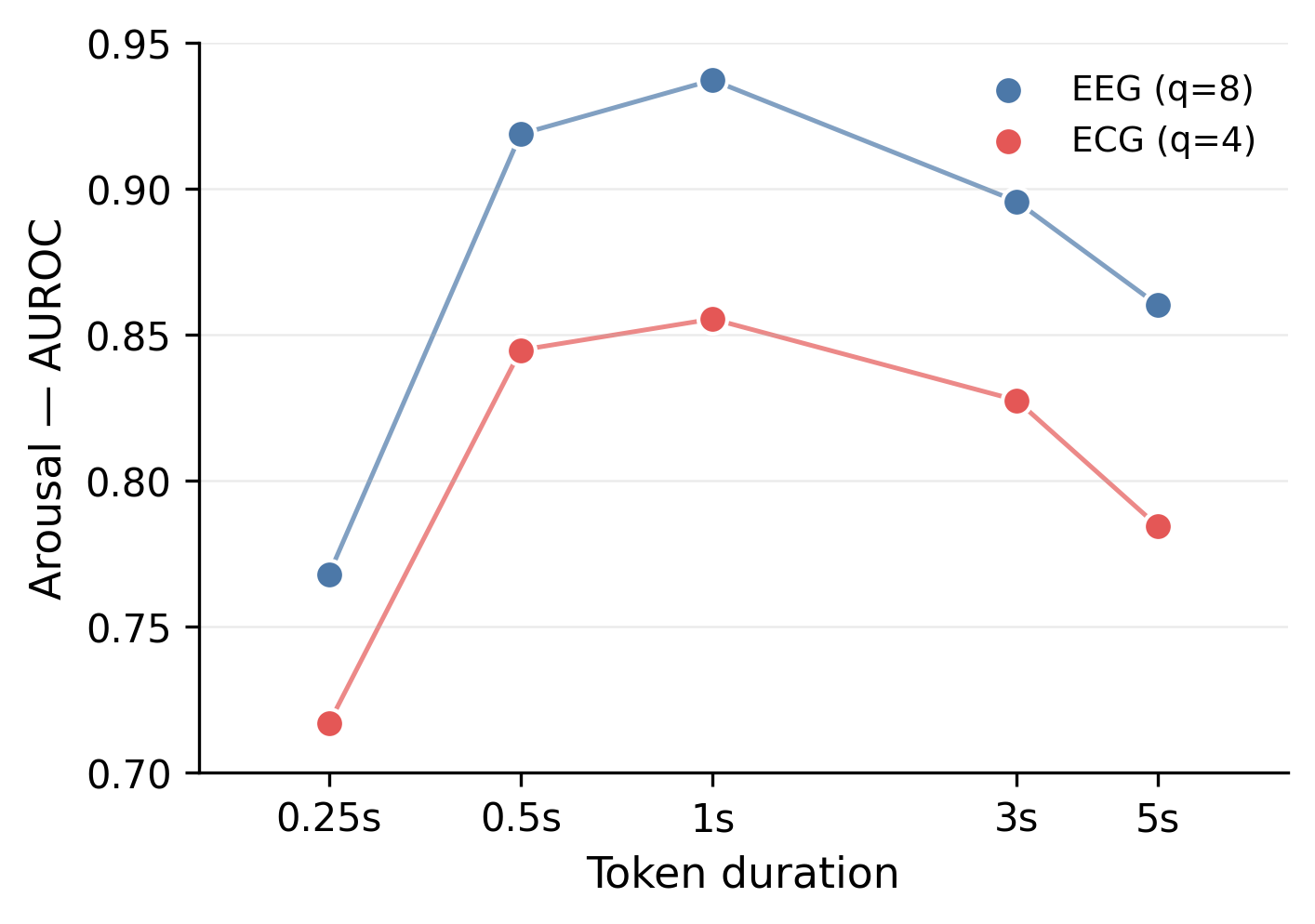}
    \caption{Arousal detection}\label{fig:tokdur-arousal}
  \end{subfigure}\hfill
  \begin{subfigure}[t]{0.19\linewidth}
    \centering
    \includegraphics[width=\linewidth]{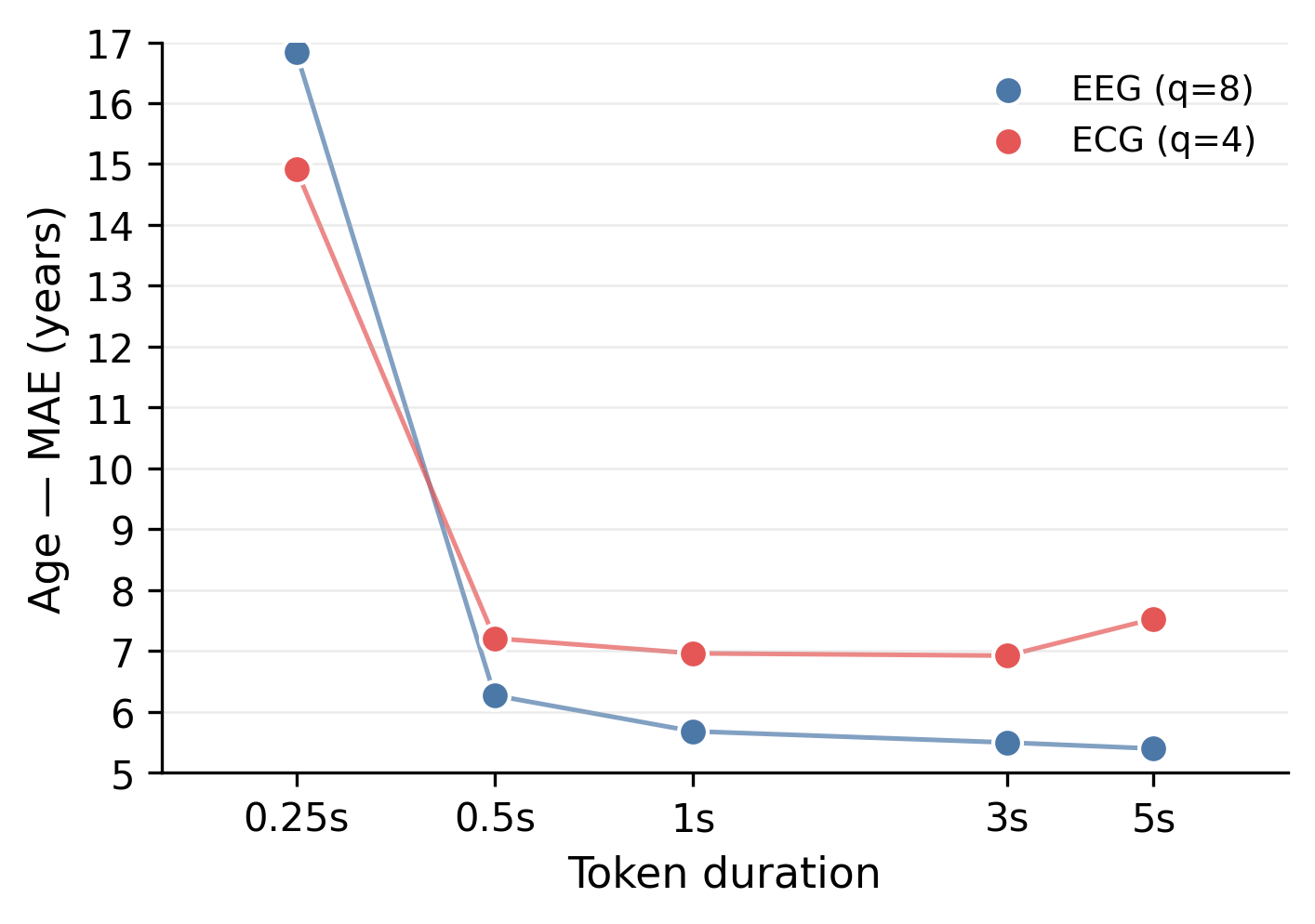}
    \caption{Age regression}\label{fig:tokdur-age}
  \end{subfigure}
  \caption{\textbf{Effect of token duration on downstream performance.} (left) Reconstruction quality decreases as token duration is varied from 0.25\,s to 5\,s, i.e. the compression rate increases. However, performance is worst at high token rates (0.25\,s) and saturates or regresses beyond 1\,s. We adopt a 1\,s token duration in all other experiments.\label{fig:token-duration}}
\end{figure}

\paragraph{Tokenization length}
In our main experiments, we designed our tokenizers to produce tokens at a rate of 1 Hz. This also determines the rate at which unique output embeddings are produced by the model. 1 Hz is a natural choice for real-world applications, aligning with standard units. Additionally, relevant physiological events such as heartbeats and sleep spindles~\cite{andrillonSleepSpindlesHumans2011} commonly occur on this timescale. Here we investigate the sensitivity of our approach to the tokenization length of different signals. For EEG and ECG signals, we trained 5 tokenizers with tokenization lengths, $\tau \in (0.25, 0.5, 1.0, 3.0, 5.0)$ seconds. This was achieved by modifying the convolutional stride lengths in the convolutional components of the tokenizers. For 3-second and 5-second tokenization lengths, we also increased the input sequence length by 3x and 5x respectively, so that the input sequence length to Transformer components in the tokenizers remained constant. We then re-tokenized all datasets and investigated the effect of tokenization length on downstream performance. \Cref{fig:token-duration} shows both the reconstruction SNR and downstream performances as we vary the tokenization length. As expected, increasing tokenization length with fixed model capacity leads to a decrease in signal-to-noise ratio as the compression rate increases. Above $\tau=0.25$ s, performance is reasonably stable across a range of token durations for both ECG and EEG data, with $\tau=1$ s working well across tasks and inputs. Performance is noticeably poorer with $\tau=0.25$ across tasks. For very small tokenization lengths $\tau$, the auto-regression task reduces to trivial extrapolation over short timescales, empirically reducing the quality of the learnt representations.

\paragraph{Adversarial losses} D\'efossez~\etal~\cite{defossezMoshiSpeechtextFoundation2024} recently observed that removing reconstruction losses and solely relying on adversarial losses led to better performance in downstream audio modelling tasks. In early experiments, we tried incorporating adversarial losses but found that this did not have a significant effect on downstream performance. Additionally, it significantly increased computational requirements: each training run required substantially higher activation memory for the discriminator; and, more training runs were required to find stable hyper-parameters given the well-known instability issues in adversarial training~\cite{salimansImprovedTechniquesTraining2016}.

\subsection{Scaling unimodal EEG models}\label{app:scaling-eeg}
\begin{figure}[h]
  \centering
  \begin{subfigure}[t]{0.24\linewidth}
    \centering
    \includegraphics[width=\linewidth]{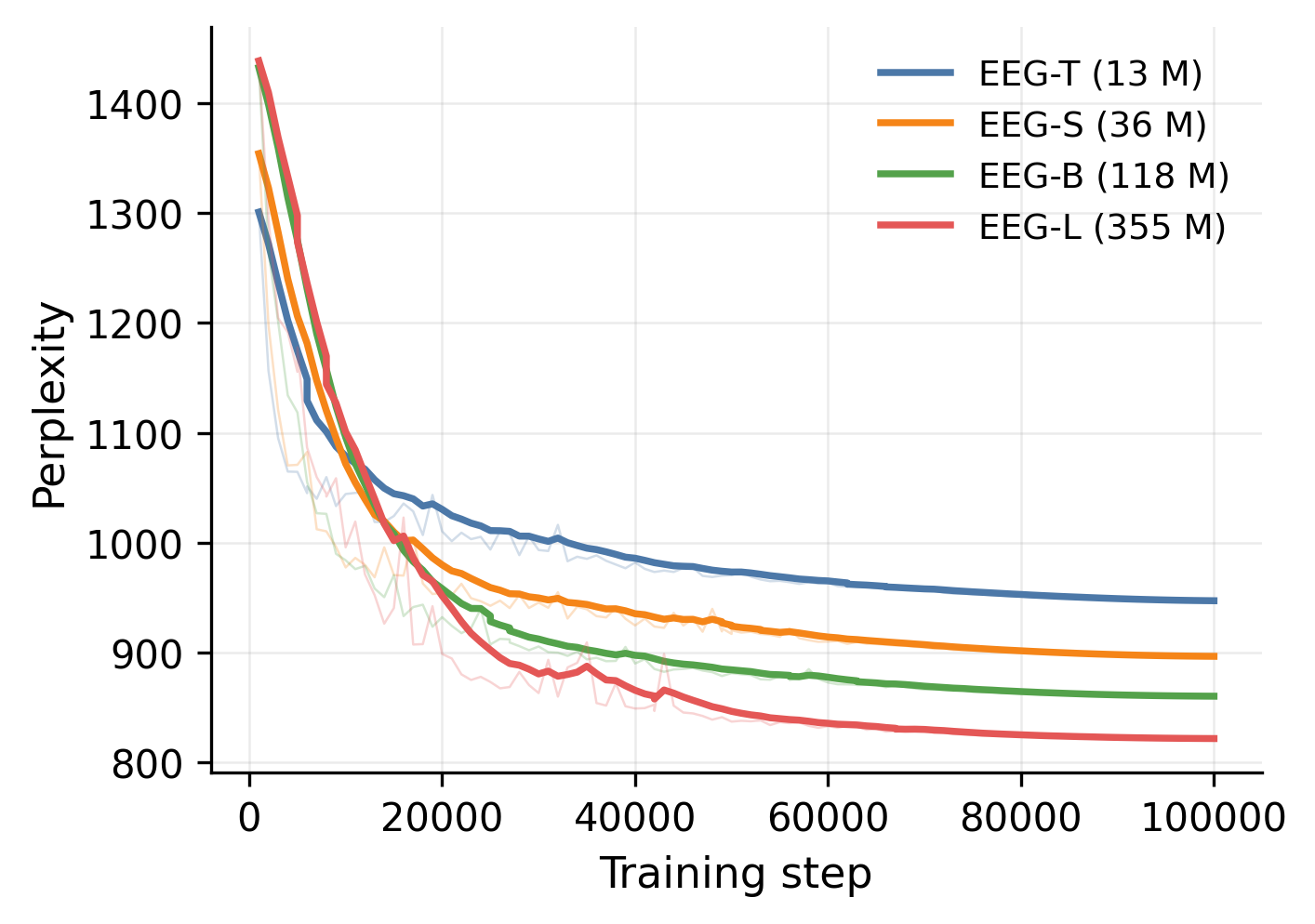}
    \caption{Validation loss}\label{fig:scaling-eeg-loss}
  \end{subfigure}\hfill
  \begin{subfigure}[t]{0.24\linewidth}
    \centering
    \includegraphics[width=\linewidth]{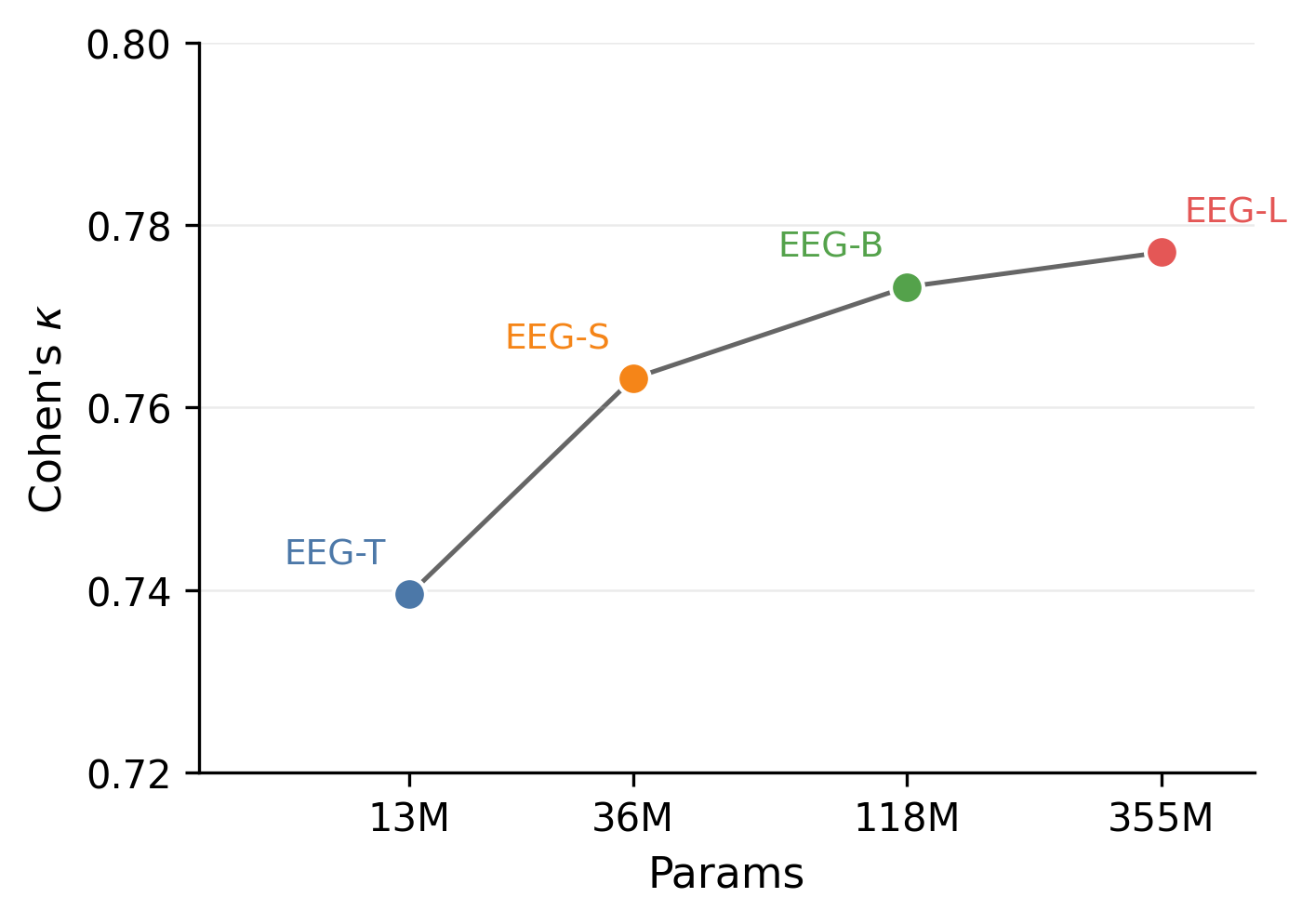}
    \caption{Sleep staging}\label{fig:scaling-eeg-staging}
  \end{subfigure}\hfill
  \begin{subfigure}[t]{0.24\linewidth}
    \centering
    \includegraphics[width=\linewidth]{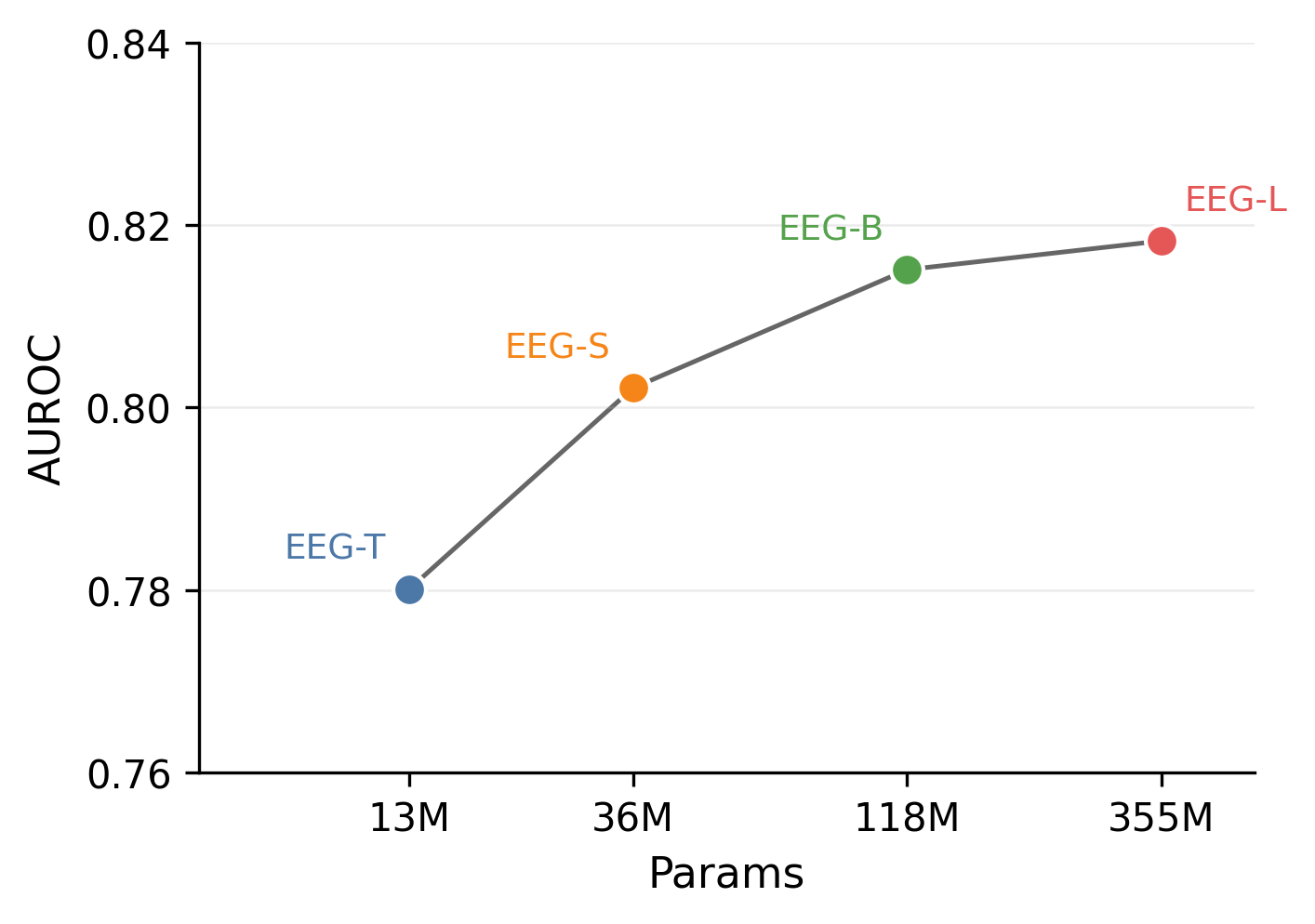}
    \caption{Apnoea detection}\label{fig:scaling-eeg-apnea}
  \end{subfigure}\hfill
  \begin{subfigure}[t]{0.24\linewidth}
    \centering
    \includegraphics[width=\linewidth]{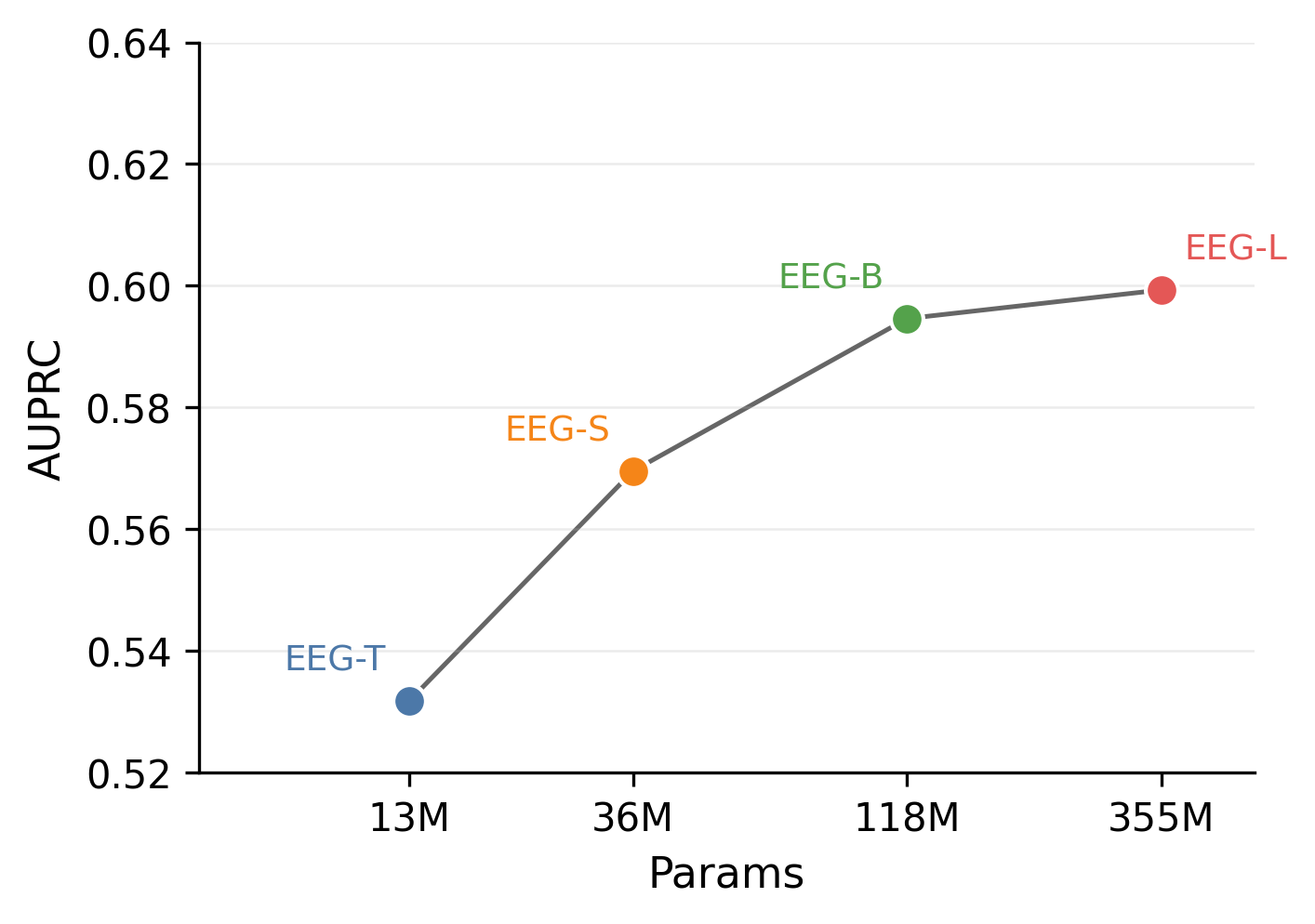}
    \caption{Arousal detection}\label{fig:scaling-eeg-arousal}
  \end{subfigure}
  \caption{\textbf{Scaling unimodal EEG models from Tiny to Large.} Next-token perplexity and downstream metrics continue to improve with model scale. Sleep stage classification, apnoea detection and arousal detection performance are reported using a linear probe on the SHHS validation set.\label{fig:scaling-eeg}}
\end{figure}

Compute requirements precluded scaling our multimodal model beyond Base size in our main experiments. To extend the scaling analysis to larger models, we trained unimodal EEG variants of Hypnos from Tiny up to Large. \Cref{fig:scaling-eeg} shows next-token perplexity and downstream probing performance on the SHHS validation set as we scale model size. Trends mirror those observed in the multimodal setting: validation loss and downstream metrics continue to improve with scale through to Large.

\subsection{Scaling context length}\label{app:context-length}
\begin{figure}[h]
  \centering
  \begin{subfigure}[t]{0.24\linewidth}
    \centering
    \includegraphics[width=\linewidth]{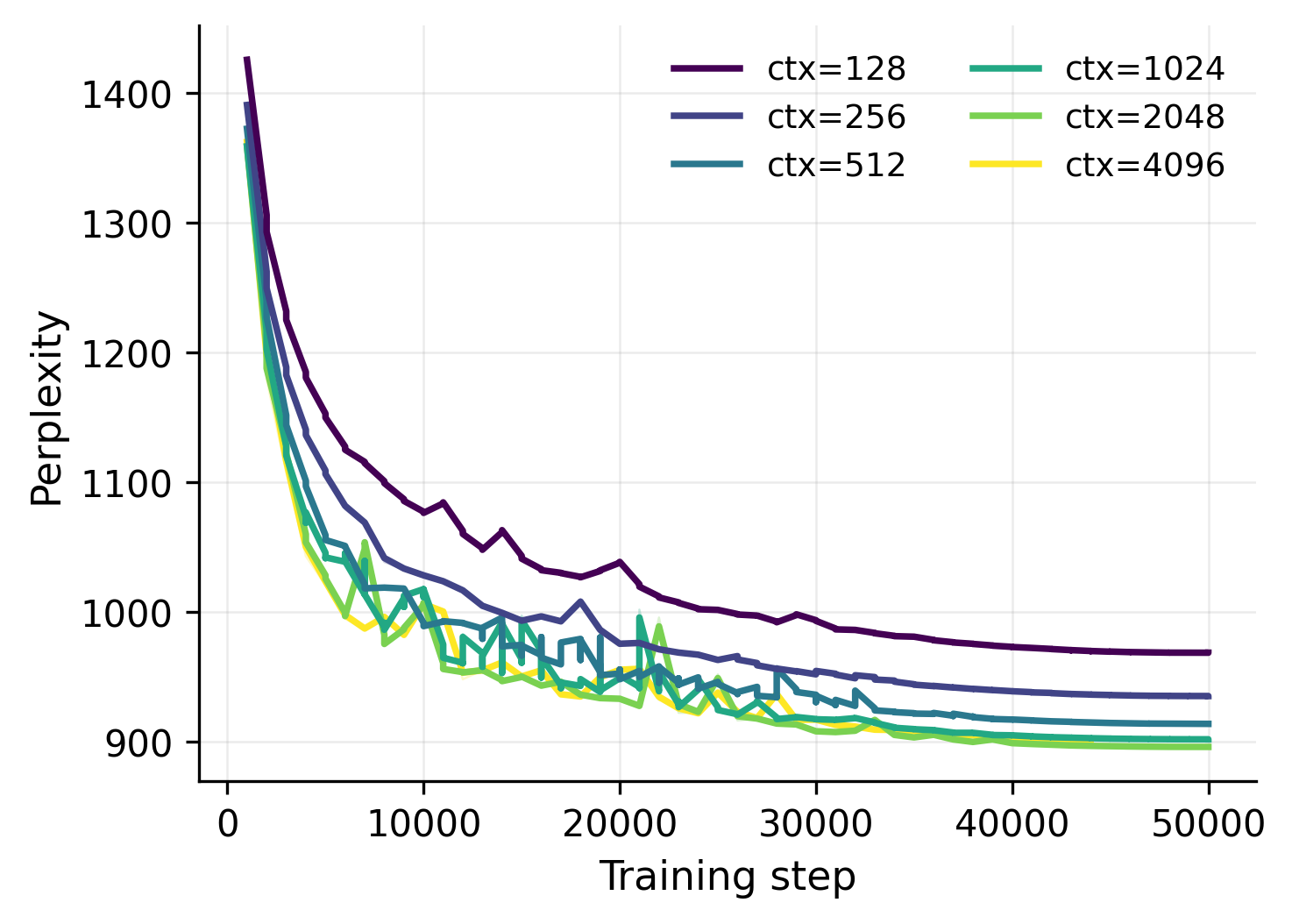}
    \caption{Validation perplexity}\label{fig:ctx-eeg-loss}
  \end{subfigure}\hfill
  \begin{subfigure}[t]{0.24\linewidth}
    \centering
    \includegraphics[width=\linewidth]{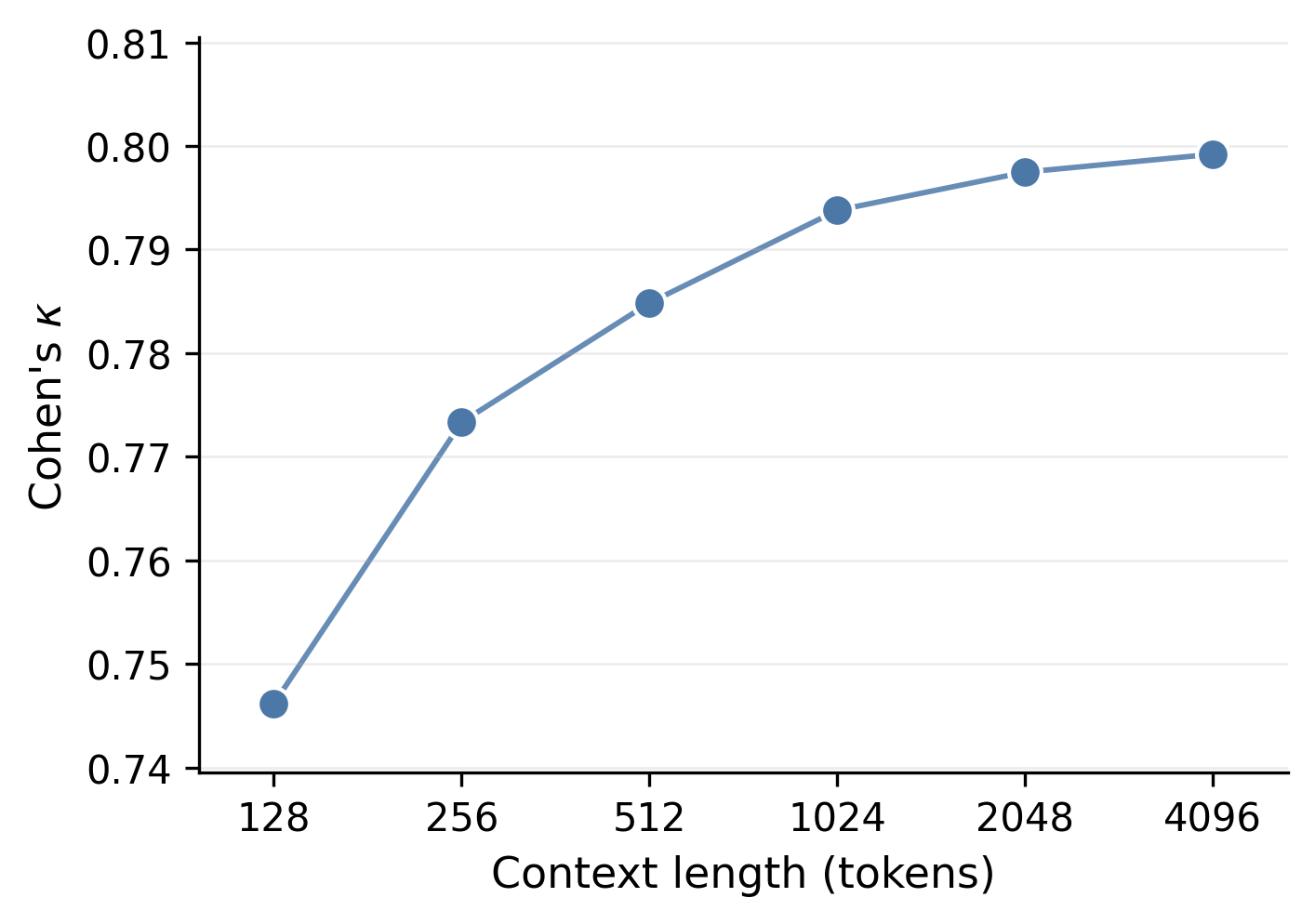}
    \caption{Sleep staging}\label{fig:ctx-eeg-staging}
  \end{subfigure}\hfill
  \begin{subfigure}[t]{0.24\linewidth}
    \centering
    \includegraphics[width=\linewidth]{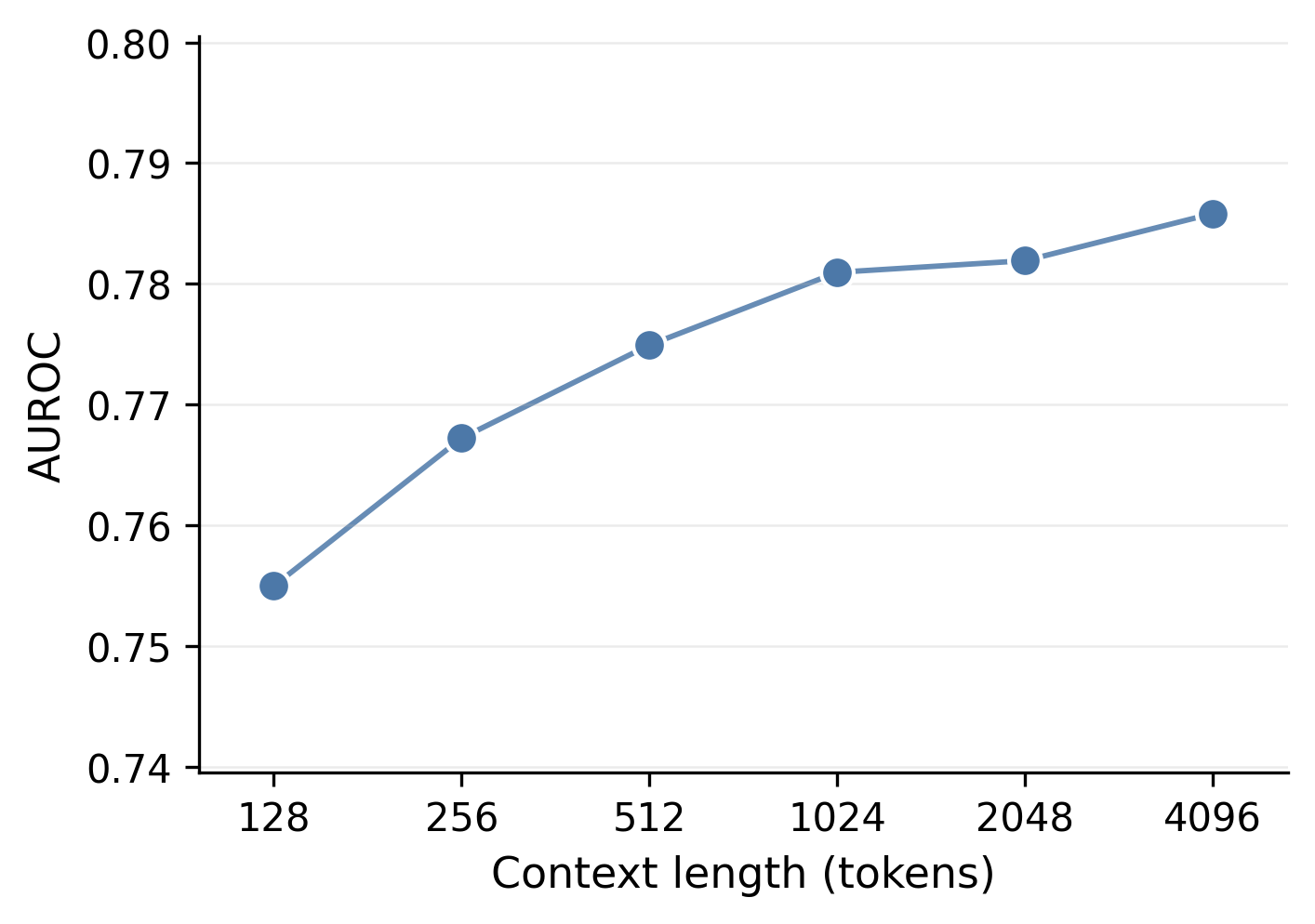}
    \caption{Apnoea detection}\label{fig:ctx-eeg-apnea}
  \end{subfigure}\hfill
  \begin{subfigure}[t]{0.24\linewidth}
    \centering
    \includegraphics[width=\linewidth]{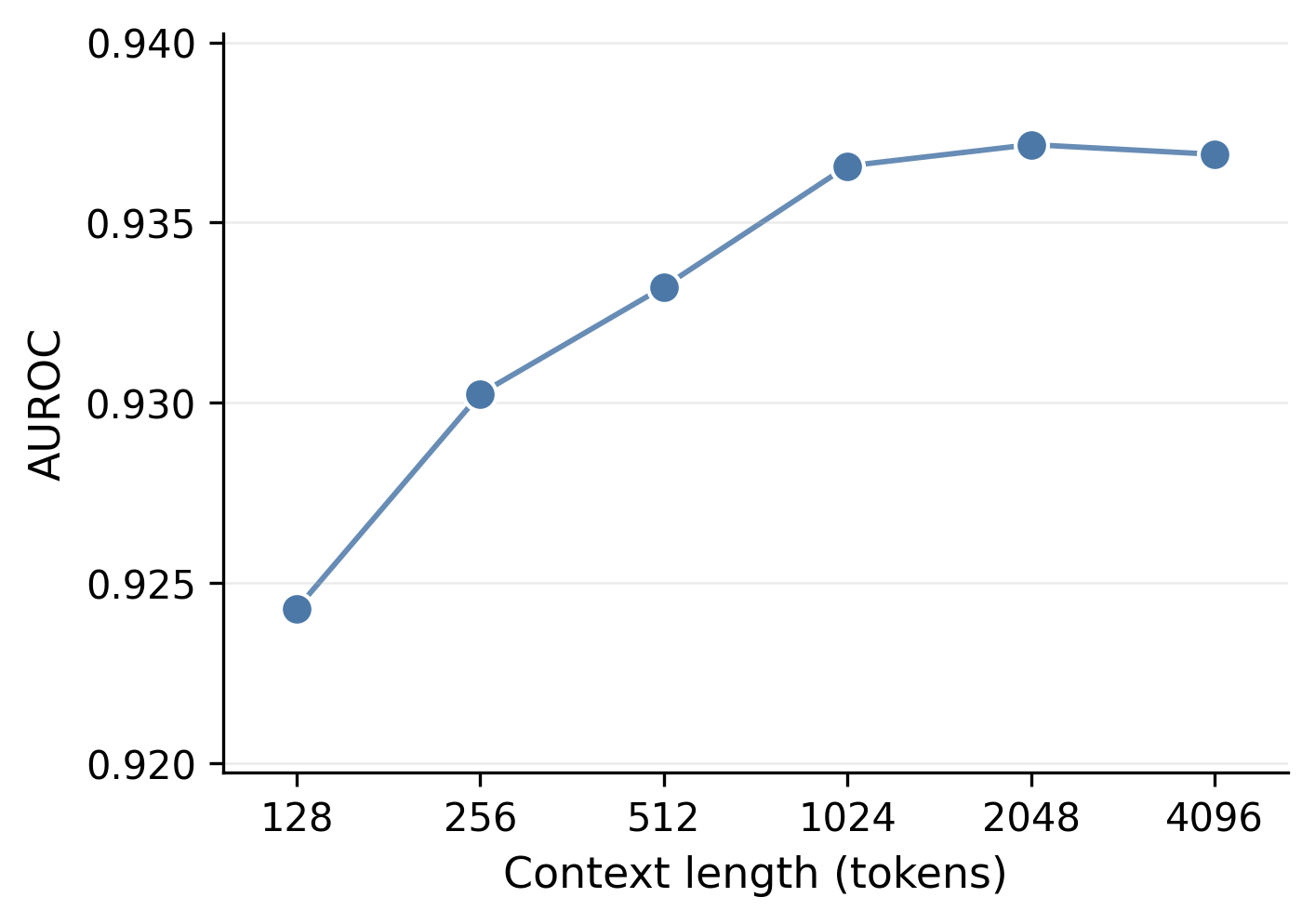}
    \caption{Arousal detection}\label{fig:ctx-eeg-arousal}
  \end{subfigure}

  \vspace{0.5em}
  \begin{subfigure}[t]{0.24\linewidth}
    \centering
    \includegraphics[width=\linewidth]{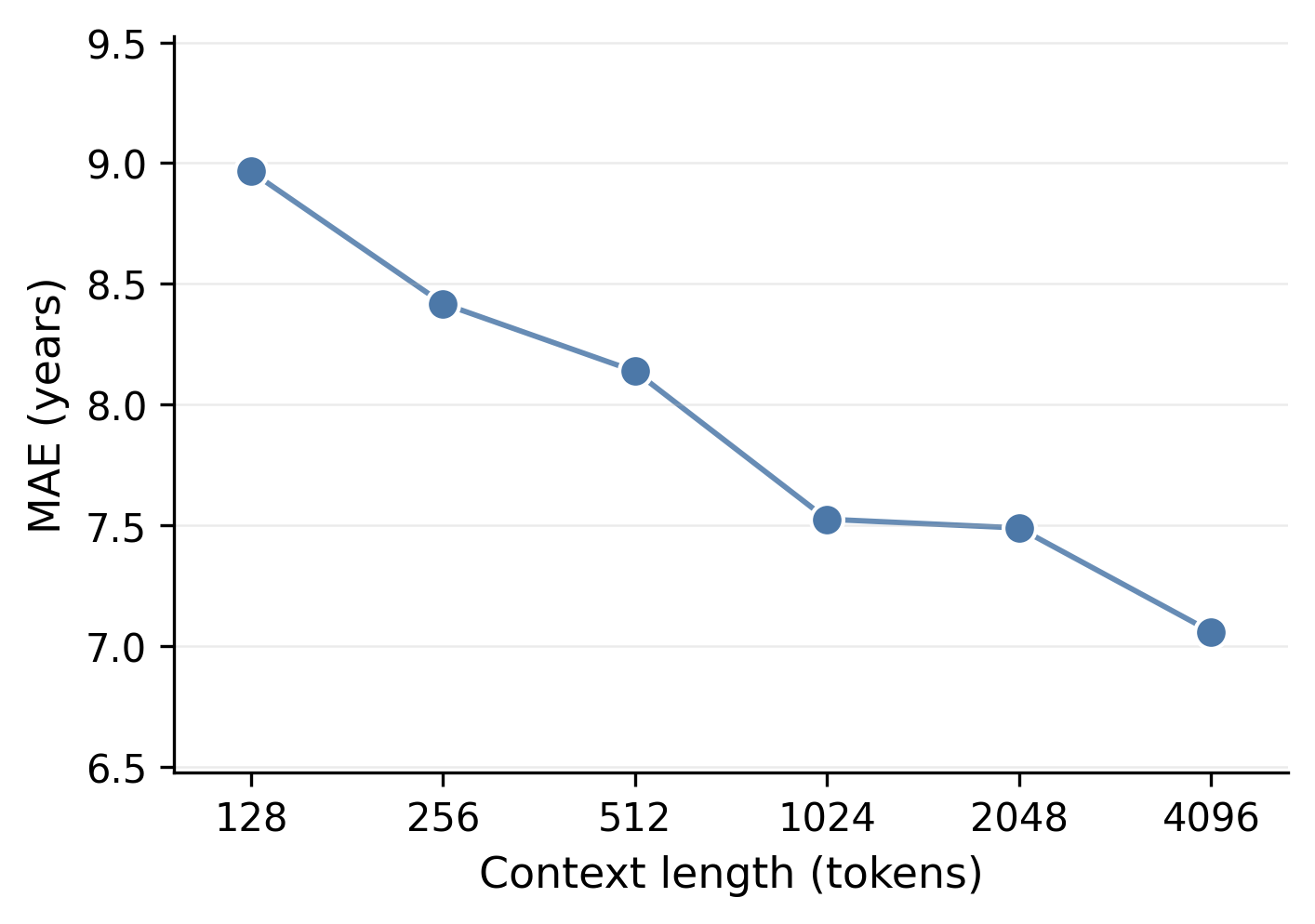}
    \caption{Age regression}\label{fig:ctx-eeg-age}
  \end{subfigure}\hfill
  \begin{subfigure}[t]{0.24\linewidth}
    \centering
    \includegraphics[width=\linewidth]{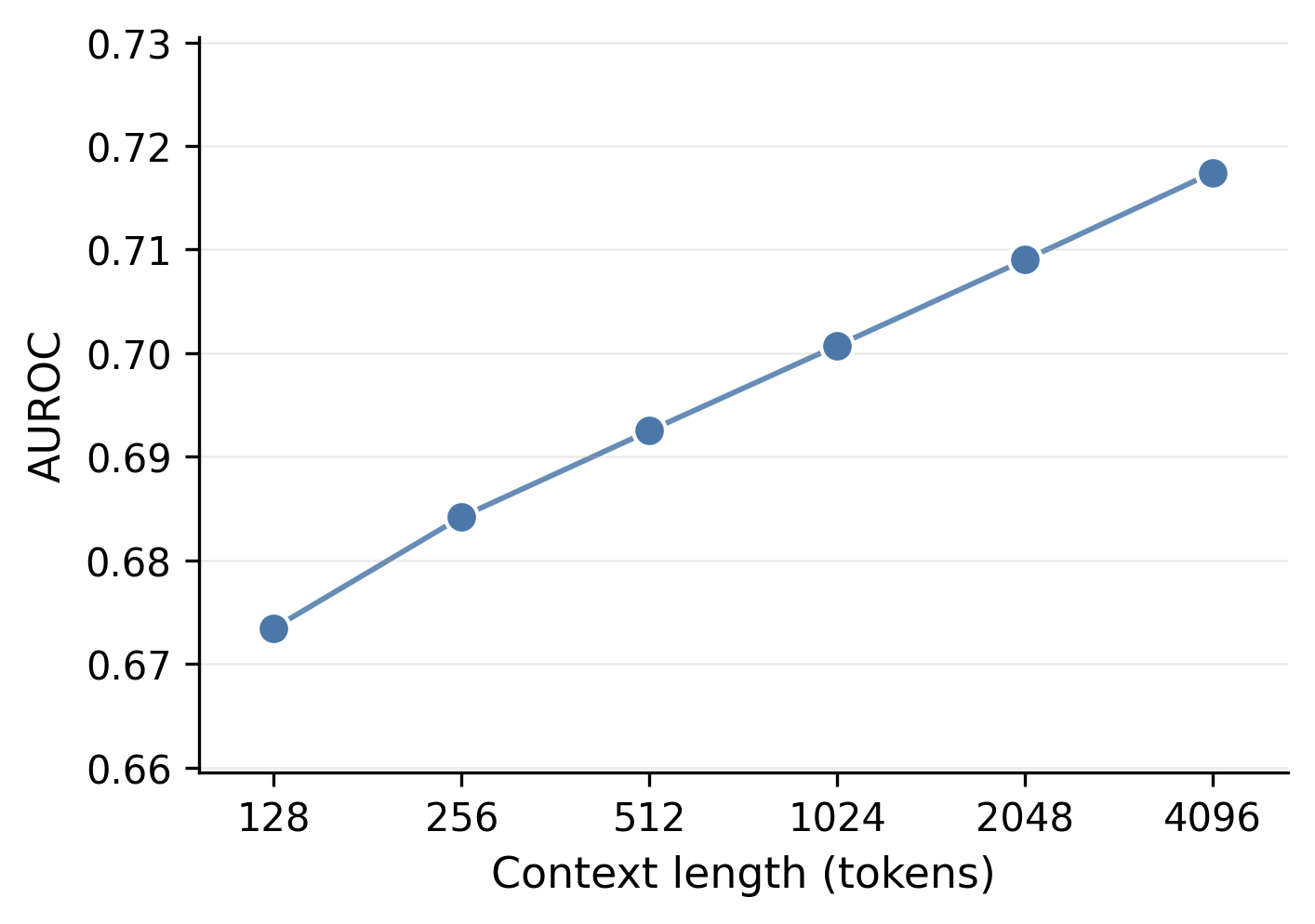}
    \caption{CVD risk}\label{fig:ctx-eeg-cvd}
  \end{subfigure}\hfill
  \begin{subfigure}[t]{0.24\linewidth}
    \centering
    \includegraphics[width=\linewidth]{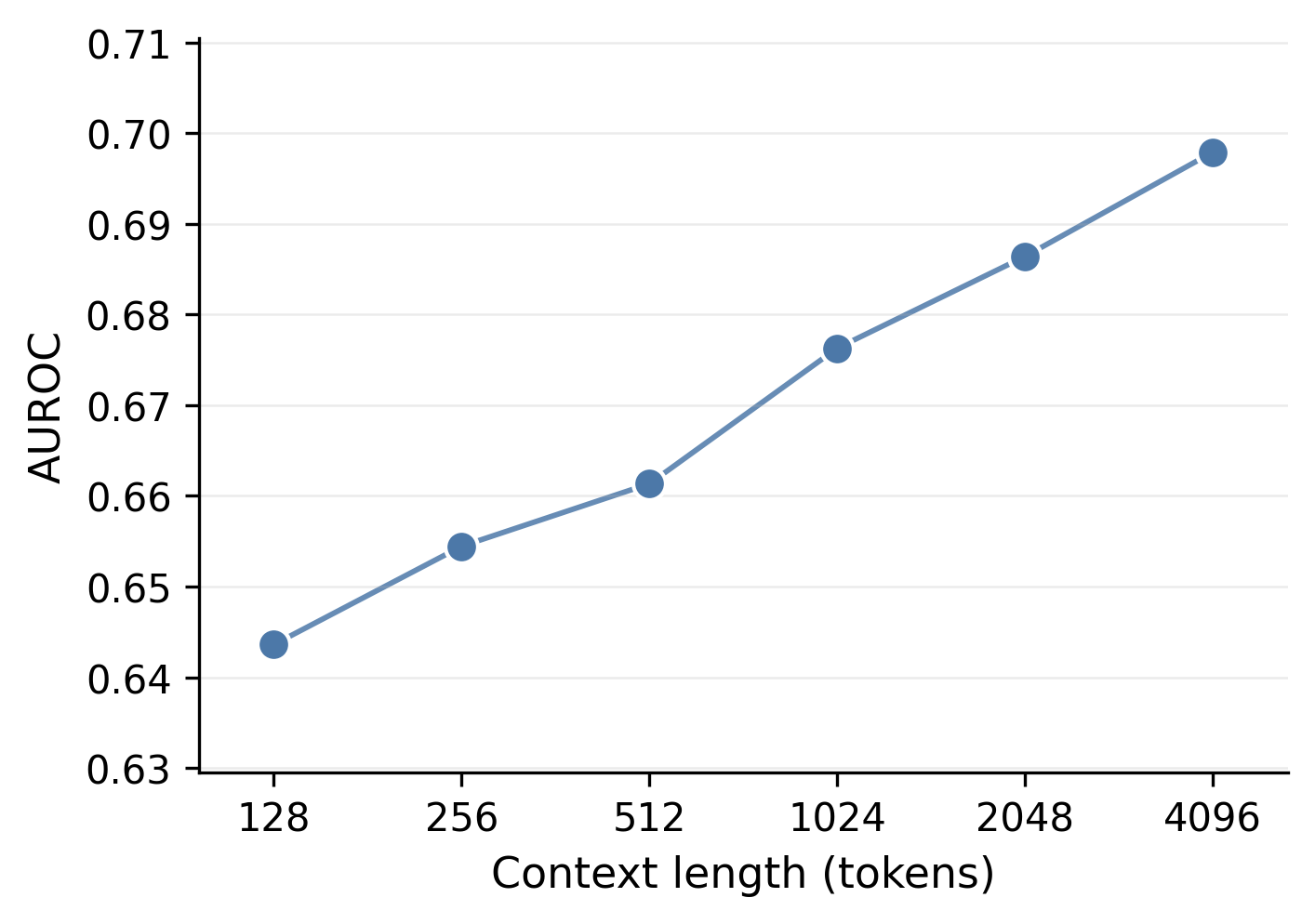}
    \caption{Moderate OSA}\label{fig:ctx-eeg-osa}
  \end{subfigure}\hfill
  \begin{subfigure}[t]{0.24\linewidth}
  \end{subfigure}
  \caption{\textbf{Effect of context length on single-channel EEG models.} Validation perplexity decreases and downstream probing performance improves as the training context length is increased from 128 to 4096 tokens. Sleep staging and arousal detection saturate at around 1024--2048 tokens, while age regression, CVD risk and moderate OSA detection continue to improve at the longest context lengths.\label{fig:ctx-eeg}}
\end{figure}

\begin{figure}[h]
  \centering
  \begin{subfigure}[t]{0.24\linewidth}
    \centering
    \includegraphics[width=\linewidth]{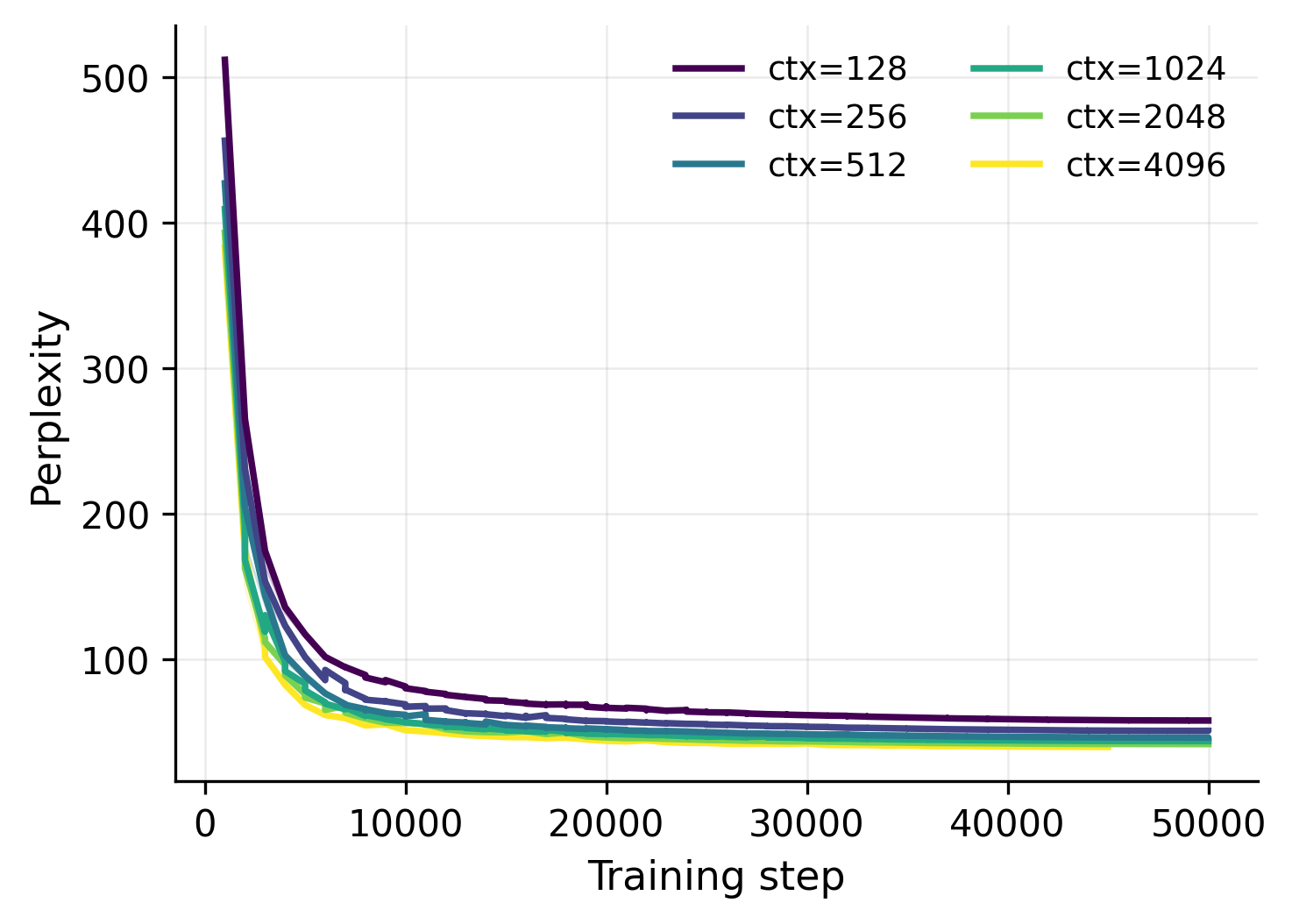}
    \caption{Validation perplexity}\label{fig:ctx-ecg-loss}
  \end{subfigure}\hfill
  \begin{subfigure}[t]{0.24\linewidth}
    \centering
    \includegraphics[width=\linewidth]{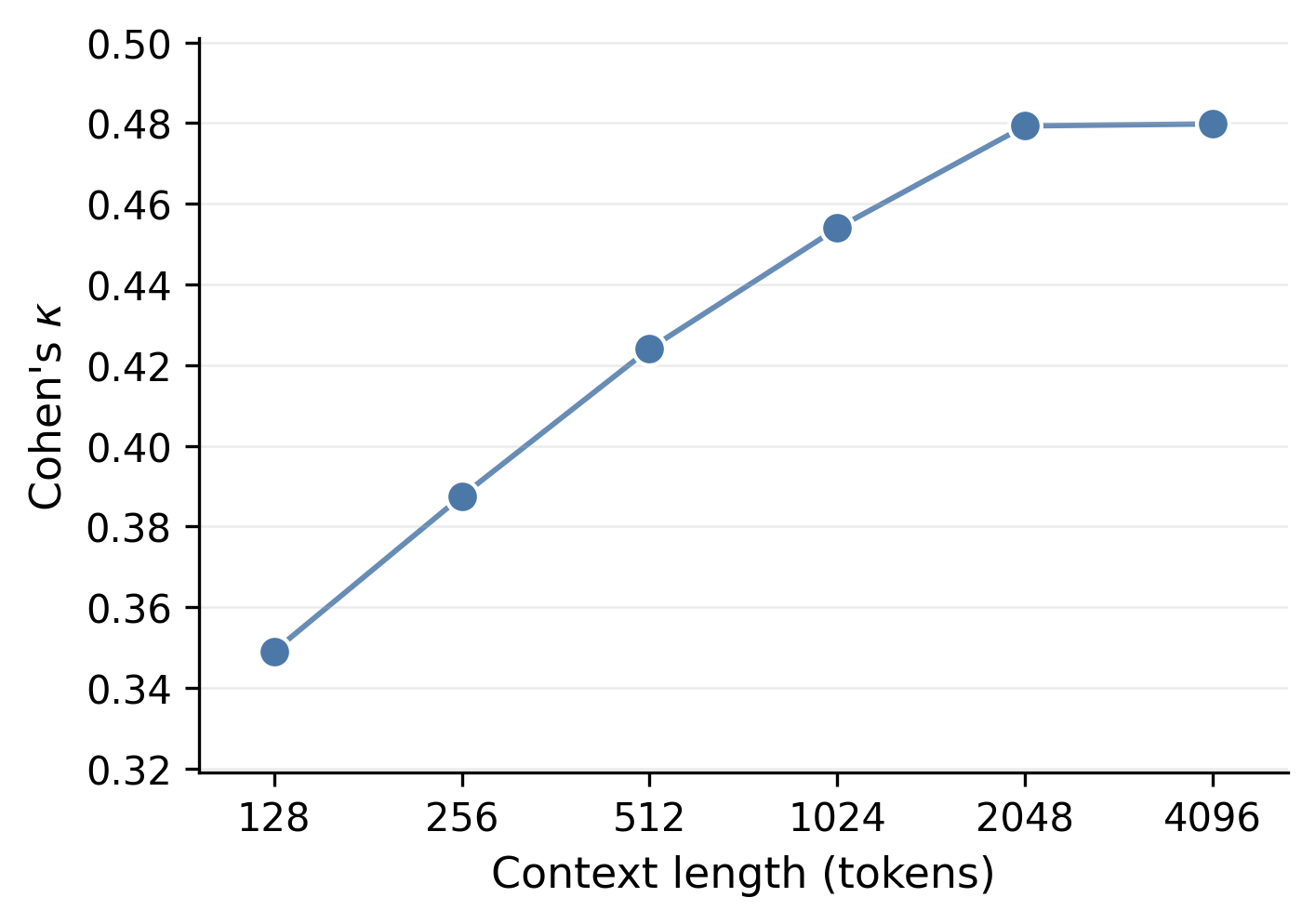}
    \caption{Sleep staging}\label{fig:ctx-ecg-staging}
  \end{subfigure}\hfill
  \begin{subfigure}[t]{0.24\linewidth}
    \centering
    \includegraphics[width=\linewidth]{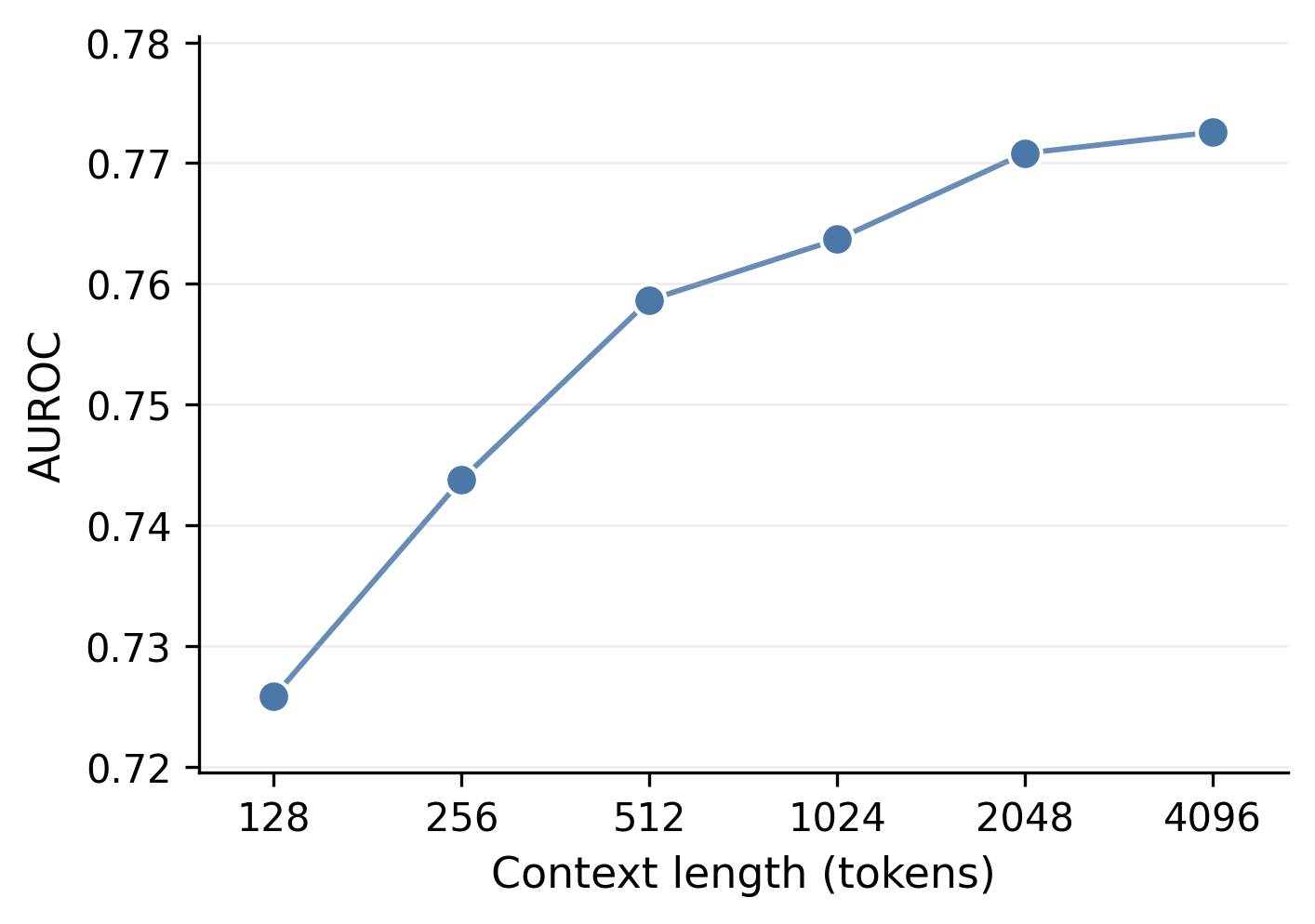}
    \caption{Apnoea detection}\label{fig:ctx-ecg-apnea}
  \end{subfigure}\hfill
  \begin{subfigure}[t]{0.24\linewidth}
    \centering
    \includegraphics[width=\linewidth]{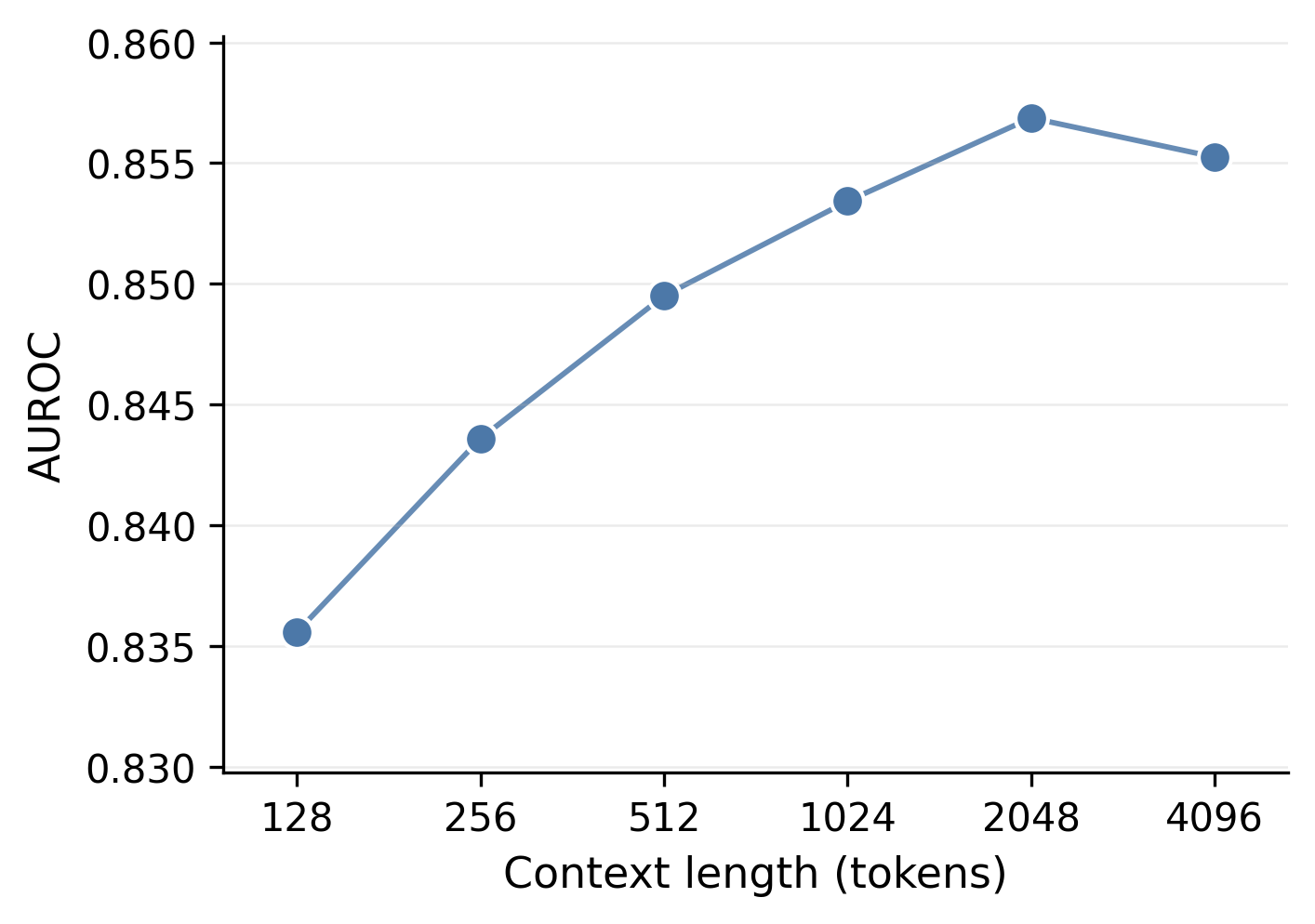}
    \caption{Arousal detection}\label{fig:ctx-ecg-arousal}
  \end{subfigure}

  \vspace{0.5em}
  \begin{subfigure}[t]{0.24\linewidth}
    \centering
    \includegraphics[width=\linewidth]{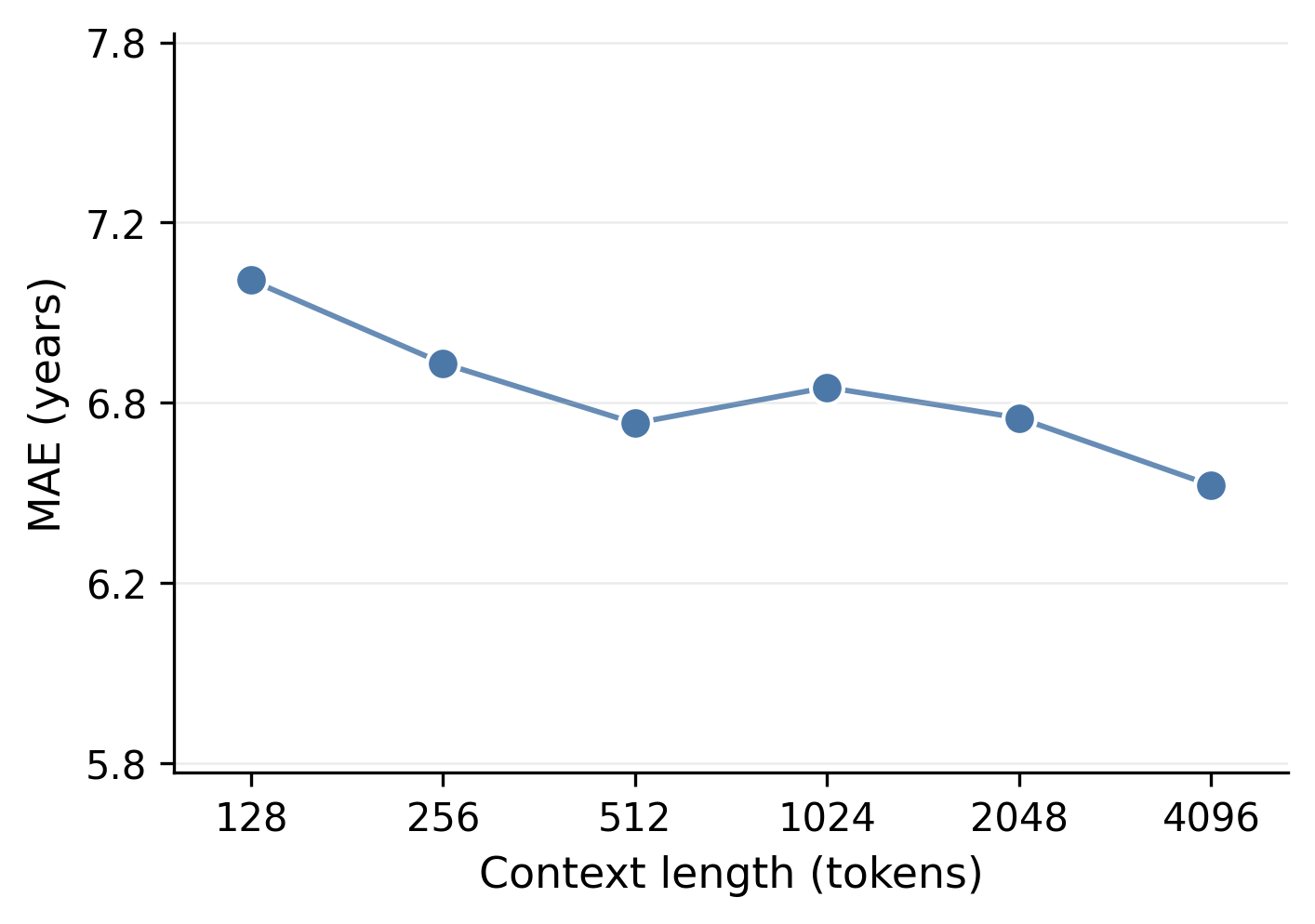}
    \caption{Age regression}\label{fig:ctx-ecg-age}
  \end{subfigure}\hfill
  \begin{subfigure}[t]{0.24\linewidth}
    \centering
    \includegraphics[width=\linewidth]{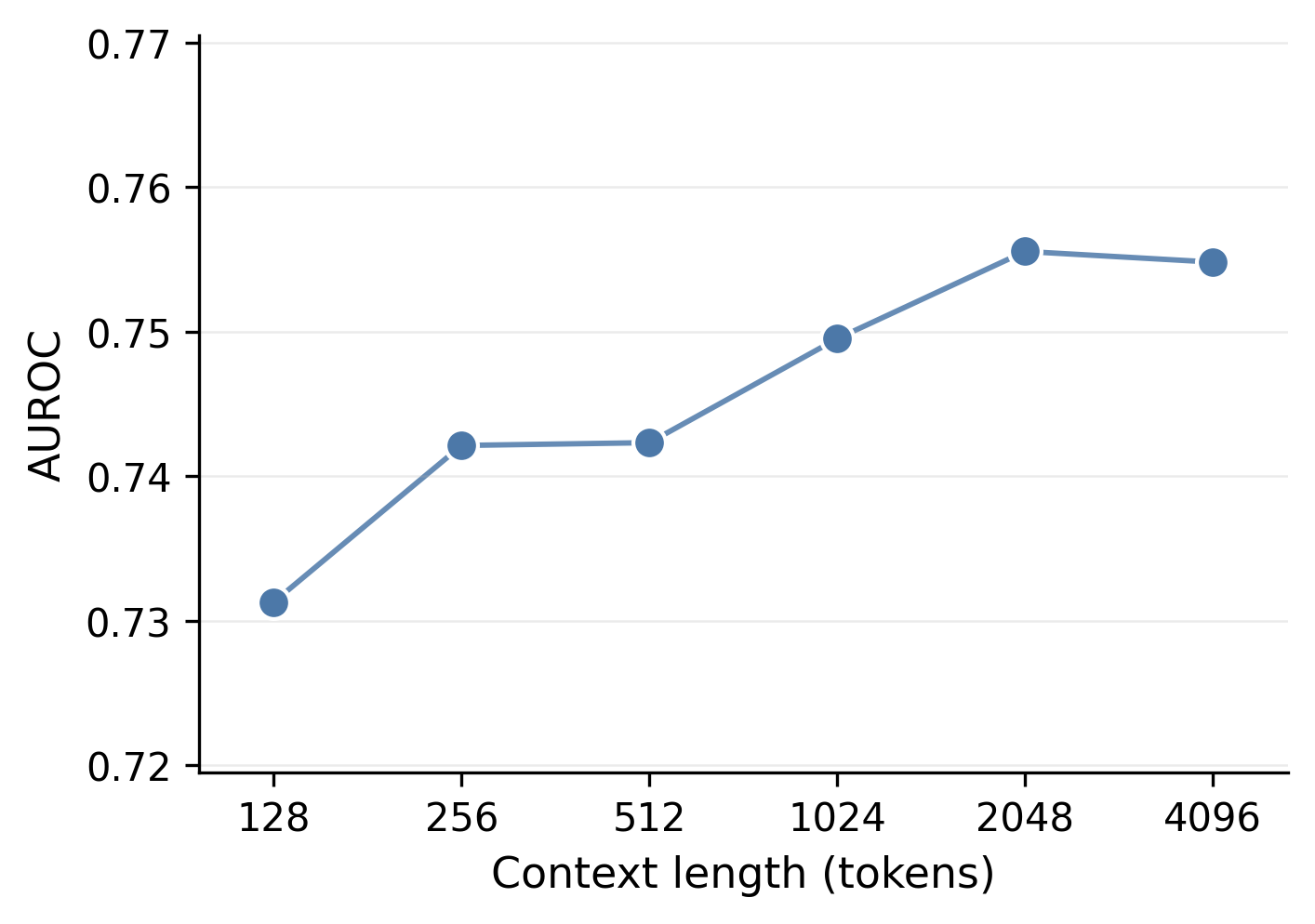}
    \caption{CVD risk}\label{fig:ctx-ecg-cvd}
  \end{subfigure}\hfill
  \begin{subfigure}[t]{0.24\linewidth}
    \centering
    \includegraphics[width=\linewidth]{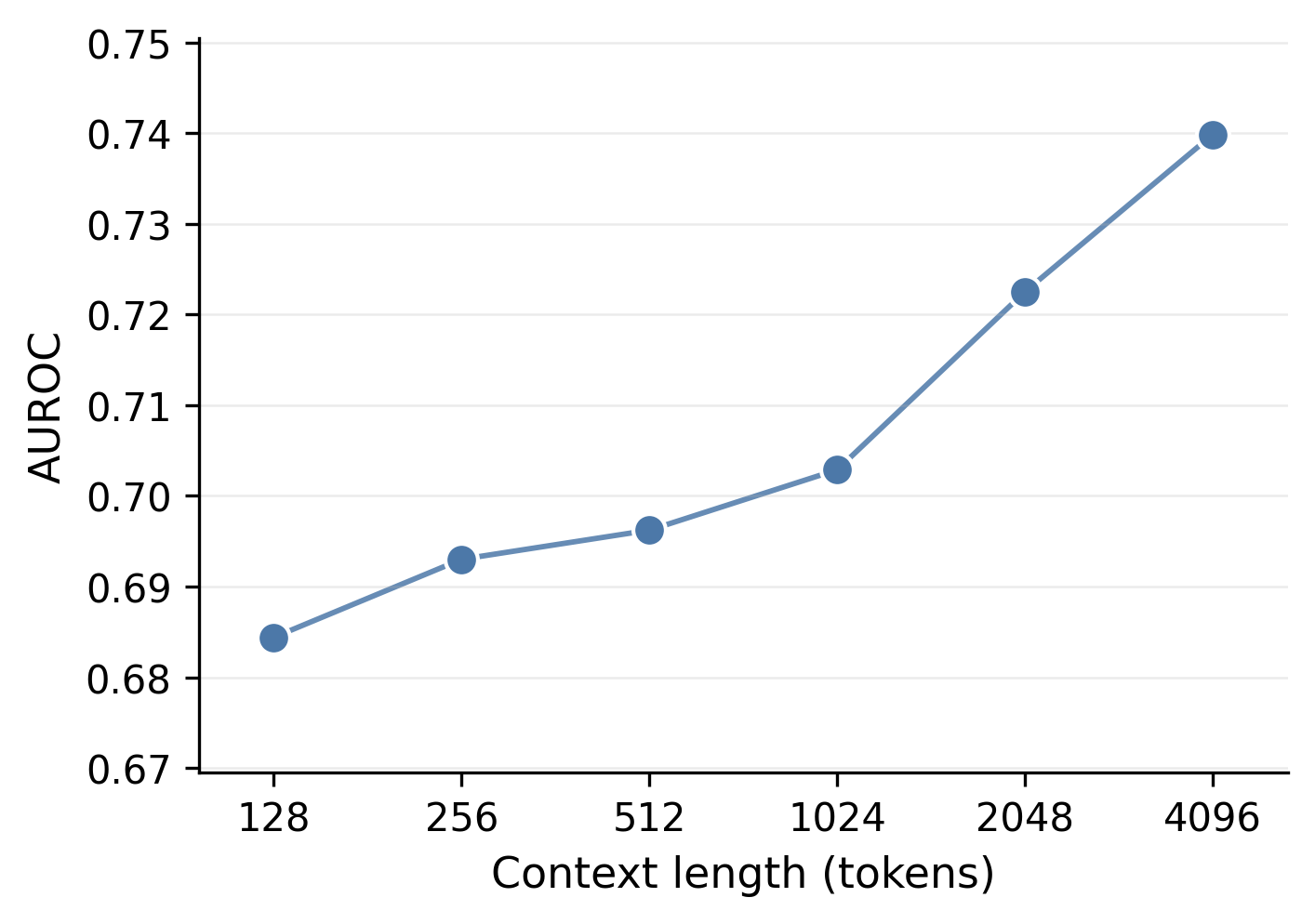}
    \caption{Moderate OSA}\label{fig:ctx-ecg-osa}
  \end{subfigure}\hfill
  \begin{subfigure}[t]{0.24\linewidth}
  \end{subfigure}
  \caption{\textbf{Effect of context length on ECG-only models.} As with EEG, perplexity and downstream metrics improve with longer context. The largest relative gains are again on summary tasks such as moderate OSA detection and CVD risk.\label{fig:ctx-ecg}}
\end{figure}

To quantify the effect of context length on Hypnos, we trained unimodal EEG and ECG variants of Hypnos-Small with context lengths $T \in \{128, 256, 512, 1024, 2048, 4096\}$ tokens, corresponding to roughly 2 minutes up to over an hour of data at 1\,Hz. Models were trained for 50k steps, whilst all other hyper-parameters were held fixed across runs. \Cref{fig:ctx-eeg,fig:ctx-ecg} show validation perplexity and downstream probing performance on the SHHS validation set as we vary $T$.

For both modalities, validation perplexity improves monotonically with context length, and most downstream metrics continue to improve out to 4096 tokens. Summary tasks benefit most: age regression, CVD risk and obstructive sleep apnoea (OSA) classification all improve steadily across the full range. Meanwhile, sleep staging and arousal detection saturate earlier, at around 1024--2048 tokens. These trends are consistent across EEG and ECG, suggesting that the benefit of longer context is not specific to a single modality.

\subsection{Training stability}\label{section:stability} To reduce compute usage, our main pre-training experiments were performed with a single random seed (42). We observed stable and predictable training dynamics across our training runs including our model scaling experiments. To further assess variance due to neural network training dynamics, we re-trained Hypnos-Small on single-channel EEG using three different random seeds. In \Cref{tab:training-stability} we see that the variance in validation loss and probing metrics is low.

\begin{table}[h]
  \centering
  \small
  \caption{\textbf{Training stability of Hypnos trained on single-channel EEG across three random seeds.} We report mean $\pm$ std and coefficient of variation (CV) across seeds. Downstream task performance is computed on validation data using an online linear probe.\label{tab:training-stability}}
  \begin{tabular}{lcc}
    \toprule
    Metric                                 & Mean $\pm$ std      & CV (\%) \\
    \midrule
    Validation perplexity (in-domain)      & $757.27 \pm 0.50$   & $0.066$ \\
    Validation perplexity (MrOS held-out)  & $876.73 \pm 0.68$   & $0.078$ \\
    \midrule
    Sleep staging $\kappa$ (SHHS)          & $0.7930 \pm 0.0004$ & $0.050$ \\
    Sleep staging $\kappa$ (MrOS held-out) & $0.7646 \pm 0.0013$ & $0.170$ \\
    \midrule
    Apnoea AUROC (SHHS)                    & $0.7825 \pm 0.0014$ & $0.179$ \\
    Apnoea AUROC (MrOS held-out)           & $0.7812 \pm 0.0095$ & $1.216$ \\
    \bottomrule
  \end{tabular}
\end{table}

\subsection{Modality-masking ablation}\label{app:modality-masking-ablation}
\Cref{tab:modality-masking-ablation-app} reports the full results of the modality-masking ablation summarised in \Cref{sec:modality-masking-ablation}, including the linear-probe head and the secondary metric per task. The overall ordering is consistent across both probes: \textit{Independent} is the weakest variant, \textit{No masking} closes most of the gap on \textit{Full}-modality evaluation but degrades on restricted-modality inputs, and \textit{Default} ($\alpha\!=\!1$ group masking) and \textit{Random} are comparable. All runs were trained on the same data splits using Hypnos-Small.

\begin{table}[h]
  \centering
  \scriptsize
  \setlength{\tabcolsep}{4pt}
  \caption{\textbf{Modality-masking ablation --- full results.} Linear and MLP probes on SHHS (in-domain). Per-subject mean. \textbf{Best} per (probe, subset, metric) column in bold (ties bolded both). Variants: \textit{Independent} ($\alpha\!\to\!\infty$, each modality stream trained independently), \textit{No masking} ($\alpha\!\to\!0$, always-full attention), \textit{Random} ($p\!=\!0.5$ per-modality random masking, following prior work), \textit{Default} (ours, Chinese Restaurant Process with $\alpha\!=\!1$; \Cref{sec:modality-masking}).\label{tab:modality-masking-ablation-app}}
  \resizebox{\textwidth}{!}{%
    \begin{tabular}{ll ccc cc cc cc}
      \toprule
                               &                                         & \multicolumn{3}{c}{Staging} & \multicolumn{2}{c}{Arousal} & \multicolumn{2}{c}{Apnoea} & \multicolumn{2}{c}{Desat.}                                                                                                \\
      \cmidrule(lr){3-5} \cmidrule(lr){6-7} \cmidrule(lr){8-9} \cmidrule(lr){10-11}
      Subset                   & Variant                                 & $\kappa$                    & AUROC                       & AUPRC                      & AUROC                      & AUPRC            & AUROC            & AUPRC            & AUROC            & AUPRC            \\
      \midrule
      \multicolumn{11}{l}{\textit{Linear probe}}                                                                                                                                                                                                                                              \\
      \hspace{1em}EEG-C3       & Independent ($\alpha\rightarrow\infty$) & $0.747$                     & $0.958$                     & $0.774$                    & $0.862$                    & $0.622$          & $0.771$          & $0.573$          & $0.766$          & $0.455$          \\
      \hspace{1em}($M{=}1$)    & No masking ($\alpha\rightarrow 0$)      & $0.755$                     & $0.960$                     & $0.784$                    & $0.884$                    & $0.670$          & $0.772$          & $0.572$          & $0.774$          & $0.474$          \\
                               & Random ($p=0.5$)                        & $0.773$                     & $0.965$                     & $0.797$                    & $0.896$                    & $0.691$          & $0.786$          & $0.597$          & $0.793$          & $\mathbf{0.508}$ \\
                               & Default ($\alpha=1$)                    & $\mathbf{0.774}$            & $\mathbf{0.966}$            & $\mathbf{0.800}$           & $\mathbf{0.898}$           & $\mathbf{0.697}$ & $\mathbf{0.788}$ & $\mathbf{0.599}$ & $\mathbf{0.795}$ & $0.507$          \\
      \midrule
      \hspace{1em}EOG          & Independent ($\alpha\rightarrow\infty$) & $0.750$                     & $0.958$                     & $0.779$                    & $0.859$                    & $0.601$          & $0.773$          & $0.588$          & $0.754$          & $0.441$          \\
      \hspace{1em}($M{=}2$)    & No masking ($\alpha\rightarrow 0$)      & $0.760$                     & $\mathbf{0.962}$            & $\mathbf{0.791}$           & $0.877$                    & $0.643$          & $0.776$          & $0.596$          & $0.745$          & $0.428$          \\
                               & Random ($p=0.5$)                        & $0.758$                     & $0.955$                     & $0.770$                    & $\mathbf{0.891}$           & $0.670$          & $0.791$          & $0.621$          & $\mathbf{0.795}$ & $\mathbf{0.513}$ \\
                               & Default ($\alpha=1$)                    & $\mathbf{0.764}$            & $0.957$                     & $0.778$                    & $0.888$                    & $\mathbf{0.671}$ & $\mathbf{0.793}$ & $\mathbf{0.622}$ & $0.794$          & $0.510$          \\
      \midrule
      \hspace{1em}Cardio-resp. & Independent ($\alpha\rightarrow\infty$) & $0.472$                     & $0.869$                     & $0.581$                    & $0.793$                    & $0.502$          & $0.851$          & $0.722$          & $\mathbf{0.810}$ & $\mathbf{0.533}$ \\
      \hspace{1em}($M{=}3$)    & No masking ($\alpha\rightarrow 0$)      & $0.463$                     & $0.869$                     & $0.575$                    & $0.831$                    & $0.565$          & $0.858$          & $0.731$          & $0.789$          & $0.490$          \\
                               & Random ($p=0.5$)                        & $0.486$                     & $0.876$                     & $0.588$                    & $0.833$                    & $0.569$          & $\mathbf{0.862}$ & $\mathbf{0.740}$ & $0.801$          & $0.512$          \\
                               & Default ($\alpha=1$)                    & $\mathbf{0.490}$            & $\mathbf{0.878}$            & $\mathbf{0.591}$           & $\mathbf{0.835}$           & $\mathbf{0.573}$ & $0.861$          & $0.737$          & $0.806$          & $0.522$          \\
      \midrule
      \hspace{1em}Full         & Independent ($\alpha\rightarrow\infty$) & $0.773$                     & $0.966$                     & $0.801$                    & $0.866$                    & $0.657$          & $0.847$          & $0.708$          & $0.817$          & $0.546$          \\
      \hspace{1em}($M{=}8$)    & No masking ($\alpha\rightarrow 0$)      & $0.792$                     & $0.970$                     & $0.816$                    & $0.900$                    & $0.710$          & $\mathbf{0.861}$ & $0.731$          & $0.830$          & $0.582$          \\
                               & Random ($p=0.5$)                        & $0.795$                     & $\mathbf{0.971}$            & $0.819$                    & $\mathbf{0.902}$           & $0.713$          & $0.860$          & $\mathbf{0.733}$ & $\mathbf{0.831}$ & $\mathbf{0.584}$ \\
                               & Default ($\alpha=1$)                    & $\mathbf{0.797}$            & $\mathbf{0.971}$            & $\mathbf{0.820}$           & $0.901$                    & $\mathbf{0.716}$ & $0.859$          & $0.730$          & $\mathbf{0.831}$ & $0.582$          \\
      \midrule[\heavyrulewidth]
      \multicolumn{11}{l}{\textit{MLP probe}}                                                                                                                                                                                                                                                 \\
      \hspace{1em}EEG-C3       & Independent ($\alpha\rightarrow\infty$) & $0.784$                     & $0.969$                     & $0.811$                    & $0.917$                    & $0.752$          & $0.788$          & $0.599$          & $0.788$          & $0.499$          \\
      \hspace{1em}($M{=}1$)    & No masking ($\alpha\rightarrow 0$)      & $0.786$                     & $0.970$                     & $0.813$                    & $0.924$                    & $0.771$          & $0.790$          & $0.606$          & $0.794$          & $0.510$          \\
                               & Random ($p=0.5$)                        & $\mathbf{0.796}$            & $\mathbf{0.972}$            & $\mathbf{0.823}$           & $0.932$                    & $0.790$          & $0.798$          & $0.617$          & $0.809$          & $0.537$          \\
                               & Default ($\alpha=1$)                    & $\mathbf{0.796}$            & $\mathbf{0.972}$            & $0.822$                    & $\mathbf{0.933}$           & $\mathbf{0.792}$ & $\mathbf{0.799}$ & $\mathbf{0.624}$ & $\mathbf{0.810}$ & $\mathbf{0.538}$ \\
      \midrule
      \hspace{1em}EOG          & Independent ($\alpha\rightarrow\infty$) & $0.781$                     & $0.969$                     & $0.808$                    & $0.903$                    & $0.712$          & $0.792$          & $0.616$          & $0.778$          & $0.487$          \\
      \hspace{1em}($M{=}2$)    & No masking ($\alpha\rightarrow 0$)      & $0.787$                     & $0.970$                     & $0.813$                    & $0.916$                    & $0.743$          & $0.794$          & $0.624$          & $0.768$          & $0.466$          \\
                               & Random ($p=0.5$)                        & $\mathbf{0.795}$            & $\mathbf{0.972}$            & $\mathbf{0.820}$           & $\mathbf{0.923}$           & $\mathbf{0.760}$ & $\mathbf{0.804}$ & $\mathbf{0.640}$ & $\mathbf{0.812}$ & $\mathbf{0.544}$ \\
                               & Default ($\alpha=1$)                    & $0.794$                     & $\mathbf{0.972}$            & $\mathbf{0.820}$           & $0.922$                    & $0.759$          & $0.803$          & $0.634$          & $0.810$          & $0.542$          \\
      \midrule
      \hspace{1em}Cardio-resp. & Independent ($\alpha\rightarrow\infty$) & $0.539$                     & $0.895$                     & $0.626$                    & $0.853$                    & $0.613$          & $0.869$          & $0.752$          & $0.832$          & $\mathbf{0.583}$ \\
      \hspace{1em}($M{=}3$)    & No masking ($\alpha\rightarrow 0$)      & $0.543$                     & $0.897$                     & $0.628$                    & $0.873$                    & $0.650$          & $0.879$          & $0.771$          & $0.815$          & $0.539$          \\
                               & Random ($p=0.5$)                        & $0.559$                     & $0.903$                     & $0.639$                    & $0.876$                    & $\mathbf{0.659}$ & $\mathbf{0.881}$ & $\mathbf{0.772}$ & $0.827$          & $0.564$          \\
                               & Default ($\alpha=1$)                    & $\mathbf{0.562}$            & $\mathbf{0.905}$            & $\mathbf{0.643}$           & $\mathbf{0.877}$           & $\mathbf{0.659}$ & $\mathbf{0.881}$ & $\mathbf{0.772}$ & $\mathbf{0.835}$ & $0.579$          \\
      \midrule
      \hspace{1em}Full         & Independent ($\alpha\rightarrow\infty$) & $0.800$                     & $0.974$                     & $0.826$                    & $0.931$                    & $0.784$          & $0.864$          & $0.737$          & $0.842$          & $0.598$          \\
      \hspace{1em}($M{=}8$)    & No masking ($\alpha\rightarrow 0$)      & $0.812$                     & $0.976$                     & $0.835$                    & $0.943$                    & $0.813$          & $\mathbf{0.880}$ & $0.766$          & $0.856$          & $0.630$          \\
                               & Random ($p=0.5$)                        & $\mathbf{0.814}$            & $\mathbf{0.977}$            & $\mathbf{0.839}$           & $\mathbf{0.944}$           & $0.815$          & $0.879$          & $\mathbf{0.768}$ & $\mathbf{0.860}$ & $\mathbf{0.636}$ \\
                               & Default ($\alpha=1$)                    & $0.813$                     & $\mathbf{0.977}$            & $0.837$                    & $\mathbf{0.944}$           & $\mathbf{0.816}$ & $\mathbf{0.880}$ & $0.767$          & $0.858$          & $0.631$          \\
      \bottomrule
    \end{tabular}%
  }
\end{table}

\subsection{Few-shot scaling across all datasets}\label{app:fewshot-all}
\Cref{fig:fewshot-staging} reports few-shot sleep-staging curves for one in-domain (SHHS) and one held-out (MrOS) dataset. For completeness, \Cref{tab:fewshot-all,tab:fewshot-all-mlp} reports the full per-dataset breakdown of sleep staging results across eight cohorts, three labelled-data fractions (1\%, 10\%, 100\%), and three metrics (Cohen's $\kappa$, macro-AUROC, macro-AUPRC) for linear and MLP probes. Hypnos consistently achieves the best macro-AUPRC in every cohort at every data fraction, and is best or near-best on $\kappa$ and AUROC throughout. Using 1\% of labelled data, Hypnos outperforms both U-Sleep and SleepTransformer using 100\% of the labelled data on 3 out of 4 held-out test sets evaluated.

\begin{table}[p]
  \centering
  \scriptsize
  \setlength{\tabcolsep}{4pt}
  \caption{\textbf{Few-shot sleep stage classification across all datasets --- linear probe.} Per-subject mean of Cohen's $\kappa$, macro-AUROC and macro-AUPRC, macro-averaged over the five AASM stages, as the fraction of labelled recordings used to train the probe is scaled from 1\% to 100\%. Foundation models are evaluated with a frozen encoder and a \textbf{linear probe}; U-Sleep and SleepTransformer are supervised (EEG+EOG), retrained at each fraction. \textbf{Best} per (fraction, metric) within each dataset block in bold.\label{tab:fewshot-all}}
  \begin{tabular}{ll ccc ccc ccc}
    \toprule
     & & \multicolumn{3}{c}{1\%} & \multicolumn{3}{c}{10\%} & \multicolumn{3}{c}{100\%} \\
    \cmidrule(lr){3-5} \cmidrule(lr){6-8} \cmidrule(lr){9-11}
    Dataset & Method & $\kappa$ & AUROC & AUPRC & $\kappa$ & AUROC & AUPRC & $\kappa$ & AUROC & AUPRC \\
    \midrule
    \multicolumn{11}{l}{\textit{In-domain}} \\
    \midrule
    \multirow{6}{*}{SHHS} & U-Sleep~\cite{perslevUSleepResilientHighfrequency2021} & $0.734$ & $0.916$ & $0.741$ & $0.790$ & $0.958$ & $0.815$ & $0.799$ & $0.969$ & $0.825$ \\
     & SleepTransformer~\cite{phanSleepTransformerAutomaticSleep2022} & $0.710$ & $0.947$ & $0.755$ & $0.782$ & $0.967$ & $0.805$ & $0.806$ & $0.974$ & $0.832$ \\
    \cmidrule[0.2pt](lr){2-11}
     & SleepFM~\cite{thapaMultimodalSleepFoundation2026} & $0.699$ & $0.942$ & $0.733$ & $0.668$ & $0.911$ & $0.671$ & $0.716$ & $0.941$ & $0.731$ \\
     & sleep2vec~\cite{yuanSleep2vecUnifiedCrossModal2026} & $0.681$ & $0.932$ & $0.716$ & $0.743$ & $0.958$ & $0.774$ & $0.763$ & $0.964$ & $0.792$ \\
     & OSF~\cite{shuaiOSFPretrainingScaling2026} & $0.754$ & $0.955$ & $0.773$ & $0.780$ & $0.964$ & $0.796$ & $0.791$ & $0.968$ & $0.809$ \\
     & Hypnos & $\mathbf{0.762}$ & $\mathbf{0.964}$ & $\mathbf{0.798}$ & $\mathbf{0.799}$ & $\mathbf{0.973}$ & $\mathbf{0.825}$ & $\mathbf{0.811}$ & $\mathbf{0.976}$ & $\mathbf{0.836}$ \\
    \cmidrule(lr){1-11}
    \multirow{6}{*}{CCSHS} & U-Sleep & $0.785$ & $0.931$ & $0.796$ & $0.843$ & $0.973$ & $0.874$ & $0.851$ & $0.981$ & $0.888$ \\
     & SleepTransformer & $0.776$ & $0.965$ & $0.819$ & $0.852$ & $0.980$ & $0.867$ & $0.873$ & $0.987$ & $0.896$ \\
    \cmidrule[0.2pt](lr){2-11}
     & SleepFM & $0.776$ & $0.960$ & $0.801$ & $0.742$ & $0.934$ & $0.738$ & $0.784$ & $0.958$ & $0.797$ \\
     & sleep2vec & $0.771$ & $0.959$ & $0.798$ & $0.819$ & $0.975$ & $0.844$ & $0.833$ & $0.980$ & $0.862$ \\
     & OSF & $0.827$ & $0.973$ & $0.842$ & $0.842$ & $0.979$ & $0.859$ & $0.849$ & $0.983$ & $0.872$ \\
     & Hypnos & $\mathbf{0.843}$ & $\mathbf{0.980}$ & $\mathbf{0.871}$ & $\mathbf{0.875}$ & $\mathbf{0.988}$ & $\mathbf{0.893}$ & $\mathbf{0.881}$ & $\mathbf{0.990}$ & $\mathbf{0.901}$ \\
    \cmidrule(lr){1-11}
    \multirow{6}{*}{CFS} & U-Sleep & $0.743$ & $0.918$ & $0.756$ & $0.808$ & $0.961$ & $0.825$ & $0.803$ & $0.971$ & $0.845$ \\
     & SleepTransformer & $0.760$ & $0.958$ & $0.798$ & $0.779$ & $0.962$ & $0.816$ & $0.804$ & $0.973$ & $0.847$ \\
    \cmidrule[0.2pt](lr){2-11}
     & SleepFM & $0.705$ & $0.937$ & $0.735$ & $0.662$ & $0.906$ & $0.677$ & $0.725$ & $0.933$ & $0.738$ \\
     & sleep2vec & $0.738$ & $0.940$ & $0.757$ & $0.792$ & $0.962$ & $0.808$ & $0.804$ & $0.966$ & $0.820$ \\
     & OSF & $0.757$ & $0.954$ & $0.784$ & $0.784$ & $0.962$ & $0.802$ & $0.805$ & $0.968$ & $0.816$ \\
     & Hypnos & $\mathbf{0.785}$ & $\mathbf{0.967}$ & $\mathbf{0.829}$ & $\mathbf{0.824}$ & $\mathbf{0.979}$ & $\mathbf{0.858}$ & $\mathbf{0.839}$ & $\mathbf{0.983}$ & $\mathbf{0.873}$ \\
    \cmidrule(lr){1-11}
    \multirow{6}{*}{NCHSDB} & U-Sleep & $0.626$ & $0.850$ & $0.671$ & $0.718$ & $0.939$ & $0.774$ & $0.733$ & $0.951$ & $0.794$ \\
     & SleepTransformer & $0.618$ & $0.925$ & $0.731$ & $0.740$ & $0.948$ & $0.783$ & $0.773$ & $0.963$ & $0.816$ \\
    \cmidrule[0.2pt](lr){2-11}
     & SleepFM & $0.595$ & $0.909$ & $0.692$ & $0.570$ & $0.880$ & $0.634$ & $0.639$ & $0.918$ & $0.707$ \\
     & sleep2vec & $0.632$ & $0.915$ & $0.708$ & $0.694$ & $0.941$ & $0.761$ & $0.713$ & $0.948$ & $0.774$ \\
     & OSF & $0.609$ & $0.909$ & $0.715$ & $0.674$ & $0.932$ & $0.749$ & $0.704$ & $0.942$ & $0.764$ \\
     & Hypnos & $\mathbf{0.723}$ & $\mathbf{0.948}$ & $\mathbf{0.784}$ & $\mathbf{0.767}$ & $\mathbf{0.964}$ & $\mathbf{0.817}$ & $\mathbf{0.777}$ & $\mathbf{0.967}$ & $\mathbf{0.826}$ \\
    \midrule
    \multicolumn{11}{l}{\textit{Held-out}} \\
    \midrule
    \multirow{6}{*}{MROS} & U-Sleep & $0.592$ & $0.872$ & $0.621$ & $0.737$ & $0.934$ & $0.730$ & $0.730$ & $0.949$ & $0.748$ \\
     & SleepTransformer & $0.673$ & $0.938$ & $0.699$ & $0.714$ & $0.943$ & $0.724$ & $0.753$ & $0.958$ & $0.758$ \\
    \cmidrule[0.2pt](lr){2-11}
     & SleepFM & $0.622$ & $0.918$ & $0.638$ & $0.601$ & $0.889$ & $0.604$ & $0.658$ & $0.919$ & $0.659$ \\
     & sleep2vec & $0.656$ & $0.925$ & $0.664$ & $0.734$ & $0.951$ & $0.719$ & $0.751$ & $0.958$ & $0.737$ \\
     & OSF & $0.698$ & $0.940$ & $0.702$ & $0.707$ & $0.948$ & $0.723$ & $0.740$ & $0.957$ & $0.738$ \\
     & Hypnos & $\mathbf{0.701}$ & $\mathbf{0.944}$ & $\mathbf{0.718}$ & $\mathbf{0.766}$ & $\mathbf{0.968}$ & $\mathbf{0.778}$ & $\mathbf{0.799}$ & $\mathbf{0.973}$ & $\mathbf{0.794}$ \\
    \cmidrule(lr){1-11}
    \multirow{6}{*}{DOD-H} & U-Sleep & $0.622$ & $0.886$ & $0.729$ & $0.721$ & $0.952$ & $0.818$ & $0.753$ & $0.956$ & $0.830$ \\
     & SleepTransformer & $0.628$ & $0.952$ & $0.787$ & $0.757$ & $0.967$ & $0.846$ & $0.785$ & $0.972$ & $0.861$ \\
    \cmidrule[0.2pt](lr){2-11}
     & SleepFM & $0.478$ & $0.911$ & $0.717$ & $0.402$ & $0.861$ & $0.617$ & $0.635$ & $0.925$ & $0.740$ \\
     & sleep2vec & $0.626$ & $0.924$ & $0.742$ & $0.701$ & $0.955$ & $0.810$ & $0.724$ & $0.963$ & $0.837$ \\
     & OSF & $0.690$ & $0.951$ & $0.808$ & $0.639$ & $0.953$ & $0.821$ & $0.679$ & $0.957$ & $0.833$ \\
     & Hypnos & $\mathbf{0.767}$ & $\mathbf{0.970}$ & $\mathbf{0.866}$ & $\mathbf{0.814}$ & $\mathbf{0.977}$ & $\mathbf{0.889}$ & $\mathbf{0.805}$ & $\mathbf{0.979}$ & $\mathbf{0.898}$ \\
    \cmidrule(lr){1-11}
    \multirow{6}{*}{DOD-O} & U-Sleep & $0.602$ & $0.873$ & $0.701$ & $0.736$ & $0.960$ & $0.813$ & $\mathbf{0.766}$ & $\mathbf{0.968}$ & $0.829$ \\
     & SleepTransformer & $0.295$ & $0.894$ & $0.610$ & $0.554$ & $0.937$ & $0.730$ & $0.720$ & $0.956$ & $0.795$ \\
    \cmidrule[0.2pt](lr){2-11}
     & SleepFM & $0.386$ & $0.891$ & $0.666$ & $0.409$ & $0.856$ & $0.593$ & $0.612$ & $0.917$ & $0.703$ \\
     & sleep2vec & $0.607$ & $0.912$ & $0.702$ & $0.679$ & $0.943$ & $0.770$ & $0.693$ & $0.950$ & $0.785$ \\
     & OSF & $0.645$ & $0.945$ & $0.768$ & $0.656$ & $0.954$ & $0.784$ & $0.717$ & $0.960$ & $0.801$ \\
     & Hypnos & $\mathbf{0.746}$ & $\mathbf{0.967}$ & $\mathbf{0.828}$ & $\mathbf{0.747}$ & $\mathbf{0.972}$ & $\mathbf{0.847}$ & $0.721$ & $\mathbf{0.968}$ & $\mathbf{0.847}$ \\
    \cmidrule(lr){1-11}
    \multirow{6}{*}{MESA} & U-Sleep & $0.576$ & $0.863$ & $0.604$ & $0.640$ & $0.904$ & $0.683$ & $0.655$ & $0.931$ & $0.712$ \\
     & SleepTransformer & $0.475$ & $0.882$ & $0.633$ & $0.663$ & $0.940$ & $0.726$ & $0.661$ & $0.942$ & $0.740$ \\
    \cmidrule[0.2pt](lr){2-11}
     & SleepFM & $0.544$ & $0.890$ & $0.619$ & $0.526$ & $0.879$ & $0.585$ & $0.562$ & $0.904$ & $0.640$ \\
     & sleep2vec & $0.628$ & $0.914$ & $0.664$ & $0.684$ & $0.939$ & $0.716$ & $0.695$ & $0.946$ & $0.731$ \\
     & OSF & $0.623$ & $0.917$ & $0.667$ & $0.637$ & $0.926$ & $0.685$ & $0.652$ & $0.936$ & $0.700$ \\
     & Hypnos & $\mathbf{0.693}$ & $\mathbf{0.948}$ & $\mathbf{0.744}$ & $\mathbf{0.747}$ & $\mathbf{0.959}$ & $\mathbf{0.773}$ & $\mathbf{0.750}$ & $\mathbf{0.964}$ & $\mathbf{0.782}$ \\
    \bottomrule
  \end{tabular}
\end{table}

\begin{table}[p]
  \centering
  \scriptsize
  \setlength{\tabcolsep}{4pt}
  \caption{\textbf{Few-shot sleep stage classification across all datasets --- MLP probe.} Per-subject mean of Cohen's $\kappa$, macro-AUROC and macro-AUPRC, macro-averaged over the five AASM stages, as the fraction of labelled recordings used to train the probe is scaled from 1\% to 100\%. Foundation models are evaluated with a frozen encoder and a \textbf{MLP probe}; U-Sleep and SleepTransformer are supervised (EEG+EOG), retrained at each fraction. \textbf{Best} per (fraction, metric) within each dataset block in bold.\label{tab:fewshot-all-mlp}}
  \begin{tabular}{ll ccc ccc ccc}
    \toprule
     & & \multicolumn{3}{c}{1\%} & \multicolumn{3}{c}{10\%} & \multicolumn{3}{c}{100\%} \\
    \cmidrule(lr){3-5} \cmidrule(lr){6-8} \cmidrule(lr){9-11}
    Dataset & Method & $\kappa$ & AUROC & AUPRC & $\kappa$ & AUROC & AUPRC & $\kappa$ & AUROC & AUPRC \\
    \midrule
    \multicolumn{11}{l}{\textit{In-domain}} \\
    \midrule
    \multirow{6}{*}{SHHS} & U-Sleep~\cite{perslevUSleepResilientHighfrequency2021} & $0.734$ & $0.916$ & $0.741$ & $0.790$ & $0.958$ & $0.815$ & $0.799$ & $0.969$ & $0.825$ \\
     & SleepTransformer~\cite{phanSleepTransformerAutomaticSleep2022} & $0.710$ & $0.947$ & $0.755$ & $0.782$ & $0.967$ & $0.805$ & $0.806$ & $0.974$ & $0.832$ \\
    \cmidrule[0.2pt](lr){2-11}
     & SleepFM~\cite{thapaMultimodalSleepFoundation2026} & $0.702$ & $0.943$ & $0.735$ & $0.748$ & $0.957$ & $0.772$ & $0.772$ & $0.964$ & $0.791$ \\
     & sleep2vec~\cite{yuanSleep2vecUnifiedCrossModal2026} & $0.699$ & $0.940$ & $0.725$ & $0.741$ & $0.957$ & $0.767$ & $0.771$ & $0.966$ & $0.799$ \\
     & OSF~\cite{shuaiOSFPretrainingScaling2026} & $0.762$ & $0.959$ & $0.777$ & $0.784$ & $0.968$ & $0.806$ & $0.800$ & $0.971$ & $0.818$ \\
     & Hypnos & $\mathbf{0.785}$ & $\mathbf{0.971}$ & $\mathbf{0.821}$ & $\mathbf{0.807}$ & $\mathbf{0.975}$ & $\mathbf{0.832}$ & $\mathbf{0.819}$ & $\mathbf{0.978}$ & $\mathbf{0.844}$ \\
    \cmidrule(lr){1-11}
    \multirow{6}{*}{CCSHS} & U-Sleep & $0.785$ & $0.931$ & $0.796$ & $0.843$ & $0.973$ & $0.874$ & $0.851$ & $0.981$ & $0.888$ \\
     & SleepTransformer & $0.776$ & $0.965$ & $0.819$ & $0.852$ & $0.980$ & $0.867$ & $0.873$ & $0.987$ & $0.896$ \\
    \cmidrule[0.2pt](lr){2-11}
     & SleepFM & $0.778$ & $0.963$ & $0.808$ & $0.831$ & $0.977$ & $0.850$ & $0.849$ & $0.983$ & $0.869$ \\
     & sleep2vec & $0.773$ & $0.961$ & $0.793$ & $0.813$ & $0.975$ & $0.838$ & $0.839$ & $0.982$ & $0.868$ \\
     & OSF & $0.829$ & $0.976$ & $0.843$ & $0.844$ & $0.982$ & $0.872$ & $0.863$ & $0.985$ & $0.884$ \\
     & Hypnos & $\mathbf{0.856}$ & $\mathbf{0.984}$ & $\mathbf{0.880}$ & $\mathbf{0.871}$ & $\mathbf{0.988}$ & $\mathbf{0.894}$ & $\mathbf{0.888}$ & $\mathbf{0.991}$ & $\mathbf{0.914}$ \\
    \cmidrule(lr){1-11}
    \multirow{6}{*}{CFS} & U-Sleep & $0.743$ & $0.918$ & $0.756$ & $0.808$ & $0.961$ & $0.825$ & $0.803$ & $0.971$ & $0.845$ \\
     & SleepTransformer & $0.760$ & $0.958$ & $0.798$ & $0.779$ & $0.962$ & $0.816$ & $0.804$ & $0.973$ & $0.847$ \\
    \cmidrule[0.2pt](lr){2-11}
     & SleepFM & $0.704$ & $0.938$ & $0.741$ & $0.740$ & $0.953$ & $0.783$ & $0.783$ & $0.965$ & $0.812$ \\
     & sleep2vec & $0.748$ & $0.944$ & $0.755$ & $0.778$ & $0.960$ & $0.794$ & $0.810$ & $0.970$ & $0.828$ \\
     & OSF & $0.774$ & $0.959$ & $0.790$ & $0.796$ & $0.968$ & $0.816$ & $0.813$ & $0.972$ & $0.829$ \\
     & Hypnos & $\mathbf{0.793}$ & $\mathbf{0.974}$ & $\mathbf{0.843}$ & $\mathbf{0.821}$ & $\mathbf{0.979}$ & $\mathbf{0.865}$ & $\mathbf{0.838}$ & $\mathbf{0.983}$ & $\mathbf{0.880}$ \\
    \cmidrule(lr){1-11}
    \multirow{6}{*}{NCHSDB} & U-Sleep & $0.626$ & $0.850$ & $0.671$ & $0.718$ & $0.939$ & $0.774$ & $0.733$ & $0.951$ & $0.794$ \\
     & SleepTransformer & $0.618$ & $0.925$ & $0.731$ & $0.740$ & $0.948$ & $0.783$ & $0.773$ & $0.963$ & $0.816$ \\
    \cmidrule[0.2pt](lr){2-11}
     & SleepFM & $0.610$ & $0.919$ & $0.708$ & $0.680$ & $0.940$ & $0.755$ & $0.721$ & $0.950$ & $0.778$ \\
     & sleep2vec & $0.628$ & $0.921$ & $0.710$ & $0.687$ & $0.940$ & $0.753$ & $0.733$ & $0.954$ & $0.787$ \\
     & OSF & $0.630$ & $0.914$ & $0.723$ & $0.689$ & $0.940$ & $0.762$ & $0.719$ & $0.947$ & $0.775$ \\
     & Hypnos & $\mathbf{0.749}$ & $\mathbf{0.960}$ & $\mathbf{0.807}$ & $\mathbf{0.770}$ & $\mathbf{0.967}$ & $\mathbf{0.822}$ & $\mathbf{0.782}$ & $\mathbf{0.969}$ & $\mathbf{0.833}$ \\
    \midrule
    \multicolumn{11}{l}{\textit{Held-out}} \\
    \midrule
    \multirow{6}{*}{MROS} & U-Sleep & $0.592$ & $0.872$ & $0.621$ & $0.737$ & $0.934$ & $0.730$ & $0.730$ & $0.949$ & $0.748$ \\
     & SleepTransformer & $0.673$ & $0.938$ & $0.699$ & $0.714$ & $0.943$ & $0.724$ & $0.753$ & $0.958$ & $0.758$ \\
    \cmidrule[0.2pt](lr){2-11}
     & SleepFM & $0.660$ & $0.934$ & $0.675$ & $0.686$ & $0.943$ & $0.696$ & $0.731$ & $0.952$ & $0.718$ \\
     & sleep2vec & $0.663$ & $0.930$ & $0.668$ & $0.720$ & $0.947$ & $0.701$ & $0.758$ & $0.960$ & $0.741$ \\
     & OSF & $0.716$ & $0.946$ & $0.709$ & $0.733$ & $0.956$ & $0.735$ & $0.750$ & $0.960$ & $0.747$ \\
     & Hypnos & $\mathbf{0.746}$ & $\mathbf{0.961}$ & $\mathbf{0.763}$ & $\mathbf{0.767}$ & $\mathbf{0.968}$ & $\mathbf{0.776}$ & $\mathbf{0.805}$ & $\mathbf{0.973}$ & $\mathbf{0.796}$ \\
    \cmidrule(lr){1-11}
    \multirow{6}{*}{DOD-H} & U-Sleep & $0.622$ & $0.886$ & $0.729$ & $0.721$ & $0.952$ & $0.818$ & $0.753$ & $0.956$ & $0.830$ \\
     & SleepTransformer & $0.628$ & $0.952$ & $0.787$ & $0.757$ & $0.967$ & $0.846$ & $0.785$ & $0.972$ & $0.861$ \\
    \cmidrule[0.2pt](lr){2-11}
     & SleepFM & $0.531$ & $0.911$ & $0.720$ & $0.684$ & $0.949$ & $0.792$ & $0.647$ & $0.943$ & $0.788$ \\
     & sleep2vec & $0.639$ & $0.934$ & $0.746$ & $0.679$ & $0.950$ & $0.797$ & $0.745$ & $0.963$ & $0.836$ \\
     & OSF & $0.698$ & $0.954$ & $0.815$ & $0.723$ & $0.959$ & $0.829$ & $0.700$ & $0.955$ & $0.820$ \\
     & Hypnos & $\mathbf{0.788}$ & $\mathbf{0.975}$ & $\mathbf{0.882}$ & $\mathbf{0.786}$ & $\mathbf{0.976}$ & $\mathbf{0.889}$ & $\mathbf{0.820}$ & $\mathbf{0.982}$ & $\mathbf{0.906}$ \\
    \cmidrule(lr){1-11}
    \multirow{6}{*}{DOD-O} & U-Sleep & $0.602$ & $0.873$ & $0.701$ & $\mathbf{0.736}$ & $0.960$ & $0.813$ & $\mathbf{0.766}$ & $\mathbf{0.968}$ & $0.829$ \\
     & SleepTransformer & $0.295$ & $0.894$ & $0.610$ & $0.554$ & $0.937$ & $0.730$ & $0.720$ & $0.956$ & $0.795$ \\
    \cmidrule[0.2pt](lr){2-11}
     & SleepFM & $0.549$ & $0.911$ & $0.700$ & $0.541$ & $0.919$ & $0.708$ & $0.581$ & $0.926$ & $0.731$ \\
     & sleep2vec & $0.615$ & $0.919$ & $0.707$ & $0.660$ & $0.940$ & $0.758$ & $0.707$ & $0.952$ & $0.788$ \\
     & OSF & $0.722$ & $0.953$ & $0.785$ & $0.694$ & $0.954$ & $0.792$ & $0.703$ & $0.953$ & $0.796$ \\
     & Hypnos & $\mathbf{0.742}$ & $\mathbf{0.966}$ & $\mathbf{0.833}$ & $0.689$ & $\mathbf{0.962}$ & $\mathbf{0.835}$ & $0.733$ & $0.966$ & $\mathbf{0.843}$ \\
    \cmidrule(lr){1-11}
    \multirow{6}{*}{MESA} & U-Sleep & $0.576$ & $0.863$ & $0.604$ & $0.640$ & $0.904$ & $0.683$ & $0.655$ & $0.931$ & $0.712$ \\
     & SleepTransformer & $0.475$ & $0.882$ & $0.633$ & $0.663$ & $0.940$ & $0.726$ & $0.661$ & $0.942$ & $0.740$ \\
    \cmidrule[0.2pt](lr){2-11}
     & SleepFM & $0.572$ & $0.909$ & $0.647$ & $0.586$ & $0.926$ & $0.671$ & $0.629$ & $0.933$ & $0.697$ \\
     & sleep2vec & $0.626$ & $0.919$ & $0.667$ & $0.669$ & $0.935$ & $0.699$ & $\mathbf{0.705}$ & $0.947$ & $0.732$ \\
     & OSF & $0.639$ & $0.925$ & $0.683$ & $0.626$ & $0.931$ & $0.690$ & $0.666$ & $0.937$ & $0.703$ \\
     & Hypnos & $\mathbf{0.690}$ & $\mathbf{0.956}$ & $\mathbf{0.757}$ & $\mathbf{0.702}$ & $\mathbf{0.958}$ & $\mathbf{0.763}$ & $0.692$ & $\mathbf{0.956}$ & $\mathbf{0.754}$ \\
    \bottomrule
  \end{tabular}
\end{table}

\clearpage
\subsection{Transfer to external single-lead ECG benchmarks}\label{app:external-ecg}
We evaluate Hypnos's frozen ECG-only embeddings against xECG~\cite{lunelliBenchECGXECGBenchmark2025}, a foundation model pretrained on 12-lead clinical ECG, on three external single-lead ECG benchmarks (\Cref{tab:external-ecg-app}). Neither model's pretraining corpus contains any of these datasets. Both models use a frozen encoder followed by a \texttt{StandardScaler} + L2-regularised logistic-regression probe, with $C$ tuned on validation AUROC over $\{10^{-4},\ldots,10^{2}\}$.

\textbf{Apnea-ECG protocol.} We follow BenchECG's \texttt{sleep\_apnea} 5-way ensemble linear-probe recipe: each sample is a 300\,s window centred on a target minute; the backbone produces per-minute features for five consecutive positions ($m{-}2,\ldots,m{+}2$); a linear head is trained on each position supervised by the central-minute label; the five predictions are averaged at test time.

\begin{table}[!htbp]
  \centering
  \small
  \setlength{\tabcolsep}{6pt}
  \caption{\textbf{External single-lead ECG benchmarks.} Frozen-encoder linear-probe AUROC across three publicly available external single-lead ECG datasets covering handheld (CinC 2017), overnight single-lead (Apnea-ECG), and dynamic Holter (CPSC 2021) settings. CinC 2017 reports 5-fold CV mean; other datasets report a single subject-disjoint split.\label{tab:external-ecg-app}}
  \begin{tabular}{l l c c}
    \toprule
    Dataset   & Task / eval unit             & Hypnos           & xECG             \\
    \midrule
    CinC 2017 & AF-vs-rest (per-record)      & $\mathbf{0.984}$ & $\mathbf{0.985}$ \\
    Apnea-ECG & Apnoea (per-minute)          & $\mathbf{0.925}$ & $0.884$          \\
    CPSC 2021 & Paroxysmal AF (30\,s window) & $\mathbf{0.985}$ & $0.934$          \\
    \bottomrule
  \end{tabular}
\end{table}

\subsection{Learning objective}\label{app:objective}

  The comparisons in \Cref{sec:experiments} differ in objective, architecture, context length and aggregation, so they establish overall model performance rather than the effect of the learning objective. Here we compare next-token prediction (NTP) directly with two alternatives.

  \paragraph{Masked-token prediction} We trained a BERT-style bi-directional masked-token prediction (MTP) model, tuning the masking ratio over $\{25, 50, 75\}\%$ and selecting $50\%$. \Cref{tab:ntp-mtp} shows that the two objectives achieve very similar downstream performance. The advantage of next-token prediction is practical: a single architecture supports both representation learning and auto-regressive generation with no train-test shift, and permits sequential state updates for streaming inference.

  \paragraph{Input-space masked autoencoding} We compared NTP with an input-space masked autoencoder (MAE) using unimodal EEG models with matched Transformer widths, context length and pre-training compute, tuning the learning rate per configuration. \Cref{tab:ntp-mae} shows that MAE is weaker at every scale.

\begin{table}[h]
  \centering
  \small
  \setlength{\tabcolsep}{4pt}
  \caption{\textbf{Next-token vs.\ masked-token prediction.} Hypnos-Small trained on all eight modalities with a context length of 512, using the same tokenizers and depth head. MTP uses a bidirectional temporal transformer with 50\% of timesteps masked, shared across modalities. Linear probes trained and evaluated on SHHS.\label{tab:ntp-mtp}}
  \begin{tabular}{l cc cc cc cc}
    \toprule
              & \multicolumn{2}{c}{Staging} & \multicolumn{2}{c}{Arousal} & \multicolumn{2}{c}{Apnoea} & \multicolumn{2}{c}{Desat.}                                         \\
    \cmidrule(lr){2-3} \cmidrule(lr){4-5} \cmidrule(lr){6-7} \cmidrule(lr){8-9}
    Objective & $\kappa$                    & AUROC                       & AUROC                      & AUPRC                      & AUROC   & AUPRC   & AUROC   & AUPRC   \\
    \midrule
    MTP       & $0.813$                     & $0.971$                     & $0.926$                    & $0.754$                    & $0.869$ & $0.785$ & $0.829$ & $0.590$ \\
    NTP       & $0.817$                     & $0.974$                     & $0.920$                    & $0.737$                    & $0.864$ & $0.777$ & $0.837$ & $0.612$ \\
    \bottomrule
  \end{tabular}
\end{table}

\begin{table}[h]
  \centering
  \small
  \caption{\textbf{Input-space MAE vs.\ NTP across model scale.} Sleep staging $\kappa$ (MLP probe) on held-out MrOS using single-channel EEG (C3).\label{tab:ntp-mae}}
  \begin{tabular}{l c c c}
    \toprule
    Model size & Width & MAE     & NTP     \\
    \midrule
    Tiny       & 192   & $0.649$ & $0.766$ \\
    Small      & 384   & $0.657$ & $0.783$ \\
    Base       & 768   & $0.725$ & $0.795$ \\
    \bottomrule
  \end{tabular}
\end{table}

\subsection{Model size and compute}\label{app:compute}

  \Cref{tab:matched-compute,tab:sleep2vec-scale} compare Hypnos with our re-trained OSF and sleep2vec respectively, using matched Transformer backbones, identical dataset splits and identical pre-processing. Encoder parameter counts exclude contrastive projection heads for sleep2vec, student parameters for OSF and discrete output codebooks for Hypnos. At each scale, Hypnos and OSF are trained with compute matched to within $10\%$. Hypnos achieves higher performance on all four tasks at every scale. Re-training also improves on the released OSF weights on some metrics (e.g.\ MrOS $\kappa$ of $0.785$ for OSF-Base vs $0.750$).

  Our sleep2vec re-implementation uses far less compute (around $10^{18}$ FLOPs for the Medium configuration). Scaling its compute beyond the original recipe, via a $4\times$ larger batch size with a tuned learning rate and up to 80k training steps, gave only a minor improvement (MrOS MLP $\kappa$ from $0.757$ to $0.765$).

\begin{table}[h]
  \centering
  \small
  \setlength{\tabcolsep}{4pt}
  \caption{\textbf{Hypnos vs.\ OSF at matched backbones and compute.} MLP probe on held-out MrOS. \textit{Enc} is the encoder parameter count. All OSF results in this table use our re-trained models, rather than the released weights used in \Cref{sec:experiments}. Results are computed on the subset of recordings common to OSF, sleep2vec and Hypnos, so may differ slightly from \Cref{tab:robustness}, which uses the subset common to all models in that table.\label{tab:matched-compute}}
  \resizebox{\textwidth}{!}{%
    \begin{tabular}{l l r c c c c c}
      \toprule
      Model        & Backbone & Enc (M) & FLOPs               & Staging $\kappa$ & Apnoea AUROC & Arousal AUROC & Desat.\ AUROC \\
      \midrule
      OSF-Tiny     & 12L/192d & $5.4$   & $7.5\times 10^{18}$ & $0.709$          & $0.761$      & $0.904$       & $0.704$       \\
      Hypnos-Tiny  & 12L/192d & $19.7$  & $8.3\times 10^{18}$ & $0.771$          & $0.780$      & $0.912$       & $0.736$       \\
      \midrule
      OSF-Small    & 12L/384d & $21.4$  & $2.5\times 10^{19}$ & $0.766$          & $0.768$      & $0.909$       & $0.728$       \\
      Hypnos-Small & 12L/384d & $53.6$  & $2.6\times 10^{19}$ & $0.793$          & $0.786$      & $0.918$       & $0.756$       \\
      \midrule
      OSF-Base     & 12L/768d & $85.3$  & $9.3\times 10^{19}$ & $0.785$          & $0.791$      & $0.918$       & $0.738$       \\
      Hypnos-Base  & 12L/768d & $163.8$ & $1.0\times 10^{20}$ & $0.815$          & $0.808$      & $0.934$       & $0.767$       \\
      \bottomrule
    \end{tabular}%
  }
\end{table}

\begin{table}[h]
  \centering
  \small
  \setlength{\tabcolsep}{4pt}
  \caption{\textbf{Hypnos vs.\ sleep2vec at matched backbones.} MLP probe on held-out MrOS, on the same recordings as \Cref{tab:matched-compute}. \textit{Enc} is the encoder parameter count.\label{tab:sleep2vec-scale}}
  \resizebox{\textwidth}{!}{%
    \begin{tabular}{l l r c c c c}
      \toprule
      Model            & Backbone  & Enc (M) & Staging $\kappa$ & Apnoea AUROC & Arousal AUROC & Desat.\ AUROC \\
      \midrule
      sleep2vec-Tiny   & 12L/192d  & $19.0$  & $0.679$          & $0.723$      & $0.829$       & $0.700$       \\
      Hypnos-Tiny      & 12L/192d  & $19.7$  & $0.771$          & $0.780$      & $0.912$       & $0.736$       \\
      \midrule
      sleep2vec-Small  & 12L/384d  & $49.3$  & $0.732$          & $0.748$      & $0.859$       & $0.710$       \\
      Hypnos-Small     & 12L/384d  & $53.6$  & $0.793$          & $0.786$      & $0.918$       & $0.756$       \\
      \midrule
      sleep2vec-Medium & 12L/768d  & $144.1$ & $0.757$          & $0.765$      & $0.873$       & $0.733$       \\
      Hypnos-Base      & 12L/768d  & $163.8$ & $0.815$          & $0.808$      & $0.934$       & $0.767$       \\
      \midrule
      sleep2vec-Large  & 16L/1024d & $282.8$ & $0.766$          & $0.777$      & $0.883$       & $0.750$       \\
      \bottomrule
    \end{tabular}%
  }
\end{table}

\subsection{Comparison with SleepMaMi}\label{app:sleepmami}

  SleepMaMi~\cite{parkSleepMaMiUniversalSleep2026} trains a micro-encoder with private and shared components using contrastive, input-space reconstruction and KoLeo losses, followed by a macro-encoder over the full night using demographic-guided contrastive learning. Both models therefore separate local learning from long-range learning, whereas Hypnos requires no patient metadata and uses causal components throughout.

  We compare the two models under linear probing on CHAT and MESA, generating SleepMaMi embeddings with its public weights, pre-processing and inference code, on the matched subset of recordings with the signals it requires. We excluded datasets such as SHHS because they appear in both pre-training corpora under different splits, and DOD-H and DOD-O because they do not record SpO$_2$, which SleepMaMi requires.

  In \Cref{tab:sleepmami}, Hypnos performs better on sleep staging and arousal detection. SleepMaMi performs better on desaturation detection, where it receives SpO$_2$ directly while Hypnos must infer desaturation from indirect signals such as ECG and respiratory effort, so this comparison is not modality-matched. Apnoea results are mixed: Hypnos is better on CHAT AUROC, SleepMaMi is better on MESA AUPRC, and the remaining differences are not significant.

  We also compare per-second detection of arousal and apnoea events in \Cref{tab:sleepmami-events}, reporting per-second macro-F1 and event-level F1, which rewards detecting whole events rather than individual seconds~\cite{mesarosMetricsPolyphonicSound2016,warbySleepspindleDetectionCrowdsourcing2014}. Hypnos leads on event-level F1 for both tasks on both cohorts, at an IoU of $0.3$ and at all thresholds from $0.1$ to $0.5$. Per-second F1 shows the same pattern, except for apnoea on CHAT, where the difference is not significant. SleepMaMi reports no evaluation under missing modalities, so we do not compare robustness.

\begin{table}[h]
  \centering
  \small
  \setlength{\tabcolsep}{4pt}
  \caption{\textbf{Hypnos vs.\ SleepMaMi.} Linear probe on a matched subset of recordings. $^{*}$ indicates a significant difference (Wilcoxon signed-rank test with Benjamini--Hochberg FDR correction, $q<0.05$).\label{tab:sleepmami}}
  \begin{tabular}{ll c cc cc cc}
    \toprule
           &           & Staging     & \multicolumn{2}{c}{Arousal} & \multicolumn{2}{c}{Apnoea} & \multicolumn{2}{c}{Desat.}                                           \\
    \cmidrule(lr){4-5} \cmidrule(lr){6-7} \cmidrule(lr){8-9}
    Cohort & Method    & $\kappa$    & AUROC                       & AUPRC                      & AUROC                      & AUPRC       & AUROC       & AUPRC       \\
    \midrule
    CHAT   & SleepMaMi & $0.799$     & $0.946$                     & $0.742$                    & $0.893$                    & $0.459$     & $0.909^{*}$ & $0.699^{*}$ \\
           & Hypnos    & $0.852^{*}$ & $0.965^{*}$                 & $0.793^{*}$                & $0.925^{*}$                & $0.460$     & $0.863$     & $0.502$     \\
    \midrule
    MESA   & SleepMaMi & $0.754$     & $0.897$                     & $0.717$                    & $0.789$                    & $0.455^{*}$ & $0.915^{*}$ & $0.885^{*}$ \\
           & Hypnos    & $0.803^{*}$ & $0.927^{*}$                 & $0.766^{*}$                & $0.787$                    & $0.436$     & $0.833$     & $0.776$     \\
    \bottomrule
  \end{tabular}
\end{table}

\begin{table}[h]
  \centering
  \small
  \caption{\textbf{Event detection: Hypnos vs.\ SleepMaMi.} Per-second macro-F1 and event-level F1 (IoU $0.3$). $^{*}$ indicates a significant difference, using the same test as \Cref{tab:sleepmami}.\label{tab:sleepmami-events}}
  \begin{tabular}{ll cc cc}
    \toprule
           &           & \multicolumn{2}{c}{Arousal} & \multicolumn{2}{c}{Apnoea}                             \\
    \cmidrule(lr){3-4} \cmidrule(lr){5-6}
    Cohort & Method    & Per-second                  & Event                      & Per-second  & Event       \\
    \midrule
    CHAT   & SleepMaMi & $0.577$                     & $0.603$                    & $0.327$     & $0.333$     \\
           & Hypnos    & $0.695^{*}$                 & $0.742^{*}$                & $0.292$     & $0.394^{*}$ \\
    \midrule
    MESA   & SleepMaMi & $0.495$                     & $0.530$                    & $0.365$     & $0.405$     \\
           & Hypnos    & $0.632^{*}$                 & $0.681^{*}$                & $0.382^{*}$ & $0.453^{*}$ \\
    \bottomrule
  \end{tabular}
\end{table}

\subsection{Rhythm-conditioned ECG generation}\label{app:ecg-generation}

\begin{figure}[h]
    \centering
    \includegraphics[width=\linewidth]{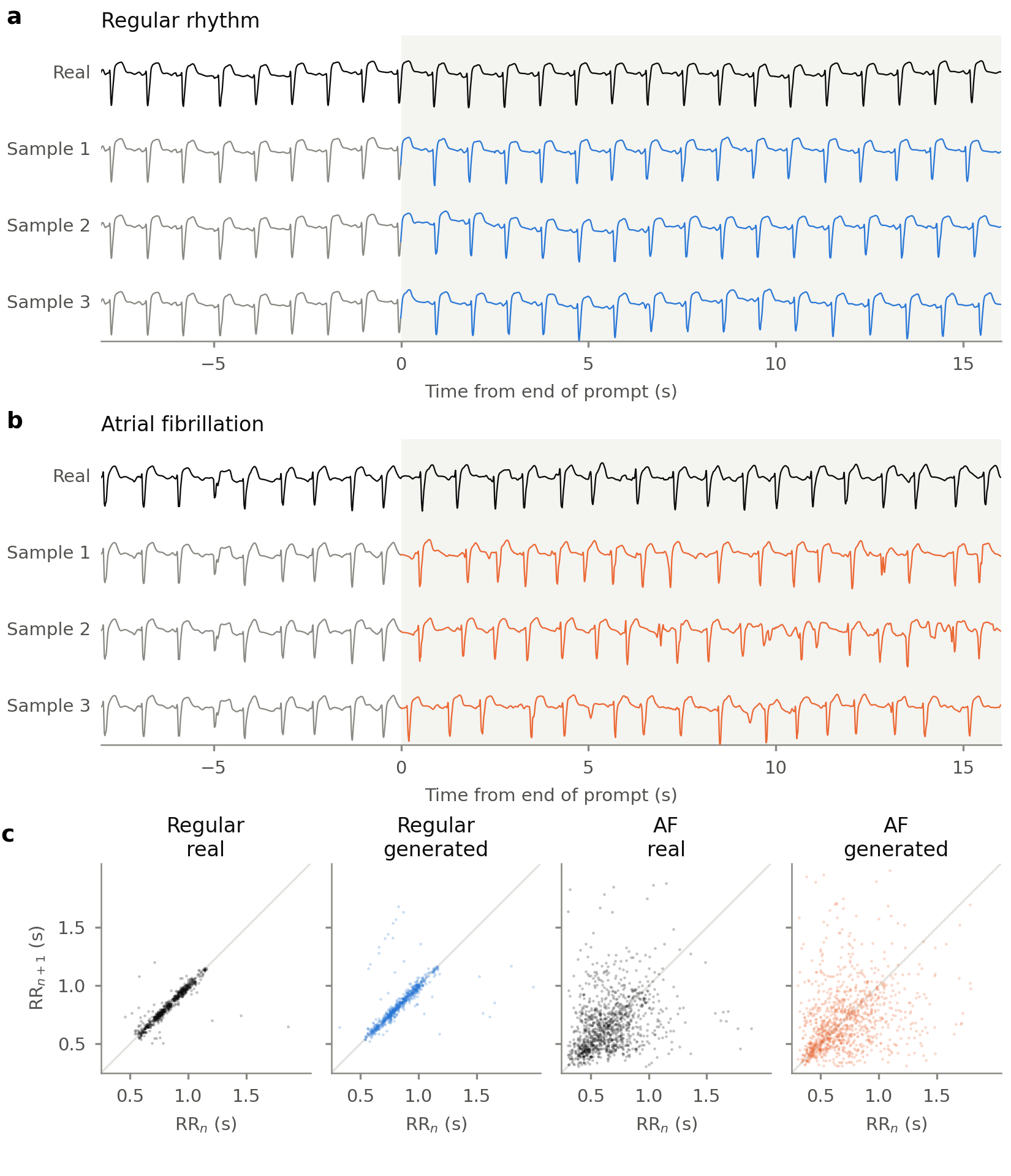}
    \caption{\textbf{Rhythm-conditioned ECG generation on MIT-BIH AFDB.} (a, b) Real continuation (black) and three generated samples (colour) from a prompt (grey; final 8\,s of 60\,s shown, first 16\,s of 30\,s generated) for regular rhythm and AF in the same patient. (c) Poincar\'e plots of successive RR intervals for real and generated electrocardiograms.\label{fig:ecg-rhythm}}
  \end{figure}

  As an example of Hypnos' ability to model stochastic structure, we auto-regressively generated single-lead ECG conditioned on either atrial fibrillation (AF) or regular rhythm using the MIT-BIH Atrial Fibrillation Database (AFDB)~\cite{moodyNewMethodDetecting1983}. From a 60\,s input segment, we generate 30\,s signal continuations. \Cref{fig:ecg-rhythm} shows that conditioning on inputs during atrial fibrillation leads to more irregular beat-to-beat intervals in the generated signals.

\section{Extended Limitations and Future Work}\label{app:limitations}

\paragraph{Clinical Outcomes} In this work, we evaluated the quality of Hypnos' embeddings for existing diagnostic tasks such as sleep stage classification. Due to data availability, we were unable to evaluate Hypnos on broader clinical outcomes data like SleepFM~\cite{thapaMultimodalSleepFoundation2026}. Evaluating Hypnos on a similar broad range of clinical outcomes is an important direction for future work. Given the large performance improvements observed across our evaluations, we expect Hypnos would also enable a significant improvement in these tasks.

\paragraph{Biomarker Discovery}
There are well-known issues with existing sleep stage definitions. For example, small changes to the brittle definition of deep (N3) sleep can have a significant effect on downstream analyses, such as how sleep architecture purportedly varies with age for women~\cite{davidsonItTimeRevisit2025a}. A promising direction is using Hypnos for data-driven biomarker discovery and identifying better continuous measures of health. For example, identifying latent directions or sparse features predictive of incident neurodegenerative or cardiovascular disease. The generative capability of Hypnos could be used for visual interpretation of such features in input space.

\section{Broader Impact}\label{section:impact}

Sleep disorders such as obstructive sleep apnoea and insomnia are common but routinely under-diagnosed, in part because polysomnography is expensive to record and labour-intensive to score. Foundation models for physiological signals offer a path to reducing the cost of sleep analysis and broadening access to sleep medicine, through improved generalisation with fewer labelled examples and via fast adaptation to novel sensor configurations, e.g.\ from consumer wearables. The recordings used in this work are drawn from nine public datasets, predominantly curated by the National Sleep Research Resource (NSRR). These cohorts are each biased by exclusion criteria; for example, MrOS is a study that solely recruited older males. For real-world deployment, Hypnos should be evaluated on populations of interest, and we encourage further work auditing performance across demographic subgroups.

\section{Additional Acknowledgements}
\label{section:appendix:acks}
The Sleep Heart Health Study (SHHS) was supported by National Heart, Lung, and Blood Institute cooperative agreements U01HL53916 (University of California, Davis), U01HL53931 (New York University), U01HL53934 (University of Minnesota), U01HL53937 and U01HL64360 (Johns Hopkins University), U01HL53938 (University of Arizona), U01HL53940 (University of Washington), U01HL53941 (Boston University), and U01HL63463 (Case Western Reserve University).

The Multi-Ethnic Study of Atherosclerosis (MESA) Sleep Ancillary study was funded by NIH-NHLBI Association of Sleep Disorders with Cardiovascular Health Across Ethnic Groups (RO1 HL098433). MESA is supported by NHLBI funded contracts HHSN268201500003I, N01-HC-95159, N01-HC-95160, N01-HC-95161, N01-HC-95162, N01-HC-95163, N01-HC-95164, N01-HC-95165, N01-HC-95166, N01-HC-95167, N01-HC-95168 and N01-HC-95169 from the National Heart, Lung, and Blood Institute, and by cooperative agreements UL1-TR-000040, UL1-TR-001079, and UL1-TR-001420 funded by NCATS.

The Wisconsin Sleep Cohort Study was supported by the U.S. National Institutes of Health, National Heart, Lung, and Blood Institute (R01HL62252), National Institute on Aging (R01AG036838, R01AG058680), and the National Center for Research Resources (1UL1RR025011).

The Childhood Adenotonsillectomy Trial (CHAT) was supported by the National Institutes of Health (HL083075, HL083129, UL1-RR-024134, UL1 RR024989).

The Cleveland Family Study (CFS) was supported by grants from the National Institutes of Health (HL46380, M01 RR00080-39, T32-HL07567, RO1-46380).

The Cleveland Children's Sleep and Health Study (CCSHS) was supported by grants from the National Institutes of Health (RO1HL60957, K23 HL04426, RO1 NR02707, M01 Rrmpd0380-39).

The National Heart, Lung, and Blood Institute provided funding for the ancillary MrOS Sleep Study, ``Outcomes of Sleep Disorders in Older Men," under the following grant numbers: R01 HL071194, R01 HL070848, R01 HL070847, R01 HL070842, R01 HL070841, R01 HL070837, R01 HL070838, and R01 HL070839.

NCH Sleep DataBank was supported by the National Institute of Biomedical Imaging and Bioengineering of the National Institutes of Health under Award Number R01EB025018.

\newpage

\section*{NeurIPS Paper Checklist}

\begin{enumerate}

  \item {\bf Claims}
  \item[] Question: Do the main claims made in the abstract and introduction accurately reflect the paper's contributions and scope?
  \item[] Answer: \answerYes{}
  \item[] Justification: \Cref{sec:experiments} shows that Hypnos outperforms prior models on most sleep analysis tasks and datasets evaluated, that performance improves with model scale, and that the model is robust to missing modalities.
  \item[] Guidelines:
        \begin{itemize}
          \item The answer \answerNA{} means that the abstract and introduction do not include the claims made in the paper.
          \item The abstract and/or introduction should clearly state the claims made, including the contributions made in the paper and important assumptions and limitations. A \answerNo{} or \answerNA{} answer to this question will not be perceived well by the reviewers.
          \item The claims made should match theoretical and experimental results, and reflect how much the results can be expected to generalise to other settings.
          \item It is fine to include aspirational goals as motivation as long as it is clear that these goals are not attained by the paper.
        \end{itemize}

  \item {\bf Limitations}
  \item[] Question: Does the paper discuss the limitations of the work performed by the authors?
  \item[] Answer: \answerYes{}
  \item[] Justification: Limitations are discussed in a dedicated section before conclusions.

  \item {\bf Theory assumptions and proofs}
  \item[] Question: For each theoretical result, does the paper provide the full set of assumptions and a complete (and correct) proof?
  \item[] Answer: \answerNA{}
  \item[] Justification: Paper does not contain theoretical results or proofs.

  \item {\bf Experimental result reproducibility}
  \item[] Question: Does the paper fully disclose all the information needed to reproduce the main experimental results of the paper to the extent that it affects the main claims and/or conclusions of the paper (regardless of whether the code and data are provided or not)?
  \item[] Answer: \answerYes{}
  \item[] Justification: All implementation details are discussed in the main paper or sign-posted to the Appendix. This includes pre-processing details, dataset splits, model hyper-parameters, and model training details.
  \item[] Guidelines:
        \begin{itemize}
          \item The answer \answerNA{} means that the paper does not include experiments.
          \item If the paper includes experiments, a \answerNo{} answer to this question will not be perceived well by the reviewers: Making the paper reproducible is important, regardless of whether the code and data are provided or not.
          \item If the contribution is a dataset and\slash or model, the authors should describe the steps taken to make their results reproducible or verifiable.
          \item Depending on the contribution, reproducibility can be accomplished in various ways. For example, if the contribution is a novel architecture, describing the architecture fully might suffice, or if the contribution is a specific model and empirical evaluation, it may be necessary to either make it possible for others to replicate the model with the same dataset, or provide access to the model. In general, releasing code and data is often one good way to accomplish this, but reproducibility can also be provided via detailed instructions for how to replicate the results, access to a hosted model (e.g., in the case of a large language model), releasing of a model checkpoint, or other means that are appropriate to the research performed.
          \item While NeurIPS does not require releasing code, the conference does require all submissions to provide some reasonable avenue for reproducibility, which may depend on the nature of the contribution. For example
                \begin{enumerate}
                  \item If the contribution is primarily a new algorithm, the paper should make it clear how to reproduce that algorithm.
                  \item If the contribution is primarily a new model architecture, the paper should describe the architecture clearly and fully.
                  \item If the contribution is a new model (e.g., a large language model), then there should either be a way to access this model for reproducing the results or a way to reproduce the model (e.g., with an open-source dataset or instructions for how to construct the dataset).
                  \item We recognize that reproducibility may be tricky in some cases, in which case authors are welcome to describe the particular way they provide for reproducibility. In the case of closed-source models, it may be that access to the model is limited in some way (e.g., to registered users), but it should be possible for other researchers to have some path to reproducing or verifying the results.
                \end{enumerate}
        \end{itemize}

  \item {\bf Open access to data and code}
  \item[] Question: Does the paper provide open access to the data and code, with sufficient instructions to faithfully reproduce the main experimental results, as described in supplemental material?
  \item[] Answer: \answerNo{}
  \item[] Justification: All datasets are publicly available. We release pre-trained model weights, an inference library and a worked example of embedding generation and linear probing at \url{https://github.com/joncarter1/hypnos}. Code for pre-training and for reproducing all experiments is not currently released.

  \item {\bf Experimental setting/details}
  \item[] Question: Does the paper specify all the training and test details (e.g., data splits, hyperparameters, how they were chosen, type of optimizer) necessary to understand the results?
  \item[] Answer: \answerYes{}
  \item[] Justification: Data splits follow prior work. All hyper-parameters are provided and were either carefully informed by prior work or tuned using validation data.

  \item {\bf Experiment statistical significance}
  \item[] Question: Does the paper report error bars suitably and correctly defined or other appropriate information about the statistical significance of the experiments?
  \item[] Answer: \answerYes{}
  \item[] Justification: Our primary results compare with prior work using Wilcoxon signed-rank tests computed over a large number of test subjects (paired across methods), as described in \Cref{sec:experiments}. We did not repeat our main pre-training experiments with different random seeds due to computational resources. However, in \Cref{section:stability} we investigate variance due to random seeds using smaller scale unimodal models, reporting mean $\pm$ standard deviation across three seeds, and find this to be very low relative to our performance improvements over prior work.

  \item {\bf Experiments compute resources}
  \item[] Question: For each experiment, does the paper provide sufficient information on the computer resources (type of compute workers, memory, time of execution) needed to reproduce the experiments?
  \item[] Answer: \answerYes{}
  \item[] Justification: In \Cref{section:compute} we report GPU types, memory usage and training time for all major model components. We additionally provide an estimate of total compute usage.

  \item {\bf Code of ethics}
  \item[] Question: Does the research conducted in the paper conform, in every respect, with the NeurIPS Code of Ethics \url{https://neurips.cc/public/EthicsGuidelines}?
  \item[] Answer: \answerYes{}
  \item[] Justification: We have reviewed the Code of Ethics and believe the paper conforms to them.

  \item {\bf Broader impacts}
  \item[] Question: Does the paper discuss both potential positive societal impacts and negative societal impacts of the work performed?
  \item[] Answer: \answerYes{}
  \item[] Justification: Discussed in \Cref{section:impact}.

  \item {\bf Safeguards}
  \item[] Question: Does the paper describe safeguards that have been put in place for responsible release of data or models that have a high risk for misuse (e.g., pre-trained language models, image generators, or scraped datasets)?
  \item[] Answer: \answerNA{}
  \item[] Justification: We do not believe release of code or models requires safeguards. Data used is already publicly available.

  \item {\bf Licenses for existing assets}
  \item[] Question: Are the creators or original owners of assets (e.g., code, data, models), used in the paper, properly credited and are the license and terms of use explicitly mentioned and properly respected?
  \item[] Answer: \answerYes{}
  \item[] Justification: Datasets used have been appropriately cited as requested by the National Sleep Research Resource which provided access. External ECG datasets, and the code and weights of baseline models, are credited by citation.

  \item {\bf New assets}
  \item[] Question: Are new assets introduced in the paper well documented and is the documentation provided alongside the assets?
  \item[] Answer: \answerYes{}
  \item[] Justification: The model weights and inference library are released under the MIT license, documented in a README and model card, with a worked example notebook.

  \item {\bf Crowdsourcing and research with human subjects}
  \item[] Question: For crowdsourcing experiments and research with human subjects, does the paper include the full text of instructions given to participants and screenshots, if applicable, as well as details about compensation (if any)?
  \item[] Answer: \answerNA{}
  \item[] Justification: Paper uses retrospective, publicly available datasets. Specific details regarding data collection can be found in the original works which are cited.

  \item {\bf Institutional review board (IRB) approvals or equivalent for research with human subjects}
  \item[] Question: Does the paper describe potential risks incurred by study participants, whether such risks were disclosed to the subjects, and whether Institutional Review Board (IRB) approvals (or an equivalent approval/review based on the requirements of your country or institution) were obtained?
  \item[] Answer: \answerNA{}
  \item[] Justification: All datasets used in this work are publicly available and have been de-identified or fully anonymised, thus not requiring IRB approval.

  \item {\bf Declaration of LLM usage}
  \item[] Question: Does the paper describe the usage of LLMs if it is an important, original, or non-standard component of the core methods in this research? Note that if the LLM is used only for writing, editing, or formatting purposes and does \emph{not} impact the core methodology, scientific rigor, or originality of the research, declaration is not required.
  \item[] Answer: \answerNA{}
  \item[] Justification: LLMs are not an original or core component of the methods described in this work.

\end{enumerate}

\end{document}